\documentclass{article}

\usepackage[main, final]{neurips_2026}

\usepackage[utf8]{inputenc} 
\usepackage[T1]{fontenc}    
\usepackage{hyperref}       
\usepackage{url}            
\usepackage{booktabs}       
\usepackage{amsfonts}       
\usepackage{nicefrac}       
\usepackage{microtype}      
\usepackage{xcolor}         
\usepackage{microtype}
\usepackage{graphicx}
\usepackage{subfigure}
\usepackage{booktabs} 
\usepackage{wrapfig}
\usepackage{amsmath}
\usepackage{amssymb}
\usepackage{mathtools}
\usepackage{amsthm}
\usepackage{algorithm}
\usepackage{algpseudocode}
\usepackage{xcolor}   
\usepackage{nccmath}
\usepackage{cleveref} 

\usepackage{tcolorbox}
\tcbuselibrary{listings, breakable} 
\usepackage{enumitem}
\usepackage{listings}

\definecolor{lightgrey}{RGB}{240, 240, 240}
\definecolor{darkgray}{RGB}{100, 100, 100}

\tcbset{
    skillbox/.style={
        colback=lightgrey,
        colframe=darkgray,
        boxrule=0.5pt,      
        arc=2mm,
        left=2mm, right=2mm, top=2mm, bottom=2mm, 
        fonttitle=\bfseries\small, 
        fontupper=\small\raggedright,   
        colbacktitle=gray,
        width=\linewidth,   
        before=\par\smallskip\noindent,
        after=\par\smallskip
    }
}
\newtcblisting{promptbox}[1][Prompt]{
    breakable,              
    colback=gray!5,         
    colframe=gray!50,       
    boxrule=1pt,            
    top=10pt,               
    bottom=10pt,            
    left=10pt,              
    right=10pt,             
    arc=5pt,                
    title={#1},             
    coltitle=black,         
    fonttitle=\bfseries,    
    colbacktitle=gray!20,   
    titlerule=1pt,          
    listing only,           
    listing options={
        basicstyle=\ttfamily\small,
        breaklines=true,        
        breakatwhitespace=false,
        columns=fullflexible,   
        keepspaces=true         
    }
}
\newcommand{\code}[1]{\texttt{\small #1}}
\newcommand{\Name}{CODE-SHARP}

\title{CODE-SHARP: Continuous Open-ended Discovery and Evolution of Skills as Hierarchical Reward Programs}

\author{%
  Richard Bornemann \\
  BOLD, Imperial College London \\
  \texttt{r.bornemann25@imperial.ac.uk} \\
  \And
  Pierluigi Vito Amadori \\
  Sony Interactive Entertainment \\
  \AND
  Antoine Cully \\
  BOLD, Imperial College London \\
}

\begin{document}
\setlength{\textfloatsep}{1\baselineskip}

\maketitle

\begin{abstract}
A core quality of general intelligence is the ability to open-endedly expand and evolve its set of mastered skills autonomously. While recent Foundation Model (FM) driven approaches have shown promising results towards this goal, they typically rely on significant human-in-the-loop engineering, limiting their transferability to novel environments. To address this, we introduce Continuous Open-ended Discovery and Evolution of Skills as Hierarchical Reward Programs (CODE-SHARP), a framework that leverages FMs to open-endedly grow and evolve an archive of Python programs encoding skills to train a generalist agent policy entirely from scratch via reinforcement learning, directly from source code. These programs, termed Skills as Hierarchical Reward Programs (SHARPs), each encode a local success condition and a set of prerequisites delegated to previously discovered SHARPs. At runtime, SHARPs dynamically route the agent through their prerequisite chain based on the current state, rewarding each completion along the way, requiring the agent to learn only the marginal behaviour each new SHARP introduces, enabling efficient learning of long-horizon skills without any pre-defined rewards. On Craftax-Classic and XLand-MiniGrid, agents trained fully autonomously by CODE-SHARP outperform previous works by 6x and 2.7x in median performance and are the only agents capable of crafting iron tools and mining diamonds. On AI2-THOR, CODE-SHARP achieves 66.7\% median success versus 0\% for ELLM and OMNI, using only the initial environment state instead of the full source code. Scaled to Craftax-Extended, CODE-SHARP trains a generalist agent on over 90 discovered SHARPs, enabling the agent to solve challenging long-horizon tasks zero-shot, matching agents trained on ground-truth rewards.
\end{abstract}

\section{Introduction}
A core quality of general intelligence is the ability to open-endedly expand and evolve its set of mastered skills autonomously \citep{hughes2024openendednessessentialartificialsuperhuman}. Consequently, recent years have seen significant progress in open-ended skill discovery for artificial intelligence \citep{wang2019paired, wang2020enhancedpoetopenendedreinforcement, adaptiveagentteam2023humantimescaleadaptationopenendedtask}. Recent Foundation Model (FM) based agents, such as Voyager \citep{wang2023voyager} and Jarvis \citep{wang2024jarvis}, have demonstrated impressive results on open-ended skill discovery in environments like Minecraft \citep{fan2022minedojo}, and similar successes have been shown in robotics with FM-based planners \citep{ahn2022can, zhang2023bootstrap}. However, these methods rely completely on human engineering to bridge the gap between high-level planning and low-level action execution via hand-crafted APIs \citep{wang2023voyager, wang2024jarvis} or models pre-trained on massive datasets of annotated human trajectories \citep{wang2024jarvis, fan2022minedojo, zhang2023bootstrap}. The scale of this engineering is considerable: the Mineflayer API used across open-ended skill discovery work in Minecraft comprises 13,000 lines of JavaScript, while the MineDojo \citep{fan2022minedojo} agent was trained on 33 years of annotated human gameplay, creating a severe bottleneck for transferring such methods to novel environments. Exploration with LLMs (ELLM) \citep{du2023guiding} and Open-Endedness via Human Notions of Interestingness (OMNI) \citep{zhangomni} reduce these requirements by training goal-conditioned agents from scratch to handle low-level action execution. However, both methods still depend on hand-engineered state captioners and translation functions requiring expert knowledge of the environment's actions, objects, and dynamics.
\begin{figure*}[t]
    \centering
    \includegraphics[width=0.9\textwidth]{Figures/ProposalMutationPipeline.pdf}
    \caption{\Name{} consists of two FM-driven iterative processes that discover and evolve skill encoding SHARPs. The \textbf{skill proposal generator}, \textbf{implementor}, and \textbf{judge} generate and filter novel SHARPs before environment evaluation. The \textbf{skill mutation generator} and \textbf{implementor} produce mutated versions of existing SHARPs, evaluated directly in the environment.}
    \label{fig:proposal_pipeline}
\end{figure*}

To overcome these limitations and move towards a more general and transferable approach to autonomous open-ended skill discovery, we introduce \textbf{C}ontinuous \textbf{O}pen-ended \textbf{D}iscovery and \textbf{E}volution of \textbf{S}kills as \textbf{H}ier\textbf{a}rchical \textbf{R}eward \textbf{P}rograms (\Name). \Name{} leverages FMs to autonomously and open-endedly grow an archive of Python programs encoding skills, termed Skills as Hierarchical Reward Programs (SHARPs), to continuously train a goal-conditioned policy, representing a generalist agent, to acquire increasingly complex skills entirely from scratch via reinforcement learning. Rather than depending on hand-crafted APIs or curated demonstrations, \Name{} ingests the environment source code directly, substantially reducing the amount of human engineering required. \Name{} structures the archive of discovered SHARPs as a continuously growing, interconnected graph, where each SHARP is directly composed of SHARPs discovered in previous iterations. Each SHARP guides the agent through its chain of prerequisite SHARPs based on the current state, directly instructing it to reuse already mastered skills and reducing learning to the marginal behaviour each new SHARP introduces. Similarly, to define a novel SHARP, the FM needs only to specify this marginal behaviour and its immediate prerequisites, rather than encoding every substep required for completion, reducing the risk of errors introduced during SHARP specification. \Name{} consists of two open-ended iterative processes: 1) FM-driven open-ended discovery of SHARPs based on the current skill archive, and 2) FM-driven evolution of the archive through targeted mutation of existing SHARPs. Simultaneously, the agent is trained exclusively on the rewards derived from the growing archive of SHARPs.

On Craftax-Classic \citep{matthews2024craftax, hafner2021benchmarking} and XLand-MiniGrid \citep{nikulin2024xland}, agents trained entirely from scratch by SHARPs discovered by \Name{} outperform baselines by a factor of 6x and 2.7x in median performance, respectively, and are the only agents capable of reliably crafting iron tools and mining diamonds. On AI2-THOR \citep{kolve2017ai2}, a photorealistic 3D household environment, \Name{} achieves 66.7\% median success across 98 achievements versus 0\% for ELLM and OMNI, using only the initial environment state instead of the full source code. To challenge \Name{}'s open-endedness, we evaluate on Craftax-Extended \citep{matthews2024craftax} with 4× the proposal iterations and 20× the training budget of our Craftax-Classic experiments. In this setting, \Name{} discovers on average 90 diverse SHARPs, which train an agent to match or outperform baselines trained on ground-truth rewards zero-shot on four challenging benchmarks.

To summarise, our principal contributions are: \textbf{1)} \Name{}, an autonomous open-ended skill discovery framework that trains generalist agents via reinforcement learning directly from environment source code, substantially reducing the human engineering required compared to prior work; \textbf{2)} the formalisation of Skills as Hierarchical Reward Programs (SHARPs), Python programs generated by FMs and built from previously discovered SHARPs, which reward the agent upon reaching a specified goal state; and \textbf{3)} a recursive transition operator that identifies the most immediate actionable prerequisite skill at each step, guiding the agent through the SHARP's prerequisite chain and reducing learning to only the novel behaviour each newly discovered SHARP introduces.

\section{Method}
\label{sec:method}
\Name{} facilitates open-ended unsupervised skill discovery by curating a continuously growing skill archive implemented as a directed acyclic graph of Skills as Hierarchical Reward Programs (SHARPs). Each SHARP is represented as an FM-generated Python program encoding a skill's success condition and prerequisite dependencies (\Cref{subsec:sharps}). Two FM-driven loops grow and refine this archive: a propose–implement–judge pipeline adds novel SHARPs for which the agent shows measurable learning progress, while a mutation loop improves existing SHARPs by revising their condition functions and prerequisite assignments. Throughout, a goal-conditioned policy $\pi$, representing the agent, is trained via PPO exclusively on SHARP-generated rewards. The full algorithm is summarised in \Cref{app:algorithm}.

\subsection{Problem Formalisation}
We model the environment as a Goal-Conditioned MDP $\mathcal{M} = 
\langle \mathcal{S}, \mathcal{A}, \mathcal{G}, P, \gamma \rangle$, with 
state and action spaces $\mathcal{S}, \mathcal{A}$, transition dynamics 
$P: \mathcal{S} \times \mathcal{A} \to \Delta(\mathcal{S})$, discount 
factor $\gamma \in [0,1)$, and semantic goal space $\mathcal{G}$.

We formalise open-ended skill discovery as the iterative construction of 
a dynamic skill archive, structured as a directed acyclic graph 
$\Lambda_t = (\mathcal{V}_t, \mathcal{E}_t)$, where each node $\sigma \in 
\mathcal{V}_t$ is a Python program encoding a skill corresponding to a semantic goal in 
$\mathcal{G}$, and edges encode prerequisite dependencies between skills. The skill discovery process corresponds to a mapping $\Lambda_{t+1} = f(\Lambda_t)$ that identifies and adds novel, learnable skills to the graph.

\subsection{Skills as Hierarchical Reward Programs}
\label{subsec:sharps}
\begin{wrapfigure}{r}{0.25\columnwidth}
    \centering
    \includegraphics[width=1.0\linewidth]{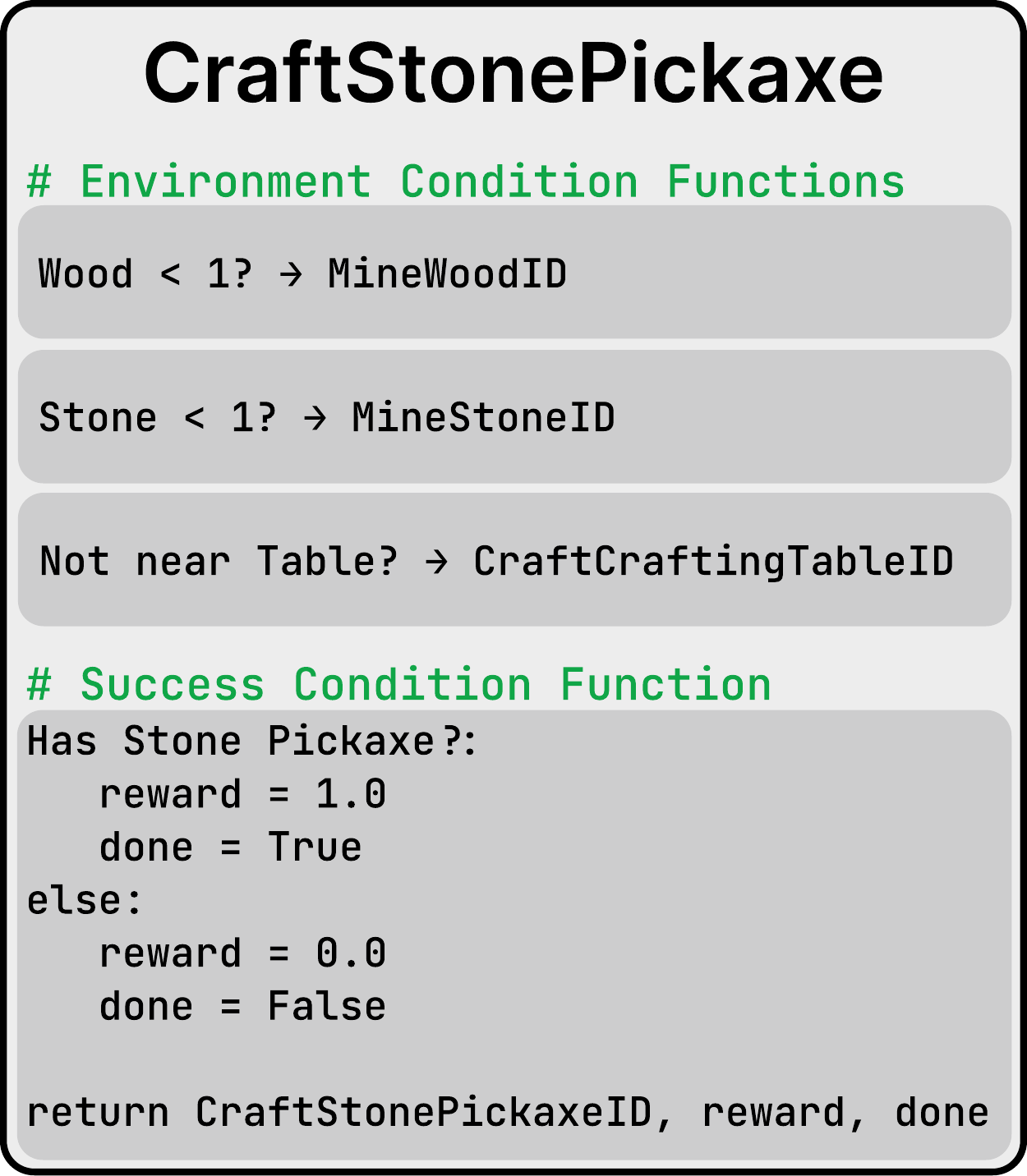}
    \vspace{-1em}
    \caption{Pseudo-code of a SHARP}
    \vspace{-1em}
    \label{fig:sharp}
\end{wrapfigure}
Each node in the archive is represented by a SHARP, an FM-generated 
executable Python program defined as $\sigma = \langle 
\phi_\sigma, \psi_\sigma, \rho_\sigma \rangle$. The FM defines $\phi_\sigma: \mathcal{S} \times \mathcal{S} \to \{0, 1\}$ 
encoding the semantic goal of a skill in code (e.g.\ 
\texttt{agent.inventory.stone\_pickaxe >= 1}) and acts as the termination 
condition. The initiation set $\psi_\sigma = ((c_i, u_i))_{i=1}^{m}$ is 
an ordered list of (condition, prerequisite) pairs, where each condition 
function $c_i$ verifies whether $s_t$ satisfies an immediate requirement for 
attempting $\phi_\sigma$ (e.g.\ \texttt{agent.inventory.stone >= 1}), and 
each $u_i \in \mathcal{V}_t$ is an existing SHARP from the archive (e.g.\ 
\texttt{MineStone}) selected by the FM to be invoked when $c_i$ is unmet.  These FM-assigned 
dependencies define the directed edges $(u_i, \sigma) \in 
\mathcal{E}_t$. At runtime, SHARPs receive the current and previous 
states $(s_t, s_{t-1})$ as input and return the reward achieved in $s_t$.

Each episode, a target SHARP $\sigma_{\text{target}} \in \mathcal{V}_t$ 
is sampled from a distribution $\mathcal{P}(\mathcal{V}_t)$ over the 
archive. To resolve the immediately actionable skill given the current state, we define a recursive transition operator $\mathcal{T}: \mathcal{V}_t \times \mathcal{S} \to \mathcal{V}_t$ that descends the prerequisite hierarchy of $\sigma_{\text{target}}$ until reaching a skill whose conditions are all satisfied in $s_t$:
\begin{equation}
    \medmath{
    \mathcal{T}(\sigma, s) =
    \begin{cases}
        \mathcal{T}(u^{\sigma}_i, s) & \text{if } \exists\, i : c^{\sigma}_i(s) = 0 \;\land\;
              (\forall\, j < i,\; c^{\sigma}_j(s) = 1) \\
        \sigma & \text{otherwise}
    \end{cases}
    }
\end{equation}
Concretely, given $\sigma_{\text{target}}$ = \texttt{CraftStonePickaxe} and its first unmet condition \texttt{agent.inventory.stone >= 1}, $\mathcal{T}$ recurses into \texttt{MineStone}; if \texttt{MineStone} in turn has an unmet condition such as \texttt{agent.inventory.wood\_pickaxe >= 1}, the recursion descends further into \texttt{CraftWoodPickaxe}, and so on until reaching a SHARP whose conditions are all satisfied in $s_t$. The resulting $\sigma_{\text{active}} = \mathcal{T}(\sigma_{\text{target}}, s_t)$ simultaneously conditions the agent's goal-conditioned policy $\pi: \mathcal{S} \times \mathcal{V}_t \to \Delta(\mathcal{A})$ and defines the reward $R(s, a, s' \mid \sigma_{\text{active}}) = \alpha(\sigma_{\text{active}}) \cdot \mathbb{I}[\phi_{\sigma_{\text{active}}}(s', s) = 1]$ to train it. This traversal repeats at every environment step, allowing $\sigma_{\text{active}}$ to adapt to changes in state, so the agent is continuously guided and rewarded through the full chain of prerequisite SHARPs towards $\sigma_{\text{target}}$. Consequently, to learn a novel skill, the hierarchy automatically guides the agent into environment states from which only the marginal new behaviour needs to be learned, rather than requiring it to rediscover the full prerequisite sequence from scratch.

\subsection{Open-Ended Skill Discovery}
\label{subsec:discovery}

\Name{} adopts an FM-based propose-implement-judge approach to generating
novel SHARPs \citep{faldor2024omni}. Throughout both the discovery and
evolution loops, every FM call receives the same base context: the current skill archive, the environment source code, failed proposals from prior iterations, and any optional auxiliary documentation such as tutorials. Component-specific inputs are noted below. All prompts
are provided in \Cref{app:discovery_prompts}.

\textbf{Skill Proposal Generator and Implementor}: The generator produces $n$ SHARP candidates as pseudo-code, each specifying a high-level description, a binary success condition $\phi$, and a dictionary mapping condition
functions of the proposed SHARP candidate to prerequisite SHARPs in the existing archive. To encourage diversity, generation is conditioned on $n$ skill categories sampled uniformly with replacement from a predefined set of category heuristics. The skill proposal implementor translates the pseudo-code candidates into SHARPs in Python code, based on a general SHARP class template.

\textbf{Skill Proposal Judge}: Recent work has demonstrated that FMs serve as effective proxies for human judgments of interestingness, novelty, and skill learnability \citep{zhangomni, faldor2024omni}. \Name{} therefore employs an FM-based judge to evaluate and select up to two of the implemented SHARP candidates for learnability evaluation based on three criteria:  \textit{1) Correctness:} The implemented SHARP must follow the provided template and correctly map all required conditions $c_i$ to valid prerequisites $u_i$ in $\mathcal{V}$. \textit{2) Feasibility:} The skill must be learnable given the agent's current capabilities. \textit{3) Novelty:} The skill must target a distinct region of the skill space compared to existing skills. A more detailed analysis of the Judge's performance is presented in \Cref{app:llm_judge_analysis}.

\textbf{Learnability Evaluation}: If a selected candidate fails to compile, the code and error trace are returned to the implementor for iterative refinement, repeated up to three times. A copy of the goal-conditioned agent is trained on the entire archive with the candidate SHARPs added to evaluate their learnability. We define learning progress as a success rate improvement of $\tau = 0.05$ between the first and final evaluation iterations, set to filter spurious successes while retaining gradually-learned long-horizon skills. SHARPs meeting this criterion enter the archive, with the rest being logged as failed proposals.

\subsection{Open-Ended Archive Evolution}
\label{subsec:refinement}

\Name{} leverages FMs to continuously evolve SHARPs already present in the skill archive to remedy potential misspecifications or replace existing prerequisite SHARPs with recently discovered, better fitting alternatives. Each iteration, SHARPs are sampled for mutation with
probability $P(k) \propto (1 - \rho_k)$, where $\rho_k$ is the SHARP's
current success rate. All prompts are provided in
\Cref{app:mutation_prompts}.

\textbf{Mutation Generation and Implementation}: The mutation proposal generator produces $m$ candidate mutations of the sampled SHARP, each conditioned on the parent SHARP's implementation and a mutation heuristic sampled from a predefined set. These heuristics target recurring, environment-independent failure modes, such as reordering condition functions, substituting prerequisite SHARPs with more suitable alternatives, or inserting missing condition–prerequisite pairs. The mutation implementor then translates each proposal into an executable SHARP in Python.

\textbf{Mutation Evaluation}: 
A central property of the SHARP formulation is that $\pi$ is conditioned only on $\sigma_{\text{active}}$ at each step, never on the structure of the parent skill that induced it. Consequently, mutations to a SHARP's condition functions and prerequisite assignments alter only how $\mathcal{T}$ traverses the archive, eliminating the need for gradient updates during mutation evaluation. We collect trajectories of the agent executing each candidate mutation in parallel, updating the archive in-place via $\sigma_k \leftarrow \sigma'_k$ if a mutation demonstrates a higher success rate than the currently best-performing $\sigma_k$.

\subsection{Agent Training}
\label{subsec:training}

The agent's policy $\pi$ is trained via PPO \citep{schulman2017proximalpolicyoptimizationalgorithms} in a continual, open-ended fashion exclusively on the rewards generated by the SHARPs discovered by \Name{}. Crucially, the FM is invoked only during the discovery and mutation loops. At runtime, only $\pi$ and the SHARP code execute, with no FM in the loop. At the beginning of each episode, a target SHARP $\sigma_{target}$ is uniformly sampled from the archive for the agent to complete. A new target SHARP is sampled if $\sigma_{target}$ is completed successfully or $200$ environment steps have elapsed without progress on $\sigma_{active}$. To condition the agent, an embedding of $\sigma_{active}$'s name is concatenated with the environment state $s_t$. We provide ablations of all training components in \Cref{app:detailed_ablation}.

\textbf{Prerequisite-Aware Opportunistic Sampling}: To exploit hard-to-reach environment states, the sampling of $\sigma_{target}$ is biased towards SHARPs whose condition functions are already satisfied but whose prerequisite SHARPs have low success rates, indicating that the current state presents a rare opportunity to practise the skill. Concretely, a sampling weight $\mathbf{B}_j$ is computed for each SHARP $\sigma_j$ as the inverse product of the success rates of its satisfied prerequisites: $\mathbf{B}_j = \frac{1}{\prod_{k=1}^{|\mathcal{V}|} (\rho_k +\epsilon)^{\mathbf{N}_{jk}}}$ where $\mathbf{N}_{jk} = 1$ if prerequisite $\sigma_k$ is assigned to a currently satisfied condition function of $\sigma_j$. SHARPs with a satisfied success condition in $s_t$ are masked for the final sampling distribution, as they would be trivial to complete.

\textbf{Adaptive Reward Scaling}: Due to the hierarchical nature of SHARPs, simply up-sampling low-performing skills to incentivise learning is insufficient, as it results in equal up-sampling of all SHARPs assigned as their prerequisites. To counter this and provide more direct learning feedback, we implement adaptive reward scaling 
\citep{kwon2025adaptiverewarddesignreinforcement}. Rewards $r_i$ provided by each SHARP $\sigma_i$ are scaled inversely to its success rate: $r_i = \min\left( \frac{1}{\rho_i + \epsilon}, 10.0 \right)$.

\section{Comparative Evaluation Experiments}
\label{sec:comp}
We first evaluate \Name{}'s ability to autonomously discover meaningful skills and train a goal-conditioned agent from scratch in three complex environments containing rich skill spaces: Craftax-Classic \citep{matthews2024craftax, hafner2021benchmarking}, XLand-MiniGrid \citep{nikulin2024xland}, and AI2-THOR \citep{kolve2017ai2}. AI2-THOR further tests transfer to a photorealistic 3D environment and skill discovery from the initial environment state without access to the full source code. We address two research questions. \textbf{RQ1}: Does \Name{} autonomously discover skills aligned with objectives that humans deem meaningful, across unseen environments? \textbf{RQ2}: Does the hierarchical composition of SHARPs enable the agent to master complex, long-horizon tasks that flat and non-compositional skill representations cannot achieve?

For RQ1, we compare against ELLM \citep{du2023guiding} and OMNI \citep{zhangomni}: like CODE-SHARP, both propose their own goals and train agents from scratch on the resulting rewards, without receiving a target task. Objective-driven reward-design methods such as Eureka \citep{maeureka}, ARM-FM \citep{castanyer2025arm}, and SCALAR \citep{zabounidis2026scalar} instead require an externally specified goal; supplying one here would change the open-ended discovery setting under evaluation. ELLM prompts the FM with a hand-crafted caption of the current state and the available actions at each step, then rewards the agent according to the cosine similarity between suggested goals and hand-crafted captions of state transitions. OMNI prompts the FM with tasks the agent can and cannot yet complete to propose new, learnable and interesting tasks, specifying each as a fixed sequence of object-state checks using a predefined schema. A learning-progress curriculum samples from this growing task set, while the agent receives a task reward only after satisfying the full sequence in order.

For RQ2, we assess SHARPs' step-adaptive conditioning on $\sigma_{\text{active}}$ through CODE-FRP, which replaces recursive traversal via $\mathcal{T}$ with Flat Reward Programs (FRPs) that explicitly enumerate complete prerequisite sequences. For example, where a SHARP corresponding to \texttt{MineStone} needs only specify the immediate prerequisite \texttt{CraftWoodPickaxe}, the corresponding FRP must enumerate the full chain \texttt{MineWood} $\rightarrow$ \texttt{PlaceCraftingTable} $\rightarrow$ \texttt{CraftWoodPickaxe}. For every discovered SHARP, we query an FM-based FRP generator to produce a counterpart conditioned on the SHARP's name, description, and success condition. Since both share identical names and success conditions, FRPs are evaluated using the agent trained on SHARPs, eliminating agent performance as a confounding factor. Performance differences therefore reflect two coupled effects: the loss of step-level adaptation and the increased specification burden of enumerating full prerequisite chains rather than immediate prerequisites.

For experiments on Craftax-Classic and XLand-MiniGrid all methods use a Transformer-XL \citep{dai2019transformerxlattentivelanguagemodels, hamon:hal-04659863v1} agent trained via PPO \citep{schulman2017proximalpolicyoptimizationalgorithms} for $1\times10^8$ steps on a single RTX A6000 ADA GPU. For \Name{} and OMNI we use 25 proposal iterations for Craftax-Classic and 40 for XLand-MiniGrid. Due to the prohibitive inference cost of ELLM's per-step querying of the FM, we reuse the authors' public FM cache for Craftax and pre-compute a cache of caption and FM response pairs for XLand-MiniGrid over nine days. For AI2-THOR we follow the evaluation protocol and agent architecture described in OMNI \citep{zhangomni} for all methods and train for $2.5\times10^6$ environment steps, with 100 proposal iterations for \Name{} and OMNI. For our Craftax-Classic and XLand-MiniGrid results we use Qwen3-235B-A22B-Thinking-2507 \citep{yang2025qwen3technicalreport} as the FM hosted across four RTX A6000 GPUs. For AI2-THOR, we  use Qwen3.5-27B as the FM, hosted on 2 RTX A6000 GPUs. 

We evaluate agent performance against a set of human-defined achievements in each environment. For \Name{}, CODE-FRP and OMNI, we utilise the discovered skill with the highest semantic similarity to the achievement name in embedding space to handle the conditioning of the respective agents, with ELLM conditioning the agent directly on the achievement name. A detailed breakdown of performance on all achievements is presented in \Cref{app:detailed_comparison}.

\subsection{Craftax-Classic Experiments}
\begin{figure*}[t]
    \centering
    \includegraphics[width=1.0\textwidth]{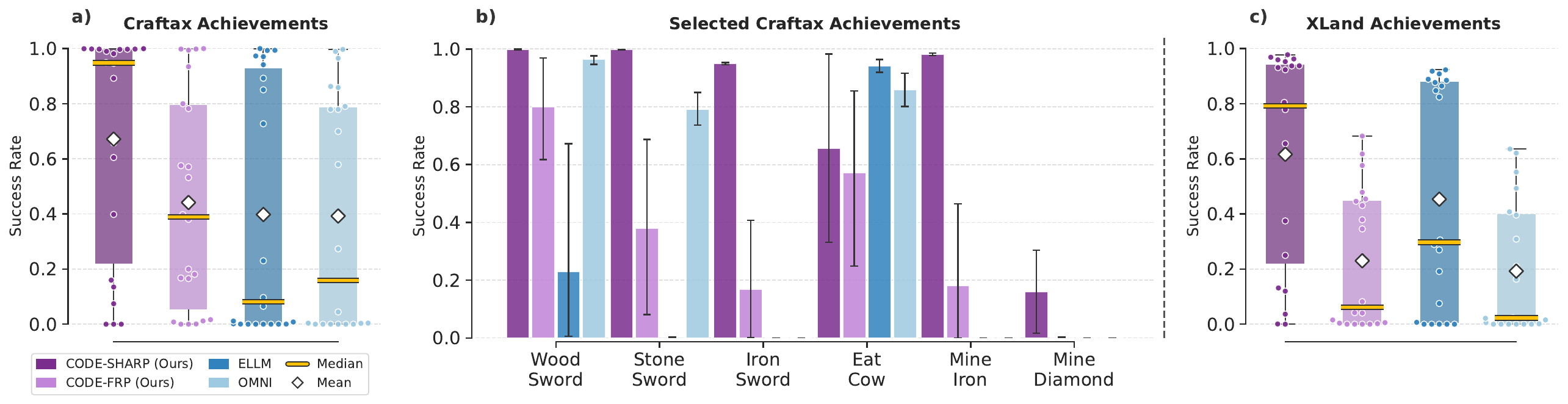}
    \caption{\textbf{a)}: Distribution of success rates on all achievements in Craftax-Classic aggregated over five runs. Shaded area indicates the interquartile range. \textbf{b)}: Average success rate on selected achievements on Craftax-Classic, error bars denote the bootstrapped 95\% CI around the mean. \textbf{c)}: Distribution of success rates on all achievements in XLand-MiniGrid across five runs.}
    \label{fig:craftax_XLand-Minigrid_results}
\end{figure*}
\textbf{Craftax-Classic} \citep{matthews2024craftax} is a procedural 2D environment based on Crafter \citep{hafner2021benchmarking}, featuring mechanics where agents collect resources, craft increasingly advanced tools, manage survival metrics, and fight mobs within a partially observable 64x64 grid. We evaluate performance across 22 predefined achievements, ranging from collecting basic resources to mining diamonds, the latter requiring up to 9 sequential prerequisite tasks. We remove all references to these achievements from the environment context provided to \Name{}, OMNI and CODE-FRP and run the experiments under standard Craftax-Classic environment settings without any modification.

\textbf{Results}: As shown in \Cref{fig:craftax_XLand-Minigrid_results}a, \Name{} outperforms both baselines, achieving a median success rate of 94.8\% and an average of 67.2\% across all 22 achievements, compared to 8.1\% (39.8\%) for ELLM and 15.8\% (39.2\%) for OMNI. \Cref{fig:craftax_XLand-Minigrid_results}b shows that the agent trained by ELLM is largely confined to shallow tasks requiring minimal prerequisites and fails to reliably learn longer-horizon skills. OMNI successfully discovers more advanced skills, training the agent to reliably craft stone tools. However, OMNI's programmatic representation of skills as a sequence of fixed target states fails to train the agent to master more complex behaviours. In contrast, \Name{}'s step-level adaptation of the active SHARP through $\mathcal{T}$ successfully trains the agent to reliably learn long-horizon skills, achieving over 94\% success rate on both the iron sword and iron pickaxe and 16\% on mining diamonds, with individual runs reaching up to 39\% success rate. The importance of this step-level adaptation is reinforced by the CODE-FRP ablation, which reaches only 38.9\% (44.2\%) and exhibits the same failure mode as OMNI. We further observe that the requirement to enumerate the entire list of prerequisite skills for FRPs instead of just the most immediate prerequisites for SHARPs results in a higher misspecification rate by the FM. Together, these results show that \Name{}'s unsupervised discovery process produces skills well-aligned with human definitions of meaningful skills to achieve in the environment (\textbf{RQ1}) and that SHARPs' hierarchical composition is central to enabling the agent to master long-horizon skills (\textbf{RQ2}). For Craftax-Classic results without access to the full source code, see \Cref{app:source_code_access}.

\subsection{XLand-MiniGrid Experiments}
\textbf{XLand-MiniGrid} \citep{nikulin2024xland} is a procedurally generated 2D grid-world that challenges \Name{} along two axes absent in Craftax-Classic: stochastic dynamics, where object interaction rules are randomly sampled each episode (yielding up to 36 possible interaction rules for a single object), and a lower level of action abstraction, where rather than executing high-level commands in a single action like in Craftax-Classic, the agent carries one object at a time and must physically navigate to other objects to trigger any interaction. We evaluate on a three-room configuration containing up to 20 objects, including a red and blue door, which first requires the agent to possess the fitting key to access further objects. As XLand-MiniGrid defines no native achievements, we introduce 20 evaluation targets: 18 for picking up each object, 10 of which are not immediately accessible to the agent and require various levels of prerequisite crafting, and 2 for unlocking the red and blue doors.

\textbf{Results}: As shown in \Cref{fig:craftax_XLand-Minigrid_results}c, \Name{} substantially outperforms both baselines, achieving a median success rate of 79.2\% and an average of 61.6\%, compared to 29.7\% (45.4\%) for ELLM and 2.2\% (19.3\%) for OMNI. ELLM again performs well on shallow achievements but fails to chain longer sequences. OMNI performs considerably worse, as its fixed target-state sequences cannot accommodate the stochastic interaction rules that vary across episodes. Similarly, our CODE-FRP ablation achieves 6.1\% (22.8\%), compounded by the FM's tendency to misspecify sub-step sequences under the stochastic rule set. SHARPs efficiently absorb this variation by dispatching to different prerequisite SHARPs depending on the environment state. On the pickup red key achievement, where the active crafting rule is sampled from a pool of 18 possible rules, the agent succeeds in over 37\% of episodes compared to just 7.5\% for ELLM. For the 11\% of achievements matched to mutated SHARPs, mutations yield an average relative performance increase to the base versions of 130.7\%, driven primarily by the addition of prerequisites missed in the original specification. These results further support our observations from Craftax-Classic and strongly affirm \textbf{RQ1} and \textbf{RQ2}.

\subsection{AI2-THOR Experiments}
\begin{wrapfigure}{r}{0.3\columnwidth}
    \centering
    \includegraphics[width=1.0\linewidth]{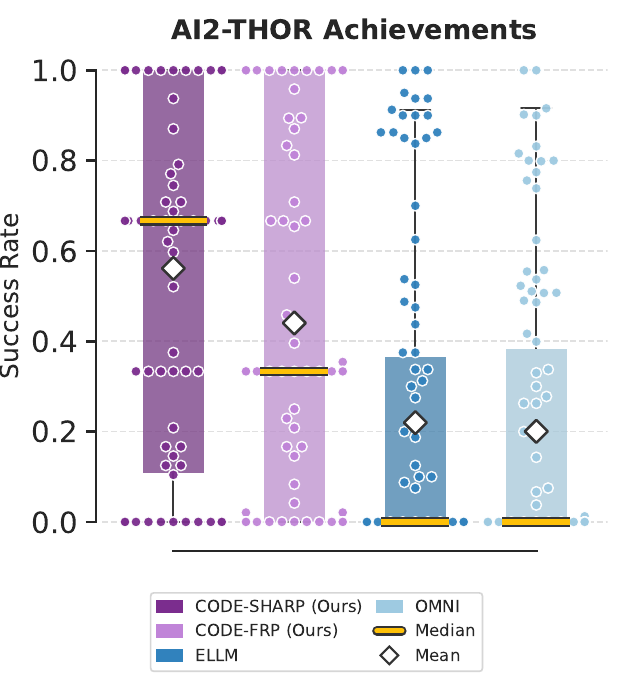}
    \vspace{-1em}
    \caption{Distribution of achievement success rates in AI2-THOR across three runs.}
    \vspace{-1em}
    \label{fig:AI2-THOR}
\end{wrapfigure}
\textbf{AI2-THOR} \citep{kolve2017ai2} is a photorealistic, first-person 3D simulator of household scenes, substantially differing from Craftax's 2D grid-world. We evaluate on the kitchen scene FloorPlan3, featuring diverse interactable utensils, foods, and appliances. For hyperparameters, agent architecture and training setup, we follow the protocol laid out in OMNI (see \Cref{app:hyperparams}). Since AI2-THOR's source code exceeds the FM context, \Name{}, CODE-FRP and OMNI receive only the initial environment state, testing the methods' ability to define meaningful skills without access to the full source code. As AI2-THOR has no native achievements, we introduce 98 hand-authored achievements covering meaningful kitchen tasks, from single-sub-task achievements (depth 1), such as storing utensils, to frying sliced potatoes, requiring six prerequisite sub-tasks. The achievement-depth distribution and our protocol for defining meaningful achievements are provided in \Cref{app:thor_achievements}.

\textbf{Results}: \Name{} achieves 66.7\% median success (56.1\% mean), compared with 0.0\% (22.0\%) for ELLM, 0.0\% (20.1\%) for OMNI, and 33.3\% (44.0\%) for CODE-FRP (\Cref{fig:AI2-THOR}). Breaking down the performance on achievements of increasing depths, we find that ELLM and OMNI achieve just 0.9\% and 2.9\% success rate on achievements of depth 3, with 0.0\% for any achievements requiring 4+ sub-tasks. In contrast, \Name{} achieves 34.2\% on tasks of depth 3 and 6.9\% for depth of 4+, compared to 13.0\% and 0.0\% for CODE-FRP. Notably, \Name{} is the only method to complete the depth 6 achievement \texttt{CookSlicedPotato} in 16.7\% of episodes. These results provide further evidence to affirm \textbf{RQ1} and \textbf{RQ2}.
\begin{figure*}[t]
    \centering
    \includegraphics[width=1.0\textwidth]{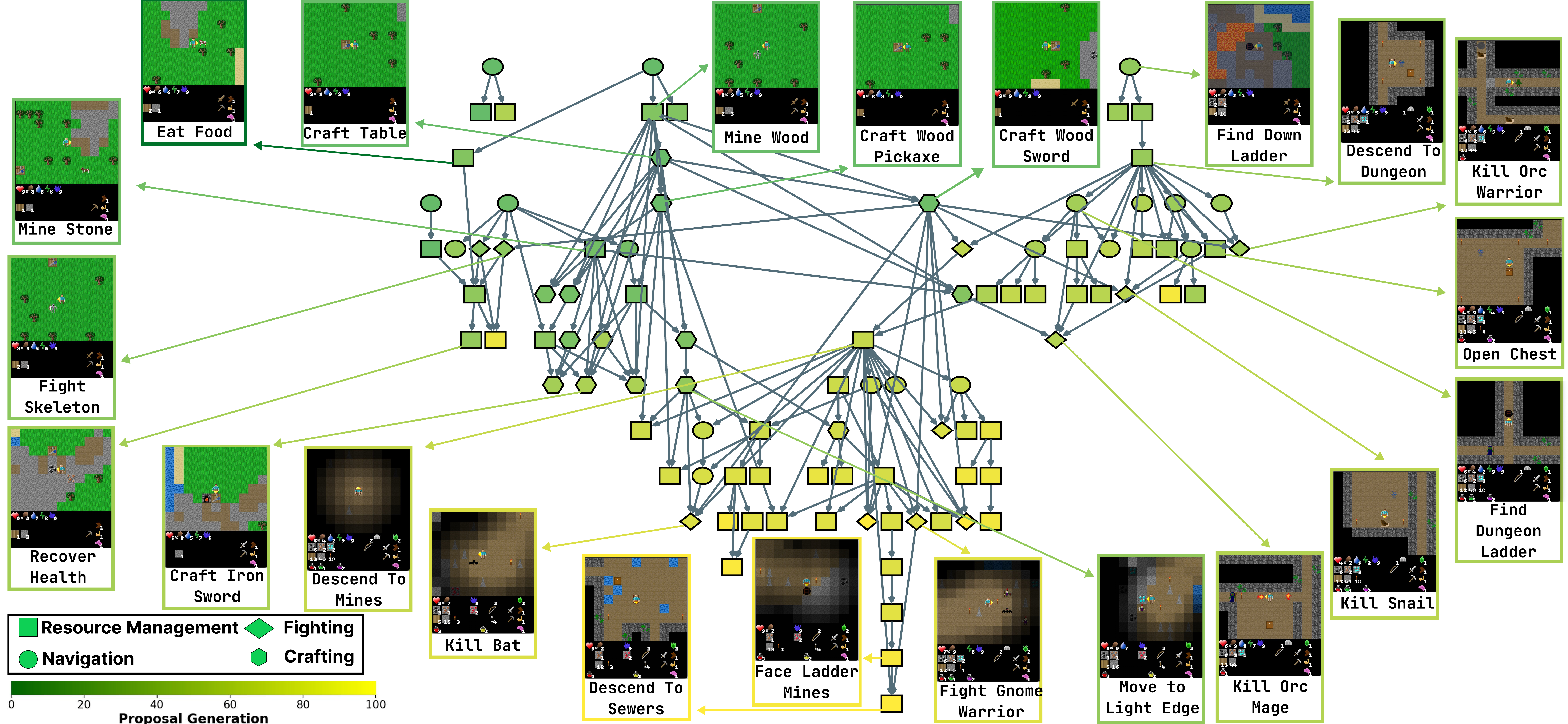}
    \caption{\Name{} continuously builds on existing SHARPs in the archive to define novel, meaningful skills mirroring the natural skill curriculum of Craftax.}
    \label{fig:found_archive}
\end{figure*}
\section{Long Run Experiment}
\label{sec:experiments}
To rigorously test \Name{}'s capacity for open-ended skill discovery, we scale our evaluation to Craftax-Extended \citep{matthews2024craftax}, which augments Craftax-Classic with NetHack's \citep{küttler2020nethacklearningenvironment} dungeon-crawling dynamics, adding 8 new levels with diverse entities and mechanics. We aim to address one main research question, \textbf{RQ3}: Does \Name{} sustain open-ended discovery at scale, producing an archive of increasingly complex SHARPs that broadly covers the environment's skill space and trains a generalist agent to learn increasingly complex skills?

We run \Name{} for 100 proposal and 85 evolution iterations, training the agent for $2\times10^9$ environment steps, requiring roughly 3 days. To better disentangle \Name{}'s skill discovery process from limitations posed by outside factors,  we enforce a minimum health threshold to prevent premature episode termination. To evaluate both the breadth of the discovered archive and the quality of the skills within it, we employ an FM-based policy planner that guides the agent through a set of challenging benchmark tasks,  each consisting of up to 11 sub-tasks, via policies-in-code \citep{klissarov2024maestromotifskilldesignartificial}. For the planner to be able to generate effective policies, \Name{} must discover SHARPs that broadly cover the environment skill space and define them such that the agent learns to reliably execute their intended behaviours. Consequently, benchmark performance serves as a joint proxy for archive breadth and skill quality. Detailed descriptions of the four benchmark tasks are provided in \Cref{app:benchmark_scenarios}. 

Without any training on the original environment or benchmark rewards, we compare the zero-shot performance of the agent trained exclusively by \Name{} against: 1) An agent pretrained on the original Craftax-Extended rewards and finetuned for each benchmark separately; 2) A task-specific agent trained only on the benchmarks; 3) A zero-shot baseline in the form of an LLM agent based on the ReAct \citep{yao2022react} framework; 4) Our CODE-FRP ablation presented in \Cref{sec:comp}, utilising the same policies-in-code as for \Name{}. All applicable baselines are trained for $2\times10^9$ steps, with an additional $1\times10^8$ for the fine-tuned agent, and evaluated with minimum health threshold enforcement. Scores are reported as the geometric mean over sub-task success rates following the Craftax scoring paradigm. For \Name{} and CODE-FRP, results are over three runs $\times$ three policies-in-code, and nine seeds for all other methods. Error bars indicate the bootstrapped 95\% CI. See \Cref{app:detailed_perf} for detailed results.

\subsection{Long Run Results}
\label{subsec:discovery_analysis}
\textbf{Results}: \Name{} discovers an average of 90 SHARPs, forming a structured curriculum that closely mirrors the natural progression of Craftax-Extended. The archive evolves from foundational \textit{Overworld} skills to increasingly complex skills spanning the \textit{Dungeon}, \textit{Mines} and \textit{Sewers} levels, which directly build on the growing hierarchical archive (\Cref{fig:found_archive}). We find that this progress is often driven by \Name{} directly exploiting environment mechanics to define unexpected and innovative behaviours. For example, in the dark \textit{Mines}, \Name{} discovers the \texttt{MoveToEdgeOfLightLevel2} SHARP, rewarding the agent for moving to the edge of a lit area. \texttt{PlaceTorchAtEdgeLevel2} directly builds on this, rewarding the agent for first moving to the light edge, then placing a torch to extend the lit area's boundary. \Name{} discovers that by using this SHARP as a prerequisite, the agent can be implicitly induced to perform exploration as it constantly alternates between moving to the edge of the light circle and placing down a new torch.
\begin{figure*}[t]
    \centering
    \includegraphics[width=0.9\textwidth]{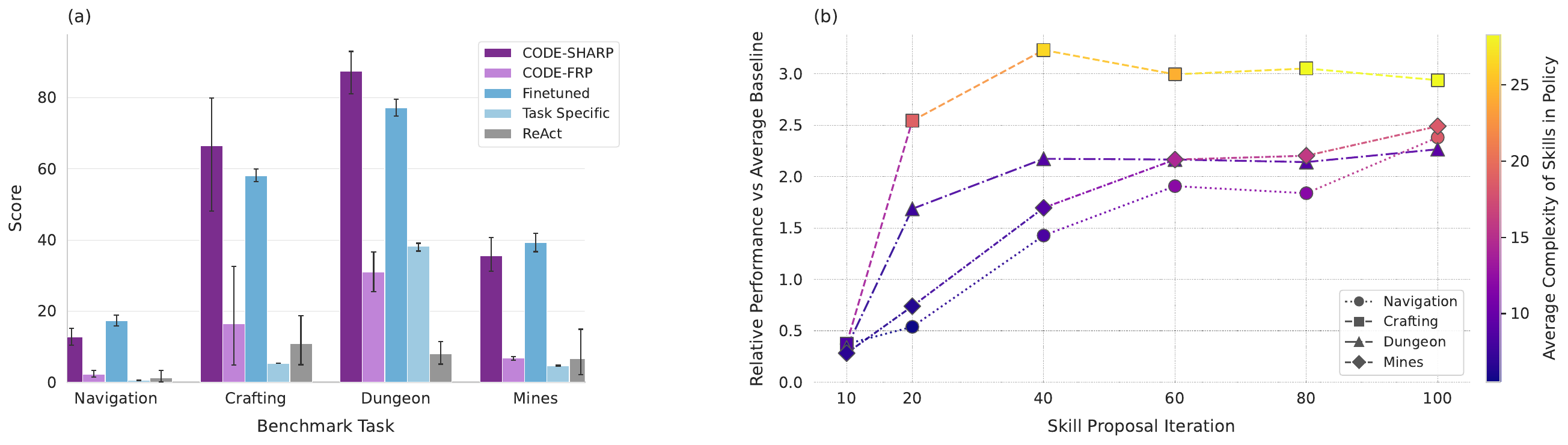}
    \caption{\textbf{(a)} Score achieved by \Name{} across benchmark tasks. \textbf{(b)} The policy planner utilises increasingly complex SHARPs to define policies-in-code throughout training, resulting in large performance gains relative to the average baseline.}
    \label{fig:performance}
\end{figure*}

On the four benchmark tasks the agent trained exclusively by \Name{} and following the policies-in-code achieves an overall benchmark score of 50.5 (\Cref{fig:performance}a). This matches the finetuned agent (47.9) and significantly outperforms all other baselines, demonstrating that the SHARPs discovered by \Name{} extensively cover the skill space of Craftax-Extended and faithfully train the generalist agent in their intended behaviours. Specifically, \Name{} demonstrates superior performance in Crafting (66.5 vs. 58.1) and Dungeon (87.4 vs. 77.1); conversely, the fine-tuned agent retains an advantage only in Navigation (12.7 vs. 17.3) and Mines (35.6 vs. 39.2), likely benefiting from ground-truth environment rewards that explicitly incentivise deeper exploration. Further illustrating the complexity of these tasks, the task-specific agents demonstrate markedly degraded performance, recording scores of 0.5 and 4.7 on Navigation and Mines respectively, with an overall average of 12.2. Similarly, the ReAct agent performs poorly across benchmark tasks (6.7), struggling with structured embodied exploration despite having source code access. These results highlight that FMs can be more effectively utilised as high-level skill discovery orchestrators via \Name{} rather than directly as embodied agents. Finally, the CODE-FRP ablation (14.1) confirms the SHARP formulation as the primary driver of benchmark performance rather than the planning provided by the policies-in-code.

We additionally analyse how the performance of the agent evolves as the archive grows in \Cref{fig:performance}b by restricting the policy planner to only use SHARPs for the policies-in-code discovered by specific iterations of the discovery process. We find that downstream scores consistently improve as the archive expands, following a natural curriculum: Benchmarks focused on earlier levels (Crafting and Dungeon) improve early on, while benchmarks focused on later levels (Navigation and Mines) see gains later. Crucially, these performance leaps correlate directly with increases in the average complexity\footnote{We define the complexity of a SHARP as one plus the sum of the complexities of its assigned prerequisite SHARPs:
$C(\sigma_k) = 1 + \sum_{(c_i,u_i)\in\psi_{\sigma_k}} C(u_i)$.} of the SHARPs utilised by the policy planner, confirming that \Name{} open-endedly discovers increasingly complex SHARPs training the agent to continuously expand its set of learned abilities across 100 iterations and producing an archive that broadly covers the skill space of Craftax-Extended (\textbf{RQ3}).

\textbf{SHARP Mutation Analysis}: \Name{} continuously improves existing SHARPs through FM-based mutation, producing an average of $20$ elite SHARP versions per run. These mutations yield a $68.8\%$ relative performance gain, increasing absolute success rates of the agent from $24.3\%$ to $41.0\%$ across all mutated SHARPs. We observe that successful mutations primarily resolve structural inefficiencies such as suboptimal condition ordering and missing or poorly chosen prerequisites. For instance, the base \texttt{KillOrcWarrior} SHARP incorrectly required descending to the \textit{Dungeon} before crafting a stone sword, whereas the refined version reorders these steps, significantly improving success rate.

\section{Related Work}
\label{sec:related_works}
\textbf{Foundation Models for Open-Ended Skill Discovery}: The ability to acquire novel skills and reflect on one's own mastered abilities is central to general intelligence \citep{hughes2024openendednessessentialartificialsuperhuman, jiang2023general, wang2020enhancedpoetopenendedreinforcement, adaptiveagentteam2023humantimescaleadaptationopenendedtask}. With the recent rise in FMs' reasoning abilities and semantic world knowledge, multiple works have proposed utilising FMs to drive open-ended autonomous skill discovery and learning. One prominent class of methods deploys FMs directly as open-ended agents \citep{yao2022react, wang2023describe, bolton2025sima}. While these approaches are very general, they typically suffer from high computational costs and latency due to the need for frequent FM inference at each decision step. To minimise computational costs, alternative frameworks utilise FMs as high-level planners, relying on hand-engineered frameworks \citep{wang2023voyager} or pre-trained models \citep{wang2024jarvis, ahn2022can, zhang2023bootstrap} to handle low-level action execution. Methods like Eureka \citep{maeureka}, ARM-FM \citep{castanyer2025arm} or SCALAR \citep{zabounidis2026scalar} generate and optimise reward functions for externally specified objectives. Unlike \Name{}, which autonomously discovers which skills to learn in the absence of a predefined goal, these methods instead address how to train an agent to efficiently learn a prespecified task. In this setting, FMs can instead propose diverse goals and provide rewards to train an agent from scratch \citep{zhangomni, du2023guiding, pourcel2024autotelic, faldor2024omni}.

\textbf{Foundation Models for Hierarchical Learning}: Decomposing long-horizon tasks into temporally extended abstractions is a fundamental challenge in AI, historically addressed through Hierarchical Task Networks (HTNs) \citep{ghallab2004automated, erol1994htn} and the Options framework in reinforcement learning \citep{sutton1999between, bacon2017option, klissarov2021flexible}. Recent neuro-symbolic approaches leverage the reasoning capabilities of FMs to automate task decomposition. These methods translate high-level instructions into sequences of sub-goals, guiding embodied agents either through executable code policies \citep{ahn2022can, liang2022code, prakash2023llm, wang2023voyager, klissarov2024maestromotifskilldesignartificial} or via auxiliary rewards and completion signals \citep{venuto2024code, castanyer2025arm, nottingham2023embodiedagentsdreampixelated, zhangomni, zhang2023bootstrap}. Most closely related to our work, SCALAR \citep{zabounidis2026scalar} also trains agents using prerequisite-linked skills, but plans backwards from a supplied goal and selects and refines skills within a fixed, human-specified skill space. In contrast, \Name{} discovers new skills without any external goals and recursively re-evaluates the active prerequisite at every environment step.

\section{Conclusion}
\label{conclusion}
We introduced \Name{}, a framework that leverages FMs to autonomously and open-endedly grow and evolve an archive of Skills as Hierarchical Reward Programs (SHARPs) to train a generalist goal-conditioned agent via reinforcement learning from scratch. Each SHARP is an FM-generated Python program that encodes a success condition and a set of prerequisite dependencies over previously discovered SHARPs, using a recursive transition operator to dynamically route the agent through its prerequisite chain at runtime, enabling efficient learning of long-horizon tasks without any hand-crafted rewards, APIs, or curated demonstrations. Empirically, we demonstrate that \Name{}'s fully autonomous open-ended discovery process produces a continuously growing, structured skill archive that spans increasingly complex behaviours, successfully training agents to master increasingly long-horizon skills, substantially outperforming prior works and matching agents trained on ground truth rewards. While \Name{} reduces the human engineering required in the loop compared to related works, it still requires structured information about the environment, such as its source code or an initial state representation. Additionally, our ablations show that performance with limited access to environment information is heavily dependent on the environment priors obtained by the FMs during pre-training. Overall, we believe \Name{} presents a meaningful next step toward fully autonomous open-ended skill discovery and training of generalist agents.

\section*{Acknowledgements}
This work was supported by the Engineering and Physical Sciences Research Council [grant number UKRI5149] and other sponsors of BOLD. RB is supported by a PhD Scholarship from Sony Interactive Entertainment.

\bibliographystyle{plainnat}
\bibliography{example_paper}

\newpage
\appendix
\crefalias{section}{appendix}
\section{Algorithm}
\label{app:algorithm}

\begin{algorithm}[H]
\caption{CODE-SHARP: Continuous Open-ended Discovery \& Evolution}
\label{alg:code_sharp}
\begin{algorithmic}[1]
\State \textbf{Input:} Maximum iterations $I$, initial archive
    $\Lambda_0 = (\mathcal{V}_0,\mathcal{E}_0)$,
    initial policy $\pi_\theta$, environment context,
    learning-progress threshold $\tau = 0.05$
\State \textbf{Initialize:} Failed proposals
    $\mathcal{H} \gets \emptyset$,
    success rates $\rho \gets \text{Init}(\Lambda_0)$

\For{$t = 1$ \textbf{to} $I$}
    \State $\Lambda_t \gets \Lambda_{t-1}$
    \State $\mathcal{C} \gets
        (\Lambda_t,\mathcal{H},\text{environment context})$
    \Comment{base context supplied to every FM call}

    \State Sample $n$ proposal categories uniformly with replacement
    \State $\mathcal{P}_{raw} \gets
        \text{ProposalGenerator}(\mathcal{C},\text{categories})$
    \State $\mathcal{P}_{impl} \gets
        \text{ProposalImplementor}(\mathcal{P}_{raw},\mathcal{C})$
    \State $\mathcal{S}_{cand} \gets
        \text{ProposalJudge}(\mathcal{P}_{impl},\mathcal{C})$
    \Comment{$|\mathcal{S}_{cand}| \leq 2$}

    \State $(\mathcal{S}_{cand},\mathcal{F}) \gets
        \text{CompileAndRepair}(\mathcal{S}_{cand},\mathcal{C})$
    \Comment{up to three implementor repair attempts}
    \State $\mathcal{H} \gets \mathcal{H} \cup \mathcal{F}$

    \If{$\mathcal{S}_{cand} \neq \emptyset$}
        \State $\pi_{copy} \gets \text{Copy}(\pi_\theta)$
        \State $\Lambda_{eval} \gets
            \text{AddSHARPs}(\Lambda_t,\mathcal{S}_{cand})$
        \State $(\boldsymbol{\rho}^{init},
                 \boldsymbol{\rho}^{final}) \gets
            \text{EvaluateLearnability}(\pi_{copy},\Lambda_{eval})$
        \Comment{train the copy on the augmented archive}

        \For{$\sigma_{new} \in \mathcal{S}_{cand}$}
            \State $\Delta\rho_{new} \gets
                \rho^{final}_{new} - \rho^{init}_{new}$
            \If{$\Delta\rho_{new} > \tau$}
                \State $\Lambda_t \gets
                    \text{AddSHARPs}(\Lambda_t,\{\sigma_{new}\})$
                \State $\rho(\sigma_{new}) \gets \rho^{final}_{new}$
            \Else
                \State $\mathcal{H} \gets
                    \mathcal{H} \cup \{\sigma_{new}\}$
            \EndIf
        \EndFor
    \EndIf

    \State Sample $\sigma_k \in \mathcal{V}_t$
        with probability $P(k) \propto 1-\rho_k$
    \State $\mathcal{C} \gets
        (\Lambda_t,\mathcal{H},\text{environment context})$
    \State $\{\sigma_k^{(1)},\ldots,\sigma_k^{(m)}\} \gets
        \text{MutationPipeline}(\sigma_k,\mathcal{C})$
    \Comment{generate and implement heuristic-conditioned mutations}

    \State $\rho_{mut}^{(i)} \gets
        \text{ZeroShotEval}(\pi_\theta,\Lambda_t,
                           \sigma_k \leftarrow \sigma_k^{(i)})$
        for $i=1,\ldots,m$ in parallel
    \Comment{no policy updates}
    \State $i^* \gets
        \arg\max_{i \in \{1,\ldots,m\}} \rho_{mut}^{(i)}$
    \If{$\rho_{mut}^{(i^*)} > \rho_k$}
        \State $\Lambda_t \gets
            \text{ReplaceSHARP}(\Lambda_t,\sigma_k,\sigma_k^{(i^*)})$
        \State $\rho(\sigma_k) \gets \rho_{mut}^{(i^*)}$
    \EndIf

    \State $\pi_\theta,\rho \gets
        \text{TrainEpoch}(\pi_\theta,\Lambda_t,\rho)$
\EndFor
\State \textbf{Return} $\Lambda_E,\pi_\theta$
\end{algorithmic}
\end{algorithm}

\begin{algorithm}[H]
\caption{Agent Training and SHARP Traversal}
\label{alg:sharp_training}
\begin{algorithmic}[1]

\Function{TrainEpoch}{$\pi_\theta,\Lambda,\rho$}
    \State Reset environment and observe $s_0$
    \State $\sigma_{\text{target}} \sim
        \operatorname{Uniform}(\mathcal{V})$
    \State $t_{\text{last}} \gets 0$
    \State Initialise rollout buffer $\mathcal{D} \gets \emptyset$

    \For{$t=0,\ldots,L-1$}
        \Comment{collect $L$ environment transitions}
        \State $\sigma_{\text{active}} \gets
            \mathcal{T}(\sigma_{\text{target}},s_t)$
        \State $z_t \gets
            \operatorname{Embed}(\operatorname{Name}(
                \sigma_{\text{active}}))$
        \State $a_t \sim
            \pi_\theta(\cdot \mid s_t,z_t)$
        \State $s_{t+1},\text{done} \gets \text{Env}(a_t)$

        \State $\alpha(\sigma_{\text{active}}) \gets
            \min\left(
                \frac{1}{\rho(\sigma_{\text{active}})+\epsilon},
                10.0
            \right)$
        \State $r_t \gets
            \alpha(\sigma_{\text{active}})
            \cdot \mathbb{I}[
                \phi_{\sigma_{\text{active}}}(s_{t+1},s_t)=1]$

        \State Store $(s_t,z_t,a_t,r_t,s_{t+1},\text{done})$
            in $\mathcal{D}$
        \State Record skill attempts and successes
            for success-rate estimation

        \If{$\phi_{\sigma_{\text{active}}}(s_{t+1},s_t)=1$}
            \State $t_{\text{last}} \gets t+1$
            \Comment{reset stall timer on progress}
        \EndIf

        \If{$\text{done}$}
        \State Reset environment and overwrite $s_{t+1}$
            with the initial state
        \State $\sigma_{\text{target}} \sim
            \operatorname{Uniform}(\mathcal{V})$
        \State $t_{\text{last}} \gets t+1$
        \ElsIf{$\phi_{\sigma_{\text{target}}}(s_{t+1},s_t)=1$
            \textbf{or} $(t+1-t_{\text{last}})\geq 200$}
            \State $\sigma_{\text{target}} \gets
                \text{SampleTarget}(s_{t+1},\Lambda,\rho)$
            \State $t_{\text{last}} \gets t+1$
        \EndIf
    \EndFor

    \State Update $\pi_\theta$ via PPO using $\mathcal{D}$
    \State Update $\rho$ from recorded skill attempts and successes
    \State \Return $\pi_\theta,\rho$
\EndFunction

\Statex

\Function{$\mathcal{T}$}{$\sigma,s$}
    \For{$(c_i,u_i) \in \psi_\sigma$ in order}
        \If{$c_i(s)=0$}
            \State \Return $\mathcal{T}(u_i,s)$
        \EndIf
    \EndFor
    \State \Return $\sigma$
\EndFunction

\Statex

\Function{SampleTarget}{$s,\Lambda,\rho$}
    \For{each $\sigma_j \in \mathcal{V}$}
        \State $N_{jk}(s) \gets
            \mathbb{I}[\exists(c_i,u_i)\in\psi_{\sigma_j}:
                u_i=\sigma_k \land c_i(s)=1]$
            for each $\sigma_k \in \mathcal{V}$
        \State $B_j(s) \gets
            \displaystyle\prod_{k=1}^{|\mathcal{V}|}
                (\rho_k+\epsilon)^{-N_{jk}(s)}$
        \State $m_j(s) \gets
            \mathbb{I}[\sigma_j
                \text{ is not already satisfied in }s]$
    \EndFor
    \State $p_j \gets
        \displaystyle\frac{B_j(s)m_j(s)}
            {\sum_\ell B_\ell(s)m_\ell(s)}$
    \State \Return
        $\sigma_{\text{target}}
            \sim \operatorname{Categorical}(\mathbf{p})$
\EndFunction

\end{algorithmic}
\end{algorithm}

\section{Description of Benchmark Scenarios}
\label{app:benchmark_scenarios}
The Craftax environment presents a vast and ostensibly open-ended skill space that remains challenging even for expert human players. Consequently, our benchmark suite is designed to probe the diverse spectrum of capabilities required within the environment, specifically targeting the first four levels, which, to our knowledge, are the only levels that have been reached by any algorithm trained on Craftax. Drawing on the evaluation methodology of MaestroMotif \citep{klissarov2024maestromotifskilldesignartificial} we defined four distinct tasks targeting essential high-level competencies: \textit{Navigation}, \textit{Crafting}, and two composite scenarios, \textit{Dungeon} and \textit{Mines}, which require the integration of multiple distinct competencies.

Each benchmark is implemented as a sequence of milestones that must be achieved in order. These tasks are intentionally designed as ambitious, long-horizon objectives intended to quantify incremental progress toward complex goals, rather than as binary pass/fail assessments. Detailed specifications for each benchmark task are provided below.

\textbf{Navigation} The navigation benchmark is comprised of $6$ milestones which follow the natural progression through the levels of Craftax. First, the agent must locate the down ladder on the \textit{Overworld} level to descend to the \textit{Orcish Dungeons}. Once on \textit{Orcish Dungeons}, the agent must find and eliminate $8$ hostile orc mobs to be able to unlock the next level, before exploring the level to locate the ladder down to the \textit{Gnomish Mines}. Here, the agent must eliminate $8$ hostile gnome mobs to unlock the \textit{Sewers} level. The \textit{Gnomish Mines} is initially fully dark, prohibiting any form of exploration unless the agent places down torches to light up a limited perimeter. Resources to craft torches can only be found on the preceding levels, requiring the agent to effectively plan ahead. After the agent has located the down ladder, the agent should descend to the \textit{Sewers} level. On the \textit{Sewers} level, paths are often blocked by water, requiring the agent to place down blocks to displace it and clear a path. The benchmark terminates once the agent has located an enchantment table, an important in-game item which can enchant weapons and armour with certain effects.

\textbf{Crafting} The crafting benchmark focuses on $5$ milestones related to crafting advanced iron tools. While diamond tools represent the most advanced tool class in Craftax they require diamonds, which are not guaranteed to spawn in the overworld. To be able to evaluate an agent's crafting ability without it being constrained by its ability to descend to subsequent levels, we therefore focus on the iron tools and armour. The first milestone requires the agent to craft a \texttt{WoodPickaxe}, which requires a total of $3$ blocks of wood. For the next milestone, the agent needs to craft a \texttt{StonePickaxe}, which requires the previously crafted pickaxe to mine stone and another block of wood. Next, the agent should craft an \texttt{IronPickaxe}, which requires a furnace adjacent to a crafting table, a block of coal and the \texttt{StonePickaxe} to mine a block of iron. Next, the agent must craft an \texttt{IronSword}, which again requires the same resources as the \texttt{IronPickaxe}. Finally, the benchmark terminates when the agent crafts a set of \texttt{IronArmour}, requiring a furnace adjacent to a crafting table as well as $3$ blocks of coal and $3$ blocks of iron. To solve this benchmark, the agent must repeatedly explore the environment to obtain the required resources and complete multi-step crafting processes.

\textbf{Dungeon} The dungeon benchmark focuses on $11$ milestones across the \textit{Overworld} and \textit{Orcish Dungeon} levels of Craftax, simulating the process of the agent first having to gather helpful resources and tools before descending to subsequent levels. The agent must first craft a \texttt{WoodPickaxe} before crafting a \texttt{StonePickaxe} and a \texttt{StoneSword}. After these essential tools, the agent should continue by crafting \texttt{Torches} and \texttt{Arrows} before heading to the grassland and collecting a \texttt{Sapling}. Next, the agent must descend to the \textit{Dungeon} level, where it should locate and open loot chests until a \texttt{Potion} is dropped. Next, the agent should eliminate at least $2$ hostile orcish mobs before locating the upwards ladder and ascending back to the \textit{Overworld}. Finally, the benchmark terminates when the agent has replenished its hunger via eating fruit or killing a cow. To solve this benchmark, the agent must complete a large number of sub-goals for each milestone, making it extremely challenging to fully complete.

\textbf{Mines} The mines benchmark focuses on the \textit{Gnomish Mines} level, an area characterised by a higher spawn rate for rare resources such as diamonds. This benchmark evaluates the agent's ability to descend through the environment and locate a diamond via $10$ sequential milestones. Similar to the \textit{Dungeon} benchmark, the agent must first craft a \texttt{WoodPickaxe}, \texttt{StonePickaxe}, and \texttt{Torches} in the \textit{Overworld} before locating the ladder to the \textit{Orcish Dungeon}. After descending, the agent is required to eliminate $8$ hostile orc mobs to unlock the passage to the \textit{Gnomish Mines}. Upon reaching the level, the agent must demonstrate survival proficiency by placing a torch and hunting a bat for food. Next the agent must navigate the \textit{Gnomish Mines} level, ideally illuminating its way via the crafted torches, to locate a water source and replenish its thirst. This benchmark is extremely challenging to solve, as it requires strong exploration skills to first descend to the \textit{Gnomish Mines} and understanding of the game mechanics to deal with the darkness before attempting to locate a diamond block comprising approximately $0.5\%$ of blocks on the \textit{Gnomish Mines} level.

\section{AI2-THOR Achievement Creation}
\label{app:thor_achievements}
AI2-THOR defines no native achievements marking the completion of meaningful tasks. To evaluate all methods on the same footing, we define a fixed set of 98 ground-truth achievements for the FloorPlan3 kitchen, following the achievement-based evaluation of Craftax \citep{matthews2024craftax}. The set was fixed before any discovery run and was not modified afterwards. The achievements are organised by interaction type. Single-step achievements (picking up, opening, switching on and slicing an object) are listed exhaustively: every object type whose simulator affordance flag allows the interaction receives one. Multi-step achievements combine an object with a destination or process: storage (placing an object in or on a receptacle), filling a container with water or coffee, cooking, and a fill-then-heat chain. Only object–destination pairings that correspond to typical kitchen use are included, such as utensils in drawers, perishable food in the fridge, cookware on the stove and dishware in the sink. Each achievement is expressed as a condition on the simulator's object metadata that must hold for a single object instance, and all attribute values were verified against the running simulator. An achievement counts only when its condition changes from false to true within an episode, so we exclude any achievement already satisfied in the initial scene. We assign each achievement a depth, defined as the minimum number of interaction actions needed to reach it from the initial state, excluding navigation. Success is determined only by these conditions, independently of any skill's own success function. Each achievement is evaluated over 16 episodes per run.

\section{Performance Breakdown Comparative Evaluation Experiments}
\label{app:detailed_comparison}

\begin{table}[H]
\centering
\caption{Per-achievement mean success rate on Craftax Achievements.}
\label{tab:craftax_achievements}
\begin{tabular}{lcccc}
\toprule
Achievement & CODE-SHARP (Ours) & CODE-FRP (Ours) & ELLM & OMNI \\
\midrule
Place Table & \textbf{1.000} & 0.998 & 0.971 & 0.989 \\
Collect Sapling & \textbf{0.999} & 0.998 & 0.994 & 0.044 \\
Make Wood Pickaxe & \textbf{0.999} & 0.994 & 0.727 & 0.779 \\
Make Wood Sword & \textbf{0.999} & 0.800 & 0.230 & 0.965 \\
Place Furnace & \textbf{0.998} & 0.575 & 0.011 & 0.273 \\
Collect Stone & \textbf{0.998} & 0.200 & 0.097 & 0.699 \\
Make Stone Sword & \textbf{0.998} & 0.380 & 0.001 & 0.790 \\
Make Stone Pickaxe & \textbf{0.997} & 0.532 & 0.000 & 0.779 \\
Collect Coal & \textbf{0.990} & 0.782 & 0.008 & 0.579 \\
Collect Iron & \textbf{0.981} & 0.181 & 0.000 & 0.000 \\
Make Iron Sword & \textbf{0.949} & 0.168 & 0.000 & 0.000 \\
Make Iron Pickaxe & \textbf{0.947} & 0.165 & 0.000 & 0.000 \\
Collect Wood & 0.892 & 1.000 & \textbf{1.000} & 0.997 \\
Eat Cow & 0.657 & 0.571 & \textbf{0.941} & 0.859 \\
Collect Drink & 0.605 & \textbf{0.934} & 0.892 & 0.004 \\
Place Plant & 0.398 & 0.398 & \textbf{0.993} & 0.004 \\
Collect Diamond & \textbf{0.160} & 0.001 & 0.000 & 0.000 \\
Defeat Zombie & 0.135 & 0.012 & \textbf{0.973} & 0.863 \\
Defeat Skeleton & \textbf{0.074} & 0.016 & 0.000 & 0.006 \\
Eat Plant & 0.000 & 0.000 & 0.000 & 0.000 \\
Place Stone & 0.000 & 0.000 & \textbf{0.065} & 0.000 \\
Wake Up & 0.000 & 0.008 & \textbf{0.850} & 0.003 \\
\midrule
\textbf{Median} & \textbf{0.948} & 0.389 & 0.081 & 0.158 \\
\textbf{Average} & \textbf{0.672} & 0.442 & 0.398 & 0.392 \\
\bottomrule
\end{tabular}
\end{table}

\begin{table}[H]
\centering
\caption{Per-achievement mean success rate on XLand-MiniGrid achievements.}
\label{tab:XLand-Minigrid_achievements}
\begin{tabular}{lcccc}
\toprule
Achievement & CODE-SHARP (Ours) & CODE-FRP (Ours) & ELLM & OMNI \\
\midrule
Pick Up Square Red & \textbf{0.978} & 0.683 & 0.908 & 0.493 \\
Pick Up Ball Green & \textbf{0.969} & 0.479 & 0.877 & 0.552 \\
Pick Up Ball Blue & \textbf{0.962} & 0.576 & 0.885 & 0.396 \\
Pick Up Square Purple & \textbf{0.959} & 0.346 & 0.848 & 0.163 \\
Pick Up Square Green & \textbf{0.953} & 0.446 & 0.864 & 0.408 \\
Pick Up Ball Red & \textbf{0.938} & 0.454 & 0.889 & 0.636 \\
Pick Up Square Blue & \textbf{0.937} & 0.379 & 0.918 & 0.309 \\
Pick Up Ball Purple & \textbf{0.931} & 0.618 & 0.824 & 0.210 \\
Pick Up Hex Red & \textbf{0.923} & 0.431 & 0.923 & 0.621 \\
Pick Up Pyramid Green & \textbf{0.805} & 0.082 & 0.270 & 0.021 \\
Pick Up Hex Purple & \textbf{0.779} & 0.004 & 0.191 & 0.006 \\
Pick Up Key Blue & \textbf{0.655} & 0.040 & 0.306 & 0.016 \\
Pick Up Pyramid Red & \textbf{0.626} & 0.015 & 0.289 & 0.024 \\
Pick Up Key Red & \textbf{0.375} & 0.006 & 0.075 & 0.001 \\
Pick Up Star Green & \textbf{0.250} & 0.000 & 0.007 & 0.000 \\
Unlock Blue Door & \textbf{0.131} & 0.002 & 0.000 & 0.000 \\
Pick Up Star Blue & \textbf{0.120} & 0.000 & 0.000 & 0.000 \\
Unlock Red Door & \textbf{0.037} & 0.000 & 0.000 & 0.000 \\
Pick Up Star Red & \textbf{0.001} & 0.000 & 0.000 & 0.000 \\
Pick Up Pyramid Blue & 0.000 & 0.000 & 0.000 & 0.000 \\
\midrule
\textbf{Median} & \textbf{0.792} & 0.061 & 0.297 & 0.022 \\
\textbf{Average} & \textbf{0.616} & 0.228 & 0.454 & 0.193 \\
\bottomrule
\end{tabular}
\end{table}

\begin{table}[H]
\centering
\caption{Per-sub-task-depth achievement success rate by method on AI2-THOR.}
\label{tab:thor_depth_breakdown}
\begin{tabular}{lrrrrr}
\toprule
Depth & \# Ach. & CODE-SHARP (OURS) & CODE-FRP (OURS) & ELLM & OMNI \\
\midrule
1       & 32 & \textbf{0.903} & 0.893 & 0.611 & 0.342 \\
2       & 37 & \textbf{0.485} & 0.314 & 0.048 & 0.218 \\
3       & 23 & \textbf{0.342} & 0.130 & 0.009 & 0.029 \\
4       &  1 & \textbf{0.125} & 0.000 & 0.000 & 0.000 \\
5       &  3 & 0.000 & 0.000 & 0.000 & 0.000 \\
6       &  2 & \textbf{0.083} & 0.000 & 0.000 & 0.000 \\
\midrule
Overall & 98 & \textbf{0.561} & 0.440 & 0.220 & 0.201 \\
\bottomrule
\end{tabular}
\end{table}

\section{Performance Breakdown Long Run Experiments}
\label{app:detailed_perf}
We present the average success rate over the individual milestones in each benchmark. The overall score is computed following the Craftax scoring paradigm, which computes the geometric mean over success rate to emphasise small improvements on hard-to-complete milestones.

\begin{table}[h]
\centering
\small
\caption{Per-milestone success rates (\%) on the Crafting benchmark. Values are mean (95\% CI), computed via bootstrap over 2000 resamples.}
\label{tab:milestones_crafting}
\resizebox{\textwidth}{!}{%
\begin{tabular}{lccccc}
\toprule
Milestone & CODE-SHARP & CODE-FRP & Finetuned & task-specific & ReAct \\
\midrule
Craft wood pickaxe & 92.6 (80.9, 98.8) & 79.0 (54.6, 96.6) & 98.7 (98.3, 99.0) & 99.0 (98.8, 99.2) & 100.0 (100.0, 100.0) \\
Craft stone pickaxe & 87.5 (65.9, 98.6) & 63.5 (38.7, 85.4) & 96.0 (95.3, 96.9) & 97.7 (96.9, 98.4) & 88.9 (66.7, 100.0) \\
Craft iron pickaxe & 75.8 (51.7, 93.0) & 17.6 (1.7, 39.6) & 84.8 (83.5, 86.1) & 0.0 (0.0, 0.0) & 33.3 (0.0, 66.7) \\
Craft iron sword & 69.7 (48.5, 86.4) & 11.7 (0.0, 28.7) & 65.9 (64.1, 67.9) & 0.0 (0.0, 0.0) & 11.1 (0.0, 33.3) \\
Craft iron armour & 34.9 (21.2, 48.6) & 3.4 (0.0, 8.5) & 12.2 (10.6, 14.0) & 0.0 (0.0, 0.0) & 0.0 (0.0, 0.0) \\
\midrule
\textbf{Overall Score} & 66.52 (48.17, 79.96) & 16.43 (4.77, 29.09) & 58.08 (56.33, 60.00) & 5.29 (5.28, 5.31) & 10.80 (4.91, 18.73) \\
\bottomrule
\end{tabular}%
}
\end{table}

\begin{table}[h]
\centering
\small
\caption{Per-milestone success rates (\%) on the Mines benchmark. Values are mean (95\% CI), computed via bootstrap 2000 resamples.}
\label{tab:milestones_mines}
\resizebox{\textwidth}{!}{%
\begin{tabular}{lccccc}
\toprule
Milestone & CODE-SHARP & CODE-FRP & Finetuned & task-specific & ReAct \\
\midrule
Craft wood pickaxe & 98.5 (98.1, 98.9) & 96.4 (95.1, 97.5) & 98.7 (98.3, 98.9) & 98.7 (98.1, 99.1) & 100.0 (100.0, 100.0) \\
Craft stone pickaxe & 98.3 (97.7, 98.7) & 81.8 (75.4, 87.7) & 96.7 (96.1, 97.4) & 98.4 (97.7, 98.9) & 88.9 (66.7, 100.0) \\
Craft torches & 97.6 (96.7, 98.4) & 73.7 (65.9, 80.3) & 96.2 (95.4, 97.0) & 97.4 (96.4, 98.3) & 77.8 (44.4, 100.0) \\
Descend to dungeon & 89.1 (84.2, 93.4) & 34.7 (28.2, 41.9) & 81.7 (80.3, 83.2) & 34.5 (28.9, 40.3) & 11.1 (0.0, 33.3) \\
Kill 8 dungeon monsters & 69.6 (58.4, 81.3) & 8.9 (4.8, 13.4) & 57.4 (54.1, 60.7) & 0.0 (0.0, 0.0) & 11.1 (0.0, 33.3) \\
Descend to gnomish mines & 47.3 (35.4, 59.6) & 2.6 (1.8, 3.3) & 41.6 (38.3, 44.9) & 0.0 (0.0, 0.0) & 11.1 (0.0, 33.3) \\
Place torch & 40.1 (29.3, 51.9) & 0.4 (0.1, 0.8) & 39.6 (36.8, 42.7) & 0.0 (0.0, 0.0) & 11.1 (0.0, 33.3) \\
Kill bat & 31.7 (19.3, 44.4) & 0.0 (0.0, 0.0) & 38.9 (36.1, 42.2) & 0.0 (0.0, 0.0) & 11.1 (0.0, 33.3) \\
Drink water & 5.9 (3.6, 8.9) & 0.0 (0.0, 0.0) & 16.6 (13.5, 19.5) & 0.0 (0.0, 0.0) & 0.0 (0.0, 0.0) \\
Find diamond & 1.3 (0.8, 1.8) & 0.0 (0.0, 0.0) & 1.2 (0.7, 1.6) & 0.0 (0.0, 0.0) & 0.0 (0.0, 0.0) \\
\midrule
\textbf{Overall Score} & 35.56 (31.24, 40.67) & 6.71 (6.23, 7.14) & 39.21 (36.60, 41.74) & 4.66 (4.56, 4.75) & 6.58 (2.13, 14.87) \\
\bottomrule
\end{tabular}%
}
\end{table}

\begin{table}[h]
\centering
\small
\caption{Per-milestone success rates (\%) on the Dungeon benchmark. Values are mean (95\% CI), computed via percentile bootstrap over 2000 resamples}
\label{tab:milestones_dungeon}
\resizebox{\textwidth}{!}{%
\begin{tabular}{lccccc}
\toprule
Milestone & CODE-SHARP & CODE-FRP & Finetuned & task-specific & ReAct \\
\midrule
Craft wood pickaxe & 98.3 (97.0, 99.2) & 93.8 (88.7, 97.2) & 99.0 (98.7, 99.4) & 99.2 (98.8, 99.6) & 100.0 (100.0, 100.0) \\
Craft stone pickaxe & 98.0 (96.4, 99.0) & 82.0 (74.3, 89.3) & 98.4 (97.7, 99.0) & 99.0 (98.4, 99.6) & 88.9 (66.7, 100.0) \\
Craft stone sword & 97.8 (96.1, 99.0) & 76.2 (65.6, 86.3) & 97.7 (96.8, 98.6) & 99.0 (98.4, 99.6) & 88.9 (66.7, 100.0) \\
Craft torches & 97.0 (95.0, 98.4) & 69.4 (57.4, 79.8) & 97.6 (96.6, 98.5) & 98.8 (98.0, 99.6) & 77.8 (44.4, 100.0) \\
Craft arrows & 96.8 (95.1, 98.3) & 67.4 (56.5, 77.7) & 97.3 (96.2, 98.4) & 98.8 (98.0, 99.6) & 77.8 (44.4, 100.0) \\
Collect sapling & 96.8 (95.0, 98.3) & 67.3 (56.9, 77.9) & 93.6 (92.0, 95.3) & 98.8 (98.0, 99.6) & 33.3 (0.0, 66.7) \\
Descend to dungeon & 88.9 (82.6, 94.3) & 23.7 (18.7, 29.1) & 79.8 (77.1, 82.6) & 38.3 (36.7, 40.2) & 11.1 (0.0, 33.3) \\
Find potion & 83.0 (74.3, 91.0) & 14.5 (9.5, 19.5) & 66.1 (62.4, 69.8) & 24.3 (18.4, 34.4) & 0.0 (0.0, 0.0) \\
Kill dungeon mob & 82.2 (72.6, 90.4) & 12.5 (7.7, 17.8) & 64.9 (60.7, 68.9) & 23.6 (17.6, 34.0) & 0.0 (0.0, 0.0) \\
Ascend to overworld & 68.9 (55.3, 80.9) & 8.1 (5.2, 11.3) & 46.0 (42.6, 49.3) & 10.7 (7.8, 13.3) & 0.0 (0.0, 0.0) \\
Replenish hunger & 68.0 (54.1, 80.1) & 4.0 (3.1, 5.0) & 43.3 (39.5, 46.8) & 0.4 (0.0, 1.2) & 0.0 (0.0, 0.0) \\
\midrule
\textbf{Overall Score} & 87.40 (81.10, 93.00) & 30.97 (25.50, 36.64) & 77.14 (74.87, 79.54) & 38.19 (36.87, 39.06) & 8.03 (5.08, 11.42) \\
\bottomrule
\end{tabular}%
}
\end{table}

\begin{table}[h]
\centering
\small
\caption{Per-milestone success rates (\%) on the Navigation benchmark. Values are mean (95\% CI), computed via bootstrap over 2000 resamples.}
\label{tab:milestones_navigation}
\resizebox{\textwidth}{!}{%
\begin{tabular}{lccccc}
\toprule
Milestone & CODE-SHARP & CODE-FRP & Finetuned & task-specific & ReAct \\
\midrule
Descend to dungeon & 86.2 (80.6, 91.1) & 36.5 (26.0, 46.1) & 92.1 (91.2, 93.0) & 10.3 (9.4, 12.1) & 33.3 (0.0, 66.7) \\
Kill 8 dungeon monsters & 78.9 (72.0, 85.1) & 10.2 (4.0, 18.4) & 56.9 (52.0, 61.2) & 0.0 (0.0, 0.0) & 11.1 (0.0, 33.3) \\
Descend to gnomish mines & 59.8 (50.6, 68.8) & 5.0 (1.4, 9.5) & 38.9 (36.2, 41.5) & 0.0 (0.0, 0.0) & 11.1 (0.0, 33.3) \\
Kill 8 gnomish mines monsters & 12.6 (4.9, 22.4) & 0.0 (0.0, 0.0) & 9.7 (8.2, 11.4) & 0.0 (0.0, 0.0) & 0.0 (0.0, 0.0) \\
Descend to sewers & 0.9 (0.3, 1.6) & 0.0 (0.0, 0.0) & 5.6 (4.7, 6.6) & 0.0 (0.0, 0.0) & 0.0 (0.0, 0.0) \\
Find enchantment table & 0.1 (0.0, 0.3) & 0.0 (0.0, 0.0) & 1.6 (1.2, 2.1) & 0.0 (0.0, 0.0) & 0.0 (0.0, 0.0) \\
\midrule
\textbf{Overall} & 12.72 (10.46, 15.08) & 2.29 (1.41, 3.27) & 17.33 (15.81, 18.84) & 0.50 (0.48, 0.54) & 1.26 (0.00, 3.27) \\
\bottomrule
\end{tabular}%
}
\end{table}



\section{Hyperparameters}
\label{app:hyperparams}

\subsection{Craftax-Classic and XLand-MiniGrid}

\begin{table}[H]
    \centering
    \caption{\Name{} Hyperparameters Craftax-Classic}
    \begin{tabular}{lr}
        \toprule
        \textbf{Hyperparameter} & \textbf{Value} \\
        \midrule
        Proposal Iterations & 25 \\
        Training Epochs per Iteration & 4 \\
        Mutation Start Iteration & 7 \\
        Proposal Evaluation Epochs & 4 \\
        Skill Proposals per Iteration & 10 \\
        Mutation Proposals per Iteration & 5 \\
        Sampled Skills for Mutation per Iteration & 1 \\

        \bottomrule
    \end{tabular}
\end{table}

\begin{table}[H]
    \centering
    \caption{\Name{} Hyperparameters XLand-MiniGrid}
    \begin{tabular}{lr}
        \toprule
        \textbf{Hyperparameter} & \textbf{Value} \\
        \midrule
        Proposal Iterations & 40 \\
        Training Epochs per Iteration & 5 \\
        Mutation Start Iteration & 20 \\
        Proposal Evaluation Epochs & 5 \\
        Skill Proposals per Iteration & 10 \\
        Mutation Proposals per Iteration & 5 \\
        Sampled Skills for Mutation per Iteration & 1 \\

        \bottomrule
    \end{tabular}
\end{table}

\begin{table}[H]
    \centering
    \caption{Hyperparameters for PPO Training (Shared Craftax-Classic and XLand-MiniGrid)}
    \begin{tabular}{lr}
        \toprule
        \textbf{Hyperparameter} & \textbf{Value} \\
        \midrule
        Learning Rate & $2 \times 10^{-4}$ \\
        Batch Size & 512 \\
        Optimizer & AdamW \\
        Discount Factor ($\gamma$) & 0.99 \\
        Entropy Coefficient & 0.01 \\
        Clip Range ($\epsilon$) & 0.2 \\
        GAE Parameter ($\lambda$) & 0.8 \\
        Total Environment Steps & $1 \times 10^8$ \\
        \bottomrule
    \end{tabular}
\end{table}

\begin{table}[H]
    \centering
    \caption{Transformer Architecture Hyperparameters (Shared Craftax-Classic and XLand-MiniGrid)}
    \begin{tabular}{lr}
        \toprule
        \textbf{Hyperparameter} & \textbf{Value} \\
        \midrule
        Architecture & Transformer-XL \\
        Number of Layers & 2 \\
        Number of Attention Heads & 8 \\
        Embedding Dimension ($d_{\text{model}}$) & 256 \\
        QKV Dimension ($d_{\text{qkv}}$) & 256 \\
        MLP Hidden Size & 256 \\
        Memory Window Length ($N_{\text{mem}}$) & 128 \\
        Gradient Window Length ($N_{\text{grad}}$) & 128 \\
        Gating & True (Bias 2.0) \\
        \bottomrule
    \end{tabular}
\end{table}

\subsection{Long Run Experiments}
\begin{table}[H]
    \centering
    \caption{\Name{} Hyperparameters}
    \begin{tabular}{lr}
        \toprule
        \textbf{Hyperparameter} & \textbf{Value} \\
        \midrule
        Proposal Iterations & 100 \\
        Training Epochs per Iteration & 10 \\
        Mutation Start Iteration & 15 \\
        Proposal Evaluation Epochs & 5 \\
        Skill Proposals per Iteration & 10 \\
        Mutation Proposals per Iteration & 5 \\
        Sampled Skills for Mutation per Iteration & 1 \\

        \bottomrule
    \end{tabular}
\end{table}

\begin{table}[H]
    \centering
    \caption{Hyperparameters for PPO Training}
    \begin{tabular}{lr}
        \toprule
        \textbf{Hyperparameter} & \textbf{Value} \\
        \midrule
        Start Learning Rate & $2 \times 10^{-5}$ \\
        Learning Rate Decay & Linear \\
        End Learning Rate & $2 \times 10^{-6}$ \\
        Batch Size & 1024 \\
        Optimizer & AdamW \\
        Discount Factor ($\gamma$) & 0.99 \\
        Entropy Coefficient & 0.02 \\
        Clip Range ($\epsilon$) & 0.2 \\
        GAE Parameter ($\lambda$) & 0.8 \\
        Total Environment Steps & $2 \times 10^9$ \\
        \bottomrule
    \end{tabular}
\end{table}

\begin{table}[H]
    \centering
    \caption{Transformer Architecture Hyperparameters}
    \begin{tabular}{lr}
        \toprule
        \textbf{Hyperparameter} & \textbf{Value} \\
        \midrule
        Architecture & Transformer-XL \\
        Number of Layers & 2 \\
        Number of Attention Heads & 8 \\
        Embedding Dimension ($d_{\text{model}}$) & 256 \\
        QKV Dimension ($d_{\text{qkv}}$) & 256 \\
        MLP Hidden Size & 256 \\
        Memory Window Length ($N_{\text{mem}}$) & 128 \\
        Gradient Window Length ($N_{\text{grad}}$) & 128 \\
        Gating & True (Bias 2.0) \\
        \bottomrule
    \end{tabular}
\end{table}

\subsection{AI2-THOR experiments}
\begin{table}[H]
    \centering
    \caption{\Name{} Hyperparameters AI2-THOR}
    \begin{tabular}{lr}
        \toprule
        \textbf{Hyperparameter} & \textbf{Value} \\
        \midrule
        Proposal Iterations & 100 \\
        Mutation Start Iteration & 25 \\
        Proposal Evaluation Epochs & 5 \\
        Skill Proposals per Iteration & 10 \\
        Mutation Proposals per Iteration & 5 \\
        Sampled Skills for Mutation per Iteration & 1 \\

        \bottomrule
    \end{tabular}
\end{table}

\begin{table}[H]
    \centering
    \caption{Hyperparameters for PPO Training}
    \begin{tabular}{lr}
        \toprule
        \textbf{Hyperparameter} & \textbf{Value} \\
        \midrule
        Start Learning Rate & $1 \times 10^{-4}$ \\
        Learning Rate Decay & None \\
        End Learning Rate & $1 \times 10^{-4}$ \\
        Batch Size & 4096 \\
        Optimizer & Adam \\
        Discount Factor ($\gamma$) & 0.99 \\
        Entropy Coefficient & 0.01 \\
        Clip Range ($\epsilon$) & 0.2 \\
        GAE Parameter ($\lambda$) & 0.95 \\
        Total Environment Steps & $2.5 \times 10^6$ \\
        \bottomrule
    \end{tabular}
\end{table}

\begin{table}[H]
    \centering
    \caption{Model Architecture Hyperparameters (AI2-THOR)}
    \begin{tabular}{lr}
        \toprule
        \textbf{Hyperparameter} & \textbf{Value} \\
        \midrule
        Architecture & CNN + FiLM + LSTM \\
        Number of Conv Layers & 3 \\
        Conv Channels & 32 \\
        Number of FiLM Blocks & 2 (BatchNorm) \\
        LSTM Hidden Size & 128 \\
        Task Embedding Dimension ($d_{\text{task}}$) & 128 \\
        Skill Embedding Input Dimension & 384 \\
        Policy/Value Head Hidden Size & 128 \\
        \bottomrule
    \end{tabular}
\end{table}

\section{Ablations on Individual Benchmark Tasks}
\label{app:detailed_ablation}
We perform a detailed ablation study to isolate the contributions of three core components described in \Cref{subsec:training}: open-ended training (OE), adaptive reward scaling (SR), and opportunistic sampling (OS). For all conditions, the agent is trained on the full generated skill archive, following the chronological order of skill discovery. We evaluate the following configurations:

\textbf{No OE+SR+OS}: To evaluate the necessity of open-ended training, we train the agent in a strictly episodic setting. At the start of each episode, a target SHARP is sampled uniformly at random. The episode terminates after the target SHARP success condition returns \texttt{True} or when the time limit of 300 steps is reached.

\textbf{No SR+OS}: We enable open-ended training but disable adaptive reward scaling and opportunistic sampling. In this setting, the next target SHARP is sampled uniformly from the skill archive discovered up to the current iteration of training.

\textbf{No OS}: We retain open-ended training and adaptive reward scaling but disable opportunistic sampling. New target SHARPs are sampled uniformly from the skill archive discovered up to the current iteration of training, with the returned rewards being scaled based on their current overall success rate.

\textbf{\Name}: The complete framework utilising open-ended training, adaptive reward scaling, and opportunistic sampling.

\begin{table}[H]
    \centering
    \caption{Ablation Impact on Average Score over Benchmarks}
    \label{tab:ablations}
    \begin{tabular}{lcc}
        \toprule
        \textbf{Condition} & \textbf{Average Score} & \textbf{Absolute Decrease} \\
        \midrule
        \Name & 50.55 & -- \\
        No OS & 31.93 & -18.62 \\
        No SR+OS & 21.20 & -29.35 \\
        No OE+SR+OS & 13.50 & -37.05 \\
        \bottomrule
    \end{tabular}
\end{table}

As detailed in \Cref{tab:ablations}, we observe a monotonic degradation in performance as components are removed. The full \Name{} framework achieves an average score of $50.55$, whereas removing opportunistic sampling (No OS) results in a significant absolute decrease of $18.62$, dropping the score to $31.93$. Reverting to a purely episodic setting (No OE+SR+OS) results in the lowest performance with an average score across all benchmark scenarios of $13.50$.

\begin{figure}[H]
    \centering
    \includegraphics[width=1.0\textwidth]{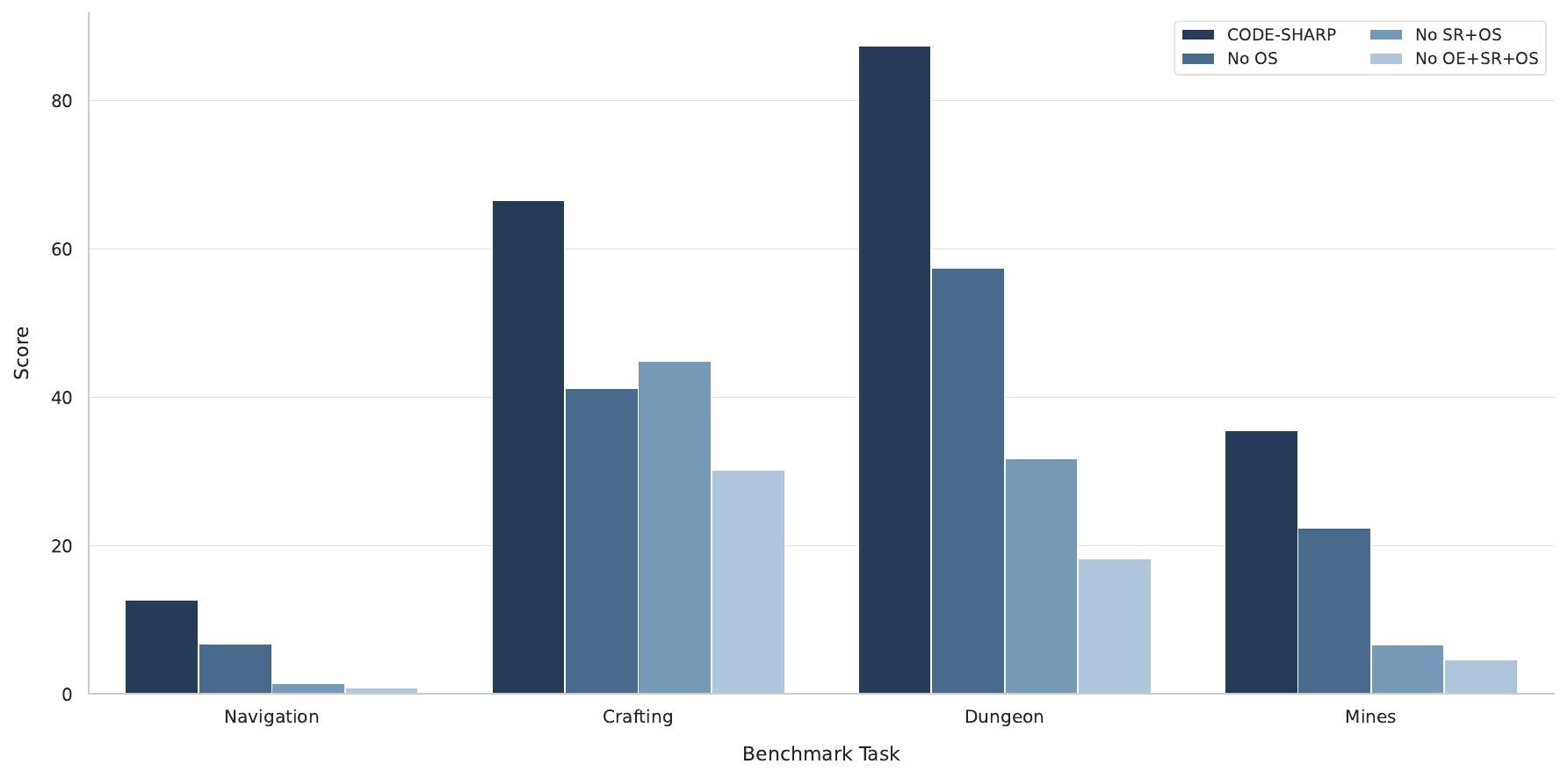}
    \caption{Detailed Ablation Results per Benchmark}
    \label{fig:ablations_scenarios}
\end{figure}

\Cref{fig:ablations_scenarios} further illustrates the impact of these components across individual benchmark tasks. We observe that opportunistic sampling is critical for mastering complex, long-horizon tasks. While all ablations contribute to the agent's success, the data suggest that opportunistic sampling, by dynamically shifting the training distribution toward the frontier of the agent's capabilities, provides the largest singular contribution to performance on challenging benchmarks.

\section{Further Analysis of LLM Judge Performance}
\label{app:llm_judge_analysis}

The LLM judge evaluates proposed SHARPs for \emph{novelty},
\emph{correctness}, and \emph{feasibility}. We audited these criteria
across Craftax-Classic, XLand, and AI2-THOR, examining duplicate
admissions, misspecified goal predicates, and subsequent learnability
rejections.

\begin{table}[ht]
    \centering
    \caption{LLM judge audit. Novelty and correctness rates are computed
    over admitted skills; learnability rejection rates are computed
    over judge-accepted candidates.}
    \label{tab:llm_judge_audit}
    \small
    \begin{tabular}{lccc}
        \hline
        Audit measure & Craftax-Classic & XLand & AI2-THOR \\
        \hline
        Exact duplicates       & 0.0\% & 0.0\%  & 4.6\%  \\
        Misspecified predicates & 9.6\% & 0.0\%  & 0.0\%  \\
        Learnability rejection & 7.1\% & 11.7\% & 18.9\% \\
        \hline
    \end{tabular}
\end{table}

\paragraph{Novelty.}
We identify exact duplicates by hashing the goal predicate together
with its declared environment conditions and prerequisites, thereby
distinguishing skills with different prerequisite routes. Duplicates
were observed only in AI2-THOR. Inspection suggests that descriptions
claimed different prerequisite chains that were not faithfully
implemented, potentially misleading the judge. This audit measures
exact duplication and does not quantify semantic near-duplicates,
whose detection remains a limitation of the judge.

\paragraph{Correctness.}
We manually associate each skill with a verified ground-truth
predicate matching its intended goal. Using a frozen policy
checkpoint, we evaluate each target skill over at least 150 episodes
and compare the success frequencies reported by the generated and
ground-truth predicates. Observed discrepancies occur only for
Craftax-Classic mob-elimination skills: generated predicates
incorrectly interpret mob disappearance as elimination, overlooking
despawning outside the agent's field of view.

\paragraph{Feasibility.}
We measure the fraction of judge-accepted candidates subsequently
rejected by the learnability gate. The judge checks whether required
preconditions are specified, relevant prerequisite skills are
available, and the goal is reachable. Passing these checks does not
establish that the current agent can learn the skill. Consequently,
the reported rejection rates capture disagreement with empirical
learnability rather than verified infeasibility alone.

\section{Sensitivity to Source-Code Access}
\label{app:source_code_access}

We evaluate the dependence of CODE-SHARP on environment source code
by removing it from the Craftax-Classic context, retaining only
environment state information and the entities present. Both
CODE-SHARP runs use Qwen3.5-27B.

\begin{table}[ht]
    \centering
    \caption{Craftax-Classic success rates with and without
    source-code access. Baselines retain full source-code access
    and use Qwen3-235B-A22B.}
    \label{tab:source_code_ablation}
    \small
    \begin{tabular}{llcc}
        \hline
        Method & Source code & Mean & Median \\
        \hline
        ELLM       & Full & 0.398 & 0.081 \\
        OMNI       & Full & 0.392 & 0.158 \\
        CODE-SHARP & None & 0.623 & 0.968 \\
        CODE-SHARP & Full & 0.733 & 0.963 \\
        \hline
    \end{tabular}
\end{table}

OMNI retains full source-code access and for both OMNI and ELLM we report results with Qwen3-235B-A22B. Our results show that CODE-SHARP despite having no source code outperforms them by a wide margin, 0.623 against 0.398 and 0.392, and the difference in medians is larger still. What degrades under reduced information is coverage rather than competence. The medians remain near the ceiling in both CODE-SHARP conditions, so the drop from 0.733 to 0.623 comes from achievements that are never solved at all rather than from degraded execution of those that are.

These failures are instructive, as they show what the method depends on once the source code is withheld. Without source code access, CODE-SHARP never defines working skills for crafting iron tools or mining diamonds. We observe that the LLM falls back on Minecraft crafting recipes, which differ from those in Craftax. Performance therefore rests on the accuracy of the FM's priors and on the capacity of RL training to absorb the errors those priors introduce. Where a prior is qualitatively wrong rather than merely imprecise, neither is sufficient. 

\section{Autonomous Pre-training and Goal-Directed Fine-tuning}
\label{app:scalar_comparison}

We evaluate whether the skill repertoire discovered by CODE-SHARP
provides useful pre-training for downstream goal-directed learning.
Starting from our five Craftax-Classic runs after $10^8$ steps of
autonomous, goal-free discovery, we fine-tune each agent for a
further $2 \times 10^8$ steps, restricting sampling to the discovered
diamond-mining skills and their prerequisites. The reward programs
and prerequisite chains remain those produced during discovery;
only the sampling distribution changes. This fine-tuning phase is
goal-directed through the selection of diamond-related skills.

We compare against SCALAR's reported Transformer-XL result on
diamond mining, matching our policy architecture. All CODE-SHARP
runs use standard Craftax-Classic dynamics and survival constraints.

\begin{table}[ht]
    \centering
    \caption{Diamond-mining success after autonomous pre-training
    and goal-directed fine-tuning, compared with SCALAR's reported
    task-specific training result.}
    \label{tab:scalar_comparison}
    \small
    \begin{tabular}{lcccc}
        \hline
        Method & Pre-training & Task training & Total steps & Success \\
        \hline
        CODE-SHARP & $10^8$ & $2 \times 10^8$ & $3 \times 10^8$
                   & $63.2\% \pm 2.14\%$ \\
        SCALAR & --- & $10^9$ & $10^9$ & $45\%$ \\
        \hline
    \end{tabular}

    \smallskip
    \footnotesize
    SCALAR: Transformer-XL result from Table~1 of its paper;
    no standard deviation is reported.
\end{table}

The result demonstrates that CODE-SHARP's autonomously discovered
skills support effective downstream specialisation. It compares
autonomous pre-training followed by goal-directed fine-tuning
against SCALAR's task-specific training, rather than comparing
the methods under identical open-ended conditions. 

\section{Archive Complexity Evolution}
\label{app:archive_complexity}
\begin{figure}[h]
    \centering
    \includegraphics[width=0.8\textwidth]{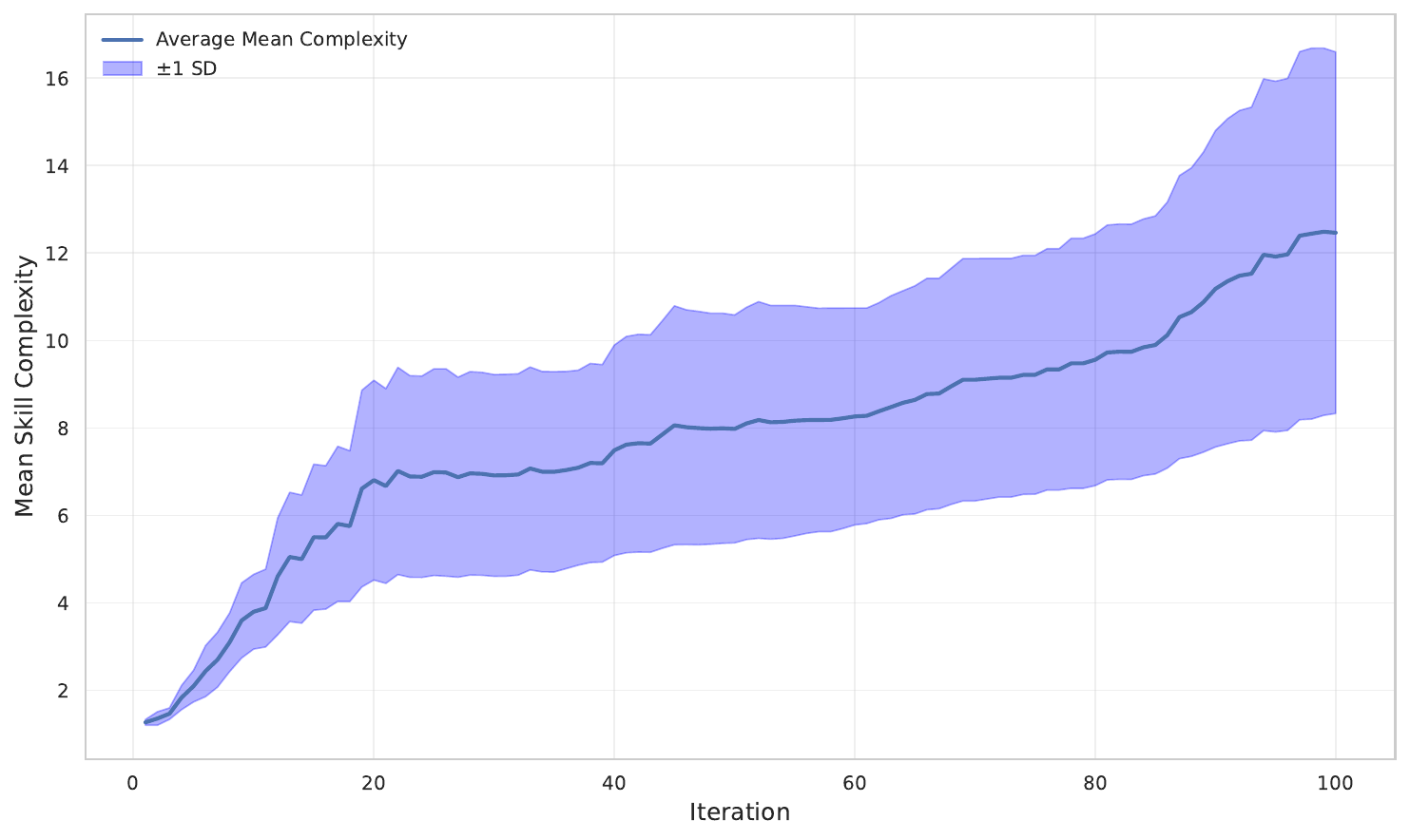}
    \caption{Evolution of average SHARP complexity present in the skill archive}
    \label{fig:ablations_complexity}
\end{figure}

We analyse the evolution of the average complexity of the SHARPs present in the skill archive. For each iteration, we compute each SHARP's complexity according to the complexity equation presented in \Cref{subsec:discovery_analysis}. We observe that the average skill complexity in the skill archive steadily increases with subsequent iterations, as seen in \Cref{fig:ablations_complexity}. This indicates that SHARPs proposed by \Name{} later on in the skill discovery process effectively build on top of the previously implemented SHARPs already present in the skill archive.

\newpage

\section{Examples of Policies-in-Code}
\label{app:example_policies}
We provide an example for the high-level policies-in-code generated by the FM-based policy planner for two benchmark tasks.

\begin{promptbox}[Policy-in-Code for the Navigation Benchmark]
class BenchmarkSolver:
    def __init__(self):
        self.parent_skill_index = 500
        self.assigned_reward = 1.0

        self.cond_fns = (
            lambda states: self._has_drunk_water(*states),
            lambda states: self._has_eaten_food(*states),
            lambda states: self._has_stone_sword(*states),
            lambda states: self._is_on_level1_or_beyond(*states),
            lambda states: self._has_killed_8_mobs_level1(*states),
            lambda states: self._is_on_level2_or_beyond(*states),
            lambda states: self._has_killed_8_mobs_level2(*states),
            lambda states: self._is_on_level3_or_beyond(*states),
            lambda states: self._is_near_enchantment_table(*states),
        )
        
        self.prereq_fns = (
            lambda states: self._do_drink_water(*states),
            lambda states: self._do_eat_food(*states),
            lambda states: self._do_craft_stone_sword(*states),
            lambda states: self._do_descend_ladder(*states),
            lambda states: self._do_kill_eight_mobs_level1(*states),
            lambda states: self._do_descend_ladder_level1(*states),
            lambda states: self._do_kill_eight_mobs_level2(*states),
            lambda states: self._do_descend_ladder_level2(*states),
            lambda states: self._do_enter_new_room_from_corridor(*states),
        )

    def _has_drunk_water(self, prev: Any, cur: Any) -> jnp.bool_:
        target = 7 + 2 * cur.player_dexterity
        on_level0 = jnp.equal(cur.player_level, 0)
        has_full_water = jnp.greater_equal(cur.player_drink, target)
        return jnp.logical_or(jnp.logical_not(on_level0), has_full_water)

    def _has_eaten_food(self, prev: Any, cur: Any) -> jnp.bool_:
        target = 7 + 2 * cur.player_dexterity
        on_level0 = jnp.equal(cur.player_level, 0)
        has_full_food = jnp.greater_equal(cur.player_food, target)
        return jnp.logical_or(jnp.logical_not(on_level0), has_full_food)

    def _has_stone_sword(self, prev: Any, cur: Any) -> jnp.bool_:
        return jnp.greater_equal(cur.inventory.sword, 2)

    def _is_on_level1_or_beyond(self, prev: Any, cur: Any) -> jnp.bool_:
        return jnp.greater_equal(cur.player_level, 1)

    def _has_killed_8_mobs_level1(self, prev: Any, cur: Any) -> jnp.bool_:
        return jnp.greater_equal(cur.monsters_killed[1], 8)

    def _is_on_level2_or_beyond(self, prev: Any, cur: Any) -> jnp.bool_:
        return jnp.greater_equal(cur.player_level, 2)

    def _has_killed_8_mobs_level2(self, prev: Any, cur: Any) -> jnp.bool_:
        return jnp.greater_equal(cur.monsters_killed[2], 8)

    def _is_on_level3_or_beyond(self, prev: Any, cur: Any) -> jnp.bool_:
        return jnp.greater_equal(cur.player_level, 3)

    def _is_near_enchantment_table(self, prev: Any, cur: Any) -> jnp.bool_:
        on_level3 = jnp.equal(cur.player_level, 3)
        near_table = is_near_block(
            cur,
            BlockType.ENCHANTMENT_TABLE_ICE.value,
        )
        return jnp.logical_and(on_level3, near_table)

    def _do_drink_water(self, prev: Any, cur: Any) -> Tuple[int, float, bool]:
        return 4, 0.0, False  # DrinkWater

    def _do_eat_food(self, prev: Any, cur: Any) -> Tuple[int, float, bool]:
        return 61, 0.0, False  # EatFood

    def _do_craft_stone_sword(self, prev: Any, cur: Any) -> Tuple[int, float, bool]:
        return 41, 0.0, False  # CraftStoneSword

    def _do_descend_ladder(self, prev: Any, cur: Any) -> Tuple[int, float, bool]:
        return 103, 0.0, False  # DescendLadder

    def _do_kill_eight_mobs_level1(self, prev: Any, cur: Any) -> Tuple[int, float, bool]:
        return 183, 0.0, False  # KillEightMobsLevel1

    def _do_descend_ladder_level1(self, prev: Any, cur: Any) -> Tuple[int, float, bool]:
        return 196, 0.0, False  # DescendLadderLevel1

    def _do_kill_eight_mobs_level2(self, prev: Any, cur: Any) -> Tuple[int, float, bool]:
        return 251, 0.0, False  # KillEightMobsLevel2

    def _do_descend_ladder_level2(self, prev: Any, cur: Any) -> Tuple[int, float, bool]:
        return 314, 0.0, False  # DescendLadderLevel2

    def _do_enter_new_room_from_corridor(self, prev: Any, cur: Any) -> Tuple[int, float, bool]:
        return 172, 0.0, False  # EnterNewRoomFromCorridorLevel1

    def _finalize(self, prev: Any, cur: Any) -> Tuple[int, float, bool]:
        success = self._is_near_enchantment_table(prev, cur)
        reward = jnp.where(success, self.assigned_reward, 0.0)
        return self.parent_skill_index, reward, success
\end{promptbox}

\begin{promptbox}[Policy-in-Code for the Dungeon Benchmark]
class BenchmarkSolver:
    def __init__(self):
        self.parent_skill_index = 500
        self.assigned_reward = 1.0

        self.cond_fns = (
            lambda states: self._has_drunk_water(*states),
            lambda states: self._has_wood_pickaxe(*states),
            lambda states: self._has_stone_pickaxe(*states),
            lambda states: self._has_stone_sword(*states),
            lambda states: self._has_torches(*states),
            lambda states: self._has_arrows(*states),
            lambda states: self._has_sapling(*states),
            lambda states: self._has_potion(*states),
            lambda states: self._has_killed_two_mobs(*states),
            lambda states: self._has_ascended(*states),
            lambda states: self._has_eaten_food(*states),
        )
        
        self.prereq_fns = (
            lambda states: self._do_drink_water(*states),
            lambda states: self._do_craft_wood_pickaxe(*states),
            lambda states: self._do_craft_stone_pickaxe(*states),
            lambda states: self._do_craft_stone_sword(*states),
            lambda states: self._do_craft_torches(*states),
            lambda states: self._do_craft_arrows(*states),
            lambda states: self._do_collect_sapling(*states),
            lambda states: self._do_find_potion(*states),
            lambda states: self._do_kill_two_mobs(*states),
            lambda states: self._do_ascend(*states),
            lambda states: self._do_eat_food(*states),
        )

    def _has_drunk_water(self, prev, cur):
        return cur.player_drink >= 9

    def _has_wood_pickaxe(self, prev, cur):
        return cur.inventory.pickaxe >= 1

    def _has_stone_pickaxe(self, prev, cur):
        return cur.inventory.pickaxe >= 2

    def _has_stone_sword(self, prev, cur):
        return cur.inventory.sword >= 2

    def _has_torches(self, prev, cur):
        return cur.inventory.torches >= 4

    def _has_arrows(self, prev, cur):
        return cur.inventory.arrows >= 2

    def _has_sapling(self, prev, cur):
        return cur.inventory.sapling > 0

    def _has_potion(self, prev, cur):
        return jnp.any(cur.inventory.potions > 0)

    def _has_killed_two_mobs(self, prev, cur):
        return cur.monsters_killed[1] >= 2

    def _has_ascended(self, prev, cur):
        return cur.player_level == 0

    def _has_eaten_food(self, prev, cur):
        return cur.player_hunger >= 9

    def _do_drink_water(self, prev, cur):
        return 11, 0.0, False  # DrinkWater skill index

    def _do_craft_wood_pickaxe(self, prev, cur):
        return 27, 0.0, False  # CraftWoodPickaxe

    def _do_craft_stone_pickaxe(self, prev, cur):
        return 38, 0.0, False  # CraftStonePickaxe

    def _do_craft_stone_sword(self, prev, cur):
        return 72, 0.0, False  # CraftStoneSword

    def _do_craft_torches(self, prev, cur):
        return 54, 0.0, False  # CraftTorches

    def _do_craft_arrows(self, prev, cur):
        return 85, 0.0, False  # CraftArrow

    def _do_collect_sapling(self, prev, cur):
        return 123, 0.0, False  # GatherSapling

    def _do_find_potion(self, prev, cur):
        return 234, 0.0, False  # LootPotionFromChest

    def _do_kill_two_mobs(self, prev, cur):
        return 118, 0.0, False  # KillOrcSoldier

    def _do_ascend(self, prev, cur):
        return 175, 0.0, False  # PositionOnUpwardLadderDungeon

    def _do_eat_food(self, prev, cur):
        return 167, 0.0, False  # EatSnail

    def _finalize(self, prev, cur):
        success = self._has_eaten_food(prev, cur)
        reward = jnp.where(success, self.assigned_reward, 0.0)
        return self.parent_skill_index, reward, success
\end{promptbox}

\section{Skill Discovery Prompts}
\label{app:discovery_prompts}

\begin{promptbox}[{Skill Proposal Generation Prompt}]
You are a curriculum designer for a reinforcement learning agent in the \$env_name environment.

## Goal


Your task is to **iteratively build a curriculum of hierarchical skills** that progressively expand the agent's capabilities. Each skill you propose becomes a building block that future, more advanced skills can depend on. Over time, this curriculum should guide the agent from basic survival and navigation towards mastering the full environment.

**Key Principle:** Every proposed skill must **meaningfully extend the agent's current abilities** by building logically on top of what it has already learned. Think of the skill archive as a growing dependency tree — your proposal adds the next logical node.

---

## How Skills Work

A skill is a sequence of **Starting Conditions** (checked in order, top to bottom) followed by a **Success Condition**:

1. If a starting condition is **not satisfied**, the system runs the **prerequisite skill** linked to that condition until it is.
2. Once satisfied, the system advances to the next condition.
3. When **all** starting conditions hold, the agent pursues the Success Condition; meeting it earns a reward.

> **Every starting condition MUST be paired with a prerequisite skill from the agent's current repertoire that can fulfil it. No exceptions.**
> If the required prerequisite does not exist, do **not** stretch an unrelated skill to fit. Stop, and propose that simpler prerequisite skill instead.


## Condition Ordering

Order conditions so fulfilling a later one cannot undo an earlier one.
- **Persistent conditions** (poessisng an item) come **first**.
- **Positional / transient conditions** (facing a tile, standing next to an object) come **last**, immediately before the success condition.


## Skill Granularity — hard rules

These are not soft guidelines; a proposal that violates either rule is invalid and will be rejected.

*   **No atomic action-wrapper skills.** If the success condition reduces to "a single primitive action has been executed once" (a rotation, a one-step move), the skill adds no hierarchical value.
*   **No over-generic umbrella skills.** If the success condition is satisfied by many qualitatively different targets (any object, any colour, any door state, any tile of a broad category), the outcome is under-specified and the skill cannot be reused as a prerequisite. If the success condition needs the word *any*, the skill is too broad.
*   A good skill sits **between an action and a generic family**: concrete enough to have a single clear success condition, abstract enough to be reusable as a prerequisite for multiple future skills.


---

## Your Task — Step by Step

1. **Environment Dynamics Analysis (MANDATORY).** Before proposing anything, write out a concrete analysis of the environment from the provided source. Cover:
   (a) the objects present and the functional role each plays;
   (b) for each property, whether it gates a rule, defines a relation between objects, or is cosmetic — justified from the code;
   (c) a layered dependency chain of capabilities, from primitive to composite, making the required prerequisite for each composite capability explicit.
   This analysis grounds every subsequent step.

2. **Analyse the Repertoire.** Study the current skills and their success rates. Identify the most impactful next skill that fits the sampled category and is reachable given the current repertoire. Within the sampled category, inspect which sub-areas are already covered and prefer filling a sub-gap over deepening an already well-covered sub-area. The sampled category itself is binding — this only steers your choice of skill *within* it.

3. **Propose Exactly One Skill.** It must be:
   - in the sampled category,
   - genuinely novel (not a duplicate, trivial variant, or re-implementation of a failed proposal). Example: MineTwoWood if MineWood exists is not novel,
   - not an existing skill with simple changes in the starting conditions,
   - neither a wrapper around a single primitive action nor an under-specified `*Any`-style umbrella (see Skill Granularity above — this is a hard rule, re-check it before committing),
   - consistent with the environment dynamics from Step 1.

4. **Define the Success Condition.** Clear, unambiguous, and expressible from the code's state representation.
   - For crafting and resource gathering: use absolute values (e.g., agent has more than 1).


5. **Define Starting Conditions, each paired with a prerequisite.** For each condition:
   - State the precise environment predicate.
   - Name the prerequisite skill **from the existing repertoire** that fulfils it. Every condition needs at least one.
   - Order conditions so later ones do not invalidate earlier ones (see Condition Ordering).
   - For skills which require objects to be present in the environment, verify they are present by default. If not, starting conditions must include an existence check and an according prerequisite skill which is able to create the object in cases it is not already present.
   - The agent is able to learn atomic skills like moving forwards or placing down objects in free spots by itself. You can assume the agent to be able to perform such atomic actions by itself.
   - If no starting conditions are needed, write: `"No implementation for a condition needed"`.

6. **Assign a reward of 1.**

7. **Learn from Failures.** Review previously failed proposals; do not re-propose them.


---

ENVIRONMENT DESCRIPTION

\$environment_description

---

SAMPLED CATEGORY

\$category

---

SKILL REPERTOIRE:

Current Agent Skill Repertoire: \$skill_repertoire

Success rates: \$success_rates

FAILED PROPOSALS:
Use the previously failed proposals as inspiration for new proposals. Do not directly reimplement them.

\$failed_proposals


OUTPUT FORMAT:
-------------

End your text with the skill proposal in this exact format — do not generate anything after it:

<proposal_start>

"Skill Name": "SkillName1", #Do not include spaces or _ in your skill names
"Skill Type": "The skill type that the proposed skill belongs to",
"Skill Description": "Describe the skill...",
"Assigned Reward": "Integer value for assigned reward",
"Success Condition": "The condition ...",
"Starting Conditions": {
  "Condition 1": ["description of predicate to check condition", "SkillName from repertoire to satisfy condition 1"],
  "Condition 2": ["description of predicate to check condition. If nested condition describe here.", "SkillName from repertoire to satisfy nested condition 1 | SkillName from repertoire to satisfy nested condition 2 | SkillName from repertoire to satisfy nested ... "],
  "Condition 3": ....}

<proposal_end>

<think>

\end{promptbox}

\begin{promptbox}[{Skill Proposal Implementor Prompt}]
You are a JAX coding assistant. Your goal is to translate a "skill proposal" into a valid Python class, implementing it as a new skill for an agent. The skills should be implemented as hierarchical reward programs. Each skill has Precondition Functions to check if necessary sub-goals (e.g., "possess iron," "near crafting table") are met. 
If a precondition fails, the skill calls a prerequisite skill—itself defined with its own preconditions—dedicated to achieving that sub-goal, creating a dependency chain. At each environment step, these preconditions are checked sequentially, allowing the system to traverse this chain to find the first skill whose prerequisites are currently satisfied. This sequential, step-by-step checking means previously met conditions can become false, re-triggering their respective sub-goal skills. Finally, a Success Condition verifies if the main objective has been completed, at which point the skill returns a reward to the agent. Using this formulation, your task is to iteratively expand the agents skill archive via these interconnected skills.

---------------------------------

INPUT: SKILL PROPOSAL
======================

You will receive a proposal containing:
* name: The skill's name.
* description: A brief explanation of what the skill does.
* index: The proposed integer ID for the skill.
* success_condition: A condition that is True only when the skill is completed successfully.
* start_conditions (Optional): A set of prerequisite environment states. Each is paired with a prerequisite skill that can achieve it if not specified otherwise.

---------------------------------

YOUR TASK: VALIDATION & IMPLEMENTATION
======================================

Follow these rules to implement the skill using the provided JAX class template.

1. Validate the Proposal
-------------------------

First, check the proposal for correctness. If it's fundamentally flawed and cannot be fixed, reject it and explain why.

* Success Condition:
    - Ensure it's logical and unambiguous.
    - Correct minor logical errors.
    - For crafting tools the success condition must be given as absolute values, i.e. for wood pickaxe it must be cur.inventory.pickaxe >= 1. For  resources it should be relative to the previous state.

* Start Conditions:
    - Verify that every prerequisite skill exists in the agent's repertoire.
    - If a prerequisite skill is invalid, try to replace it with a valid one. If none exists, reject the proposal.

2. Implement the JAX Class
--------------------------

If the proposal is valid (or you've corrected it), implement the class.

* Indexing:
    - Use the proposed index. If it's already taken, increment the index by one until it is unique.
    - Refer to prerequisite skills by their correct index.

* Code Requirements:
    - Your entire implementation MUST be JAX and JIT-compatible.
    - ALWAYS use JAX logical operators (e.g., jnp.logical_and, jnp.logical_or) instead of Python's native 'and', 'or'.
    - If the proposal specifies a condition with "No implementation for a condition needed" or you can not add the condition to the skill class. You are allowed to simply ignore it by either not adding it to the code or only adding it as a comment


ENVIRONMENT CODE
----------------

All functions and global variables shown in the environment code below are available for direct use in your skill implementation — you do not need to import or redefine them.

\$environment_description

class SkillName:
    """One-line description of what this skill achieves."""

    def __init__(self):
        self.parent_skill_index = \$skill_index   # unique integer, from the proposal
        self.assigned_reward    = 1.0

        self.cond_fns = (
            # Sub-goal 1: one branch per possible case the environment may assign.
            # Exactly one of these returns a meaningful False per episode; the rest pass through.
            lambda states: self._cond_subgoal1_case1(*states),
            lambda states: self._cond_subgoal1_case2(*states),
            # ... one entry per case in sub-goal 1's pool ...

            # Sub-goal 2: same pattern, different pool of cases.
            lambda states: self._cond_subgoal2_case1(*states),
            lambda states: self._cond_subgoal2_case2(*states),
            # ... one entry per case in sub-goal 2's pool ...

            ...
        )
        self.prereq_fns = (
            # Same length and order as cond_fns.
            lambda states: self._do_subgoal1_case1(*states),
            lambda states: self._do_subgoal1_case2(*states),

            lambda states: self._do_subgoal2_case1(*states),
            lambda states: self._do_subgoal2_case2(*states),
        )

    # --- Sub-goal 1 branches ---
    # All branches check the same sub-goal; they differ only in which case is active.

    def _cond_subgoal1_case1(self, prev, cur) -> jnp.bool_:
        applicable = <True when the environment assigned case1 for this sub-goal>
        met        = <True when sub-goal 1 is already satisfied>
        return jnp.logical_or(jnp.logical_not(applicable), met)

    def _do_subgoal1_case1(self, prev, cur):
        return <PREREQ_SKILL_FOR_CASE1>, 0.0, False

    def _cond_subgoal1_case2(self, prev, cur) -> jnp.bool_:
        applicable = <True when the environment assigned case2 for this sub-goal>
        met        = <True when sub-goal 1 is already satisfied>
        return jnp.logical_or(jnp.logical_not(applicable), met)

    def _do_subgoal1_case2(self, prev, cur):
        return <PREREQ_SKILL_FOR_CASE2>, 0.0, False

    # --- Sub-goal 2 branches ---
    def _cond_subgoal2_case1(self, prev, cur) -> jnp.bool_:
        applicable = <True when the environment assigned case1 for this sub-goal>
        met        = <True when sub-goal 2 is already satisfied>
        return jnp.logical_or(jnp.logical_not(applicable), met)

    def _do_subgoal2_case1(self, prev, cur):
        return <PREREQ_SKILL_FOR_CASE1>, 0.0, False

    def _cond_subgoal2_case2(self, prev, cur) -> jnp.bool_:
        applicable = <True when the environment assigned case2 for this sub-goal>
        met        = <True when sub-goal 2 is already satisfied>
        return jnp.logical_or(jnp.logical_not(applicable), met)

    def _do_subgoal2_case2(self, prev, cur):
        return <PREREQ_SKILL_FOR_CASE2>, 0.0, False

    def _finalize(self, prev, cur):
        success = <True when the overall skill goal is achieved>
        reward  = jnp.where(success, self.assigned_reward, 0.0)
        return self.parent_skill_index, reward, success

    #### DO NOT CHANGE ANYTHING BEYOND THIS ####
    def _dummy_condition(self, *args):
        return jnp.bool_(True)

    def _dummy_prereq(self, *args):
        return (-1, 0.0, False)


SKILL REPERTOIRE:
-----------------

Current Agent Skill Repertoire: 

\$skill_repertoire_dicts

\$skill_repertoire


SKILL PROPOSAL
---------------

The following is the skill proposal with its name, description, success condition, and necessary starting conditions. First, verify that all properties are correctly specified according to the rules and environment mechanics, and that all prerequisite skills are present in the agent's skill repertoire. If not, correct these errors before creating the skill class in Python. If the proposal cannot be corrected, do not output a skill class.

\$skill_proposal_one


OUTPUT FORMAT:
-------------

To make it easier to parse your skill proposal, follow this exact format for describing the novel skill. End your text with the skill proposal. Do not generate anything after that:


<start_proposal>

"Skill Name": "SkillName1",
"Skill Index": "The skill's index corresponds to the index provided by the skill proposal",
"Skill Description": "Describe the skill...",
"Skill Function": "The implemented JAX skill function goes here"

<end_proposal>


<think>
\end{promptbox}

\begin{promptbox}[{Skill Proposal Judge Prompt}]
You are the official Skill Novelty Judge for an agent's open-ended skill learning curriculum.
Your mandate is to select at most two of the most valuable new skills from a batch of proposals.

Each skill has Precondition Functions that check whether necessary sub-goals are met.
If a precondition fails, the skill delegates to a prerequisite SHARP — itself defined with its own preconditions — dedicated to achieving that sub-goal, forming a dependency chain.
A Success Condition then verifies that the main objective is complete, at which point the skill returns a reward.

Your decision must balance curriculum coherence (skills that incrementally build on the agent's current abilities) with diversity (skills that open new branches of the skill tree).


---

### Step 1 — Environment Dynamics Analysis (MANDATORY)

Before looking at any proposal, write out a concrete analysis of the environment dynamics from the provided source. This is the single source of truth for every downstream judgement — do not skip or shortcut it. Cover:

(a) **Objects:** every object present and its functional role.
(b) **Properties:** for each property, state whether it gates a rule, defines a relation between objects, or is cosmetic — justified from the code.
(c) **Dependency chain:** the high-level chain of capabilities the dynamics induce, e.g. objects required for crafting, keys for doors etc, all the possible rules which apply to an object or skill.

You will reference this analysis in Step 2 to verify each proposal.


You will reference this analysis in Step 2 to verify each proposal.


### Step 2 — Per-Proposal Scrutiny

Evaluate every proposal against the dynamics analysis from Step 1. A proposal must pass ALL of the following checks; any single failure is an immediate disqualification. You are **not** allowed to correct code or silently add missing prerequisites — if a change would be required, reject the proposal.

*   **Dynamics & Prerequisite Compliance (GATING CHECK):**
    *   Trace the proposal's start conditions and success condition through the dynamics chain from Step 1. The skill must produce its claimed outcome under the real rules. Refer to the point on Feasibility for examples of what is complient with the rules and what is not.
    *   The proposal must list **every** prerequisite skill implied by that chain. If the success condition requires a sub-capability and the corresponding prerequisite is missing, wrong, or not present in the agent's repertoire, reject the proposal — even if it is otherwise novel.
    *   Success and start conditions must be unambiguous, self-contained, and actually reachable by the skill (e.g. do not require picking up an object type absent from the environment by default).
    *   *Note: Fine-grained positional gaps between a prerequisite's end-position and the parent skill's start-position do not count as a missing prerequisite under this check — see the Feasibility bullet for how those are handled.*

*   **Granularity (GATING CHECK):**
    *   **Atomic action wrapper → reject.** If the success condition reduces to "a single primitive action has been executed once" (a rotation, a one-cell move), the skill adds no hierarchical value and must be rejected.
    *   **Over-generic umbrella → reject.** If the success condition is satisfied by many qualitatively different targets (any object, any colour, any door state, any tile of a broad category), the outcome is under-specified and the skill cannot serve as a reusable prerequisite. Reject. If the success condition needs the word *any*, the skill is too broad.
    *   A valid skill sits between a single primitive action and an `*Any`-style family: one precise, named capability with an unambiguous success condition.

*   **Novelty & Precision:**
    *   Genuinely new capability, not a trivial variation or duplicate of an archived skill.
    *   Not a minimal change in capabilities, or an existing skill with added prerequsite skills.
    *   Not a strict re-implementation of a previously failed proposal.

*   **Feasibility:**
    *   The proposal must be feasible given the agent's skill repertoire. If prerequisite skills for high-level sub-tasks are missing, the skill is not feasible.
    *   Focus on the high-level sub-steps required to complete the skill and whether each is covered by a prerequisite from the repertoire. If a proposal assumes the agent can perform atomic tasks — moving forward a step, rotating to face an object, toggling, placing down objects in adjacent spots — but is otherwise well defined, accept it.
    *   **Positional tolerance:** A prerequisite skill that delivers the agent into the vicinity of a target satisfies the positional requirement of the parent skill even if the agent still needs a small number of primitive steps to reach the exact interaction position. The RL agent will learn to bridge these residual steps autonomously when the parent skill becomes active. Do NOT reject a proposal solely because a prerequisite skill ends slightly short of the precise target position.
    *   **Hard limit on tolerance:** This positional tolerance covers only fine-grained proximity adjustments. It does NOT excuse the absence of a high-level prerequisite skill — such as acquiring a required object, unlocking a door, navigating to a different room, or crafting an ingredient. If an entire capability class is unrepresented in the prerequisite list, reject.
    *   Example — valid reason for rejection: An interaction skill requires an object but has no prerequisite skill to acquire or create it. Example — valid reason for rejection: A composite skill requires navigating to a location but has no navigation prerequisite at all. Example — NOT a valid reason for rejection: A prerequisite navigation skill brings the agent near the target, and the agent needs a couple of primitive steps to close the remaining gap before interacting.

*   **Curriculum Value:**
    *   **Natural curriculum progression.** Keep accepting simple skills as long as they are valuable prerequisites for more complex capabilities the agent has not yet reached. The moment **all** prerequisite skills a more complex capability depends on are present in the repertoire **and reliably learned**, that complex capability becomes the preferred pick — graduate to it instead of stacking further simple skills. If even one required prerequisite is missing, the complex capability is not yet reachable; accept the missing prerequisite first.
    *   **Category balance (validity-gated tie-breaker).** If a category is under-represented in the current repertoire and a proposal in that category passes **every** gating check above, prefer it over equally-valid proposals in over-represented categories. This is a tie-breaker among proposals that are already valid. It is **never** grounds to lower the bar on Dynamics, Prerequisite, Granularity, Feasibility, or Technical checks — reject an invalid proposal even if doing so leaves its category empty for this round.
    *   **Genuine capability gain.** Favour proposals that unlock a qualitatively new skill and fill an obvious gap in the skill tree over cosmetic variants of existing skills.

*   **Technical Compliance:**
    *   Fully JIT-compatible, JAX-only code; strict adherence to the provided JAX skill-class template.
    *   If the proposed index is already taken, assign the next available unique integer.


### Step 3 — Final Verdict

Select at most two proposals that pass every check above; reject all if none qualify.
For each selected skill, your justification must explicitly name:
(i) which dynamics-chain prerequisites you verified,
(ii) the curriculum stage the skill targets — simple stepping-stone vs. graduation to a more complex capability — with a direct reference to the prerequisite success rates in `\$skill_performances` that justify that choice,
(iii) whether category-balance diversity played a role in the selection and, if so, an explicit confirmation that the choice was made **among already-valid candidates** and did not lower the bar on any gating check.

Adhere strictly to the rules above. Previously added skills that violated any rule do not license you to repeat the violation.


ENVIRONMENT CODE
----------------

\$environment_description

SKILL REPERTOIRE:
-----------------

Current Agent Skill Repertoire: \$skill_repertoire

Skill performances: \$skill_performances


FAILED PROPOSALS:
----------------

\$failed_proposals


SKILL PROPOSAL
---------------

The following is the skill proposal with its name, description, success condition, and necessary starting conditions.

\$candidate_dict

\$candidate_code


OUTPUT FORMAT:
-------------

If the skill proposal passes your judgement, output your decision following this exact template. Do not output anything after the <decision_end> token.

<decision_start>
[
    {
        "Skill Names": "The name of the first skill you choose",
        "Proposal Index": Must correspond to the index of the chosen skill proposal,
        "Decision Reason": "Explain your reasoning for choosing this specific skill"
    },
    {
        "Skill Names": "The name of the second skill you choose",
        "Proposal Index": Must correspond to the index of the chosen skill proposal,
        "Decision Reason": "Explain your reasoning for choosing this specific skill"
    }
]
<decision_end>

If the skill does not pass your judgement, simply output "Skill did not pass judgement".

<think>

\end{promptbox}

\section{Skill Mutation Prompts}
\label{app:mutation_prompts}

\begin{promptbox}[{Skill Mutation Proposal Prompt}]
You are a helpful skill engineer with the job of designing skills as hierarchical reward programs (SHARP) to train an agent in the \$env_name environment. Each SHARP consists of precondition functions which define the necessary environment conditions that need to be satisfied in order to be able to execute the parent skill. Each precondition can call an already existing skill from the agents archive of learned skills to satisfy the respective condition. Each SHARP further has a success condition which checks if the parent skill is completed and returns the reward.

During the initial implementation there can often be mistakes, oversights or inefficiencies in the defined SHARPs. Your task is to evolve and mutate an already existing SHARP to optimise it based on its current structure and the other skills present in the agents skill archive. For this you will receive a heuristic which you will use as guidance on how to mutate the sampled SHARP. Your mutation will be evaluated against the parent skill. If it has a better performance in terms of the success rate of completing the skill, the mutation will be accepted as the new elite.

Mutation Rules:

1. Ensure your mutations follow the exact heuristic specification given to you and are sensible.
2. If you add a new precondition function, ensure that a relevant skill to satisfy it is present in the agents skill archive.
3. Directly follow the output template given to you to define your mutation proposal.
4. All precondition functions should be clearly marked in the mutation preconditions.
5. Carefully analyse the previous failed mutation attempts, if available, to intelligently propose a next mutation. You should not directly reimplement one of the previous failed mutations.
6. Under no circumstances should you mutate the success condition of the parent skill.

ENVIRONMENT DESCRIPTION

\$environment_description

SKILL ARCHIVE

\$skill_repertoire

PARENT SKILL

\$sampled_parent_skill

\$sampled_parent_skill_code

PREVIOUSLY FAILED PROPOSED MUTATIONS

\$previous_failed_elites

MUTATION HEURISTIC

\$heuristic_name

\$heuristic

OUTPUT

The mutation for this index is \$mutation_index . Be sure to use this exact value for the mutation output dict.
To make it easier to parse your skill proposal, follow this exact format. End your text with the skill proposal — do not generate anything after that:


</mutation_start>

"Skill Name": "ParentSkillName",
"Mutation Name": "Name of the mutated skill",
"Skill Index": "The skill's index corresponds to the skill idx of the parent skill",
"Mutation Index": "\$mutation_index",
"Skill Description": "Describe the skill...",
"Assigned Reward": "The reward you assign to the skill",
"Mutation Description": "Describe the mutation and how it should improve the skill success rate...",
"Success Condition": "The goal state describing succesful completion of the skill...",
"Mutation Starting Conditions": {
  "Condition 1": ["environment predicate", "SkillName from repertoire to satisfy condition 1"],
  "Condition 2": ["environment predicate", "SkillName from repertoire to satisfy condition 2"]
}

</mutation_end>

<think>

\end{promptbox}

\begin{promptbox}[{Mutation Heuristics}]
{
  "Crossover": "Add new preconditions and prerequisite skills to the parent skill that synergize with the sampled skill to enhance its success rate. Focus on synergies that improve survival, ability to navigate the environment, or resource management. You can also replace prerequisite skills for existing preconditions with skills from the agent's repertoire that are more likely to successfully fulfil the specified preconditions.",
  
  "Efficiency": "Streamline the skill's execution flow. Analyze the logical sequence of preconditions, ensuring they are ordered for maximum effectiveness (e.g., gathering tools before attempting tasks).",
  
  "Simplicity": "Optimize the skill's implementation to be as efficient and streamlined as possible by identifying at least one redundant or distracting precondition that can be removed. Redundant preconditions are those that repeat other preconditions objectives. Distracting preconditions are those that force the agent to complete unnecessary objectives, diverting attention away from the skill's primary objective. Carefully analyse which conditions are absolutetly necessary to achieve the sampled skills stated goal and then remove all preconditions deemed distracting or redundant",
  
  "Hyperparameters": "Optimize the numerical parameters governing the skill, such as precondition thresholds. Adjust these parameters to minimize the need for re-visiting preconditions. Ensure the agent gathers sufficient resources in the first instance to sustain the skill's execution."
}
    
\end{promptbox}

\begin{promptbox}[{Skill Mutation Implementor Prompt}]
You are a JAX coding assistant. Your goal is to translate a "skill mutation proposal" into a valid Python class, implementing it as a new skill for an agent.

The skills should be implemented as hierarchical reward programs. Each skill has Precondition Functions to check if necessary sub-goals are met. If a precondition fails, the skill calls a prerequisite skill — itself defined with its own preconditions — dedicated to achieving that sub-goal, creating a dependency chain. At each environment step, these preconditions are checked sequentially, allowing the system to traverse this chain to find the first skill whose prerequisites are currently satisfied. Finally, a Success Condition verifies if the main objective has been completed, at which point the skill returns a reward to the agent. Your task is to translate a proposed mutation of an existing skill into a valid Python class.

INPUT: MUTATION PROPOSAL
======================

You will receive a mutation proposal containing:
* `SkillName`: The skill's name.
* `SkillIndex`: The integer ID for the skill, which you must use.
* `SkillDescription`: A brief explanation of what the skill does.
* `SuccessCondition`: A condition that is `True` only when the skill is completed successfully.
* `Mutation Starting Conditions`: A set of prerequisite environment states for the mutated skill. Each is paired with a prerequisite skill that can achieve it.

---------------------------------

YOUR TASK: VALIDATION & IMPLEMENTATION
======================================

Follow these rules to implement the mutated skill using the provided JAX class template.

**1. Validate the Proposal**
-------------------------

First, check the mutation proposal for correctness. If it's fundamentally flawed and cannot be fixed, reject it and explain why.

* **Success Condition:**
    * Ensure it's logical and unambiguous.
    * Correct minor logical errors.
    * The success condition must exactly match that of the original parent skill — do not modify it.
* **Mutation Starting Conditions:**
    * Verify that every prerequisite skill exists in the agent's repertoire.
    * If a prerequisite skill is invalid, try to replace it with a valid one. If no valid replacement exists, reject the proposal.

**2. Implement the JAX Class**
--------------------------

If the proposal is valid (or you've corrected it), implement the class.

* **Indexing:**
    * Use the exact `SkillIndex` provided in the proposal.
    * Refer to prerequisite skills by their correct index from the skill repertoire.
* **Code Requirements:**
    * Your entire implementation **MUST** be JAX and JIT-compatible.
    * **ALWAYS** use JAX logical operators (e.g., `jnp.logical_and`, `jnp.logical_or`) instead of Python's native `and` or `or`.
    * If the proposal specifies a condition with "No implementation for a condition needed," implement the function body but **must not** append it to `self.cond_fns`, and the respective prerequisite must not be added to `self.prereq_fns`.

---------------------------------

ENVIRONMENT CODE
----------------

\$environment_description

class SkillName:
    """One-line description of what this skill achieves."""

    def __init__(self):
        self.parent_skill_index = \$skill_index   # unique integer, from the proposal
        self.assigned_reward    = 1.0

        self.cond_fns = (
            # Sub-goal 1: one branch per possible case the environment may assign.
            # Exactly one of these returns a meaningful False per episode; the rest pass through.
            lambda states: self._cond_subgoal1_case1(*states),
            lambda states: self._cond_subgoal1_case2(*states),
            # ... one entry per case in sub-goal 1's pool ...

            # Sub-goal 2: same pattern, different pool of cases.
            lambda states: self._cond_subgoal2_case1(*states),
            lambda states: self._cond_subgoal2_case2(*states),
            # ... one entry per case in sub-goal 2's pool ...

            ...
        )
        self.prereq_fns = (
            # Same length and order as cond_fns.
            lambda states: self._do_subgoal1_case1(*states),
            lambda states: self._do_subgoal1_case2(*states),

            lambda states: self._do_subgoal2_case1(*states),
            lambda states: self._do_subgoal2_case2(*states),
        )

    # --- Sub-goal 1 branches ---
    # All branches check the same sub-goal; they differ only in which case is active.

    def _cond_subgoal1_case1(self, prev, cur) -> jnp.bool_:
        applicable = <True when the environment assigned case1 for this sub-goal>
        met        = <True when sub-goal 1 is already satisfied>
        return jnp.logical_or(jnp.logical_not(applicable), met)

    def _do_subgoal1_case1(self, prev, cur):
        return <PREREQ_SKILL_FOR_CASE1>, 0.0, False

    def _cond_subgoal1_case2(self, prev, cur) -> jnp.bool_:
        applicable = <True when the environment assigned case2 for this sub-goal>
        met        = <True when sub-goal 1 is already satisfied>
        return jnp.logical_or(jnp.logical_not(applicable), met)

    def _do_subgoal1_case2(self, prev, cur):
        return <PREREQ_SKILL_FOR_CASE2>, 0.0, False

    # --- Sub-goal 2 branches ---
    def _cond_subgoal2_case1(self, prev, cur) -> jnp.bool_:
        applicable = <True when the environment assigned case1 for this sub-goal>
        met        = <True when sub-goal 2 is already satisfied>
        return jnp.logical_or(jnp.logical_not(applicable), met)

    def _do_subgoal2_case1(self, prev, cur):
        return <PREREQ_SKILL_FOR_CASE1>, 0.0, False

    def _cond_subgoal2_case2(self, prev, cur) -> jnp.bool_:
        applicable = <True when the environment assigned case2 for this sub-goal>
        met        = <True when sub-goal 2 is already satisfied>
        return jnp.logical_or(jnp.logical_not(applicable), met)

    def _do_subgoal2_case2(self, prev, cur):
        return <PREREQ_SKILL_FOR_CASE2>, 0.0, False

    def _finalize(self, prev, cur):
        success = <True when the overall skill goal is achieved>
        reward  = jnp.where(success, self.assigned_reward, 0.0)
        return self.parent_skill_index, reward, success

    #### DO NOT CHANGE ANYTHING BEYOND THIS ####
    def _dummy_condition(self, *args):
        return jnp.bool_(True)

    def _dummy_prereq(self, *args):
        return (-1, 0.0, False)


SKILL REPERTOIRE
Current Agent Skill Repertoire: \$skill_repertoire

MUTATION PROPOSAL
The following is the mutation proposal. First, verify that all properties are correctly specified according to
the rules and environment mechanics and that all prerequisite skills are present in the agent's skill repertoire.
If not, correct these errors before creating the skill class in Python. If the proposal cannot be corrected,
do not output a skill class.

\$mutation_proposal

SAMPLED PARENT SKILL CODE
This is the code of the sampled parent skill. It can diverge from the code in the base repertoire. Always use the version
given to you here as the base for your code.

\$sampled_parent_skill

OUTPUT FORMAT
To make it easier to parse your skill definition, follow this exact format:

<start_proposal>

"Skill Name": "SkillName1",
"Skill Index": "The skill's index corresponds to the index provided by the skill proposal",
"Skill Description": "Describe the skill...",
"Skill Function": "The implemented JAX skill function goes here"

<end_proposal>


<think>

\end{promptbox}

\section{Policy Planner Prompt}
\label{app:policy_planner_prompts}

\begin{promptbox}[{Policy Planner Prompt}]
You are an expert AI architect guiding an agent to solve complex tasks in the Craftax environment. Your goal is to design a high-level policy skill, `BenchmarkSolver`, that orchestrates existing skills from a repertoire to complete a specific benchmark.

The `BenchmarkSolver` acts as a meta-policy. You must define a sequence of **preconditions** (checks for specific states) and map each to a **prerequisite skill** (an action to achieve that state). The agent will execute these in order.

YOUR TASK: IMPLEMENT BENCHMARK SOLVER
======================================

Implement the `BenchmarkSolver` skill using the JAX class template provided. This skill orchestrates the agent to achieve the provided **Benchmark Milestones**.

1. Skill Structure
------------------
* **Class Name:** `BenchmarkSolver`
* **Index:** $next_skill_index
* **Template:** Use the provided JAX class structure.

2. Logic & Strategy Requirements
--------------------------------
You must define the `cond_fns` (preconditions) and `prereq_fns` (actions) lists. Construct them using the following logic:

### A. Strategic Planning
Before strictly following the provided milestones, you must analyze the task requirements to ensure agent survival and efficiency.
* **Preparation:** You are encouraged to define a limited set of auxiliary preconditions for essential tools that are not explicitly listed in the milestones but can assist the agent in completing the milestones.
* **Focus:** Focus on the only the most critical tools that are required to complete the milestones. Avoid adding too many preparatory steps as it may lead to the agent getting distracted from the main task of solving milestones.
* **Efficiency:** The agent can get distracted if you add too many preparatory steps. Ensure that you limit the number of preparatory steps to a minimum and that they can be completed easily by the agent.
* **Ordering:** These preparatory steps must be placed only at the **beginning** of your precondition sequence, ensuring the agent is equipped *before* attempting to solve the milestones.

### B. Milestone Execution
* After your preparatory steps, define preconditions that mirror the **Benchmark Milestones** in order.
* Ensure the logical flow allows the agent to progress from one milestone to the next without getting stuck.

### C. Skill Selection
* For each precondition, assign the most effective skill from the **Skill Repertoire**.
* **Best Fit:** Choose the skill with the highest success rate that achieves the goal.
* **Approximation:** If no skill perfectly matches a milestone, use the closest available alternative or a skill that covers multiple milestones (e.g., a "Mine Stone" skill might naturally cover a "Craft Stone Pickaxe" milestone).
* **Constraint:** You must use *existing* skills only. Do not invent new skill names.

3. Code Requirements
--------------------
* **JAX Compatibility:** The implementation must be pure JAX and JIT-compatible.
* **Operators:** ALWAYS use `jnp.logical_and`, `jnp.logical_or`, etc., instead of Python native operators.
* **Imports:** Do not add import statements; assume the environment is pre-loaded.

---------------------------------

CRAFTAX ENVIRONMENT
-------------------

$environment_description

BENCHMARK MILESTONES
--------------------

The benchmark task consists of the following milestones that should be achieved in order:

$milestones

---------------------------------

SKILL REPERTOIRE
----------------

Current Agent Skill Repertoire:

$skill_repertoire

Skill Success Rates:

$success_rates

---------------------------------

SKILL FUNCTION TEMPLATE
--------------------

class BenchmarkSolver:
    # Top-level policy for solving the benchmark task.

    def __init__(self):
        self.parent_skill_index = $next_skill_index
        self.assigned_reward = 1.0   

        #### Implement the tuples containing the condition and prerequisite functions ####
        # NOTE: The Wrapper requires these to be tuples of lambdas accepting a single 'states' argument.
        # Always keep the dummy functions first to prevent empty tuple errors during JIT compilation.
        
        self.cond_fns = (
            # lambda states: self._condition_1(*states),
            # lambda states: self._condition_2(*states),
        )
        
        self.prereq_fns = (
            # lambda states: self._do_condition1(*states),
            # lambda states: self._do_condition2(*states),
        )

    #### Implement the logic functions here ####

    # def _condition_1(self, prev: Any, cur: Any) -> jnp.bool_:
    #     # Return True if the condition is satisfied (e.g., item is in inventory)
    #     return cur.inventory.wood >= 1

    # def _do_condition1(self, prev: Any, cur: Any) -> Tuple[int, float, bool]:
    #     # Return the index of the skill required to satisfy _condition_1
    #     # Returns: (Prerequisite Skill Index, 0.0, False)
    #     return 5, 0.0, False # Example: Index 5 is "Collect Wood"

    #### Implement the success condition (Finalize) here ####

    def _finalize(self, prev: Any, cur: Any) -> Tuple[int, float, bool]:
        # Implement the ultimate success check for this skill
        # e.g., success = cur.inventory.crafting_table >= 1
        success = False 
        
        reward  = jnp.where(success, self.assigned_reward, 0.0)
        return self.parent_skill_index, reward, success


OUTPUT FORMAT:
-------------

To make it easier to parse your skill proposal, follow this exact format for describing the novel skill. End your text with the skill proposal. Do not generate anything after that:


</start_proposal>

"Skill Name": "BenchmarkSolver",
"Skill Description": "Describe the skill...",
"Skill Function": "The implemented JAX skill function goes here"

</end_proposal>
\end{promptbox}

\section{Discovered Skill Archive}
\label{app:discovered_skills}
In the following we present the skill archive discovered during one of the three runs, consisting of 93 discovered SHARPs. The skills  were discovered over 100 iterations of skill proposal and agent training.

\begin{tcolorbox}[skillbox, title=FindTree]
    \textbf{Description:} The agent explores the environment until it is in the proximity of a tree (within a radius of 2 blocks). \\
    \textbf{Skill Type:} Navigation \\
    \textbf{Assigned Reward:} 1 \\
    \textbf{Success Condition:} \code{agent near block: BlockType.TREE} \\
    \textbf{Starting Conditions:} None
\end{tcolorbox}

\vspace{0.5cm}

\begin{tcolorbox}[skillbox, title=FindLake]
    \textbf{Description:} The agent explores the environment to locate a lake, a vital water source to manage its thirst. This skill aims to teach the agent to identify and approach water bodies for survival. \\
    \textbf{Skill Type:} Navigation \\
    \textbf{Assigned Reward:} 1 \\
    \textbf{Success Condition:} \code{agent near block: BlockType.WATER} \\
    \textbf{Starting Conditions:} None
\end{tcolorbox}

\vspace{0.5cm}

\begin{tcolorbox}[skillbox, title=FindCow]
    \textbf{Description:} The agent explores the overworld until it is within 2 blocks of a cow, a passive mob that can be killed for food. \\
    \textbf{Skill Type:} Navigation \\
    \textbf{Assigned Reward:} 1 \\
    \textbf{Success Condition:} \code{agent near passive mob: PassiveMobType.COW} \\
    \textbf{Starting Conditions:} None
\end{tcolorbox}

\vspace{0.5cm}

\begin{tcolorbox}[skillbox, title=DrinkWater]
    \textbf{Description:} The agent locates a water source and drinks from it to replenish its thirst intrinsic to maximum capacity, ensuring survival through hydration management. \\
    \textbf{Skill Type:} Survival
    \textbf{Assigned Reward:} 1 \\
    \textbf{Success Condition:} \code{agent drink >= (7 + 2 * agent dexterity)} \\
    \textbf{Starting Conditions:}
    \begin{itemize}[noitemsep, topsep=0pt]
        \item \textbf{C1:} \code{agent near block: BlockType.WATER} (Requires: FindLake)
    \end{itemize}
\end{tcolorbox}

\vspace{0.5cm}

\begin{tcolorbox}[skillbox, title=MineWood]
    \textbf{Description:} The agent mines a tree to obtain wood, a fundamental resource required for crafting tools and structures. This skill builds upon tree location by teaching the agent to interact with trees to gather raw materials. \\
    \textbf{Skill Type:} Resource Gathering \\
    \textbf{Assigned Reward:} 1 \\
    \textbf{Success Condition:} \code{agent inventory wood >= 1} \\
    \textbf{Starting Conditions:}
    \begin{itemize}[noitemsep, topsep=0pt]
        \item \textbf{C1:} \code{agent near block: BlockType.TREE} (Requires: FindTree)
    \end{itemize}
\end{tcolorbox}

\vspace{0.5cm}

\begin{tcolorbox}[skillbox, title=KillCow]
    \textbf{Description:} The agent engages and defeats a cow, a passive mob found in the overworld, to obtain food resources. This skill teaches the agent basic combat mechanics against passive mobs. \\
    \textbf{Skill Type:} Fighting Mobs \\
    \textbf{Assigned Reward:} 1 \\
    \textbf{Success Condition:} \code{agent killed mob: PassiveMobType.COW} \\
    \textbf{Starting Conditions:}
    \begin{itemize}[noitemsep, topsep=0pt]
        \item \textbf{C1:} \code{agent near passive mob: PassiveMobType.COW} (Requires: FindCow)
    \end{itemize}
\end{tcolorbox}

\vspace{0.5cm}

\begin{tcolorbox}[skillbox, title=PlaceCraftingTable]
    \textbf{Description:} The agent places a crafting table using wood resources, creating a necessary structure for crafting tools and enabling progression to more advanced skills. \\
    \textbf{Assigned Reward:} 1 \\
    \textbf{Skill Type:} Crafting \\
    \textbf{Success Condition:} \code{agent near block: BlockType.CRAFTING\_TABLE} \\
    \textbf{Starting Conditions:}
    \begin{itemize}[noitemsep, topsep=0pt]
        \item \textbf{C1:} \code{agent inventory wood >= 2} (Requires: MineWood)
    \end{itemize}
\end{tcolorbox}

\vspace{0.5cm}

\begin{tcolorbox}[skillbox, title=FindStone]
    \textbf{Description:} The agent explores the overworld to locate stone deposits within a 2-block radius, which are essential for crafting stone tools and advancing resource gathering capabilities. \\
    \textbf{Skill Type:} Navigation \\
    \textbf{Assigned Reward:} 1 \\
    \textbf{Success Condition:} \code{agent near block: BlockType.STONE} \\
    \textbf{Starting Conditions:} None
\end{tcolorbox}

\vspace{0.5cm}

\begin{tcolorbox}[skillbox, title=CraftWoodPickaxe]
    \textbf{Description:} The agent crafts a wooden pickaxe at a crafting table, enabling stone mining for progression. This skill builds upon wood gathering and table placement to teach core crafting mechanics. \\
    \textbf{Skill Type:} Crafting \\
    \textbf{Assigned Reward:} 1 \\
    \textbf{Success Condition:} \code{agent inventory pickaxe >= 1} \\
    \textbf{Starting Conditions:}
    \begin{itemize}[noitemsep, topsep=0pt]
        \item \textbf{C1:} \code{agent inventory wood >= 1} (Requires: MineWood)
        \item \textbf{C2:} \code{agent near block: BlockType.CRAFTING\_TABLE} (Requires: PlaceCraftingTable)
    \end{itemize}
\end{tcolorbox}

\vspace{0.5cm}

\begin{tcolorbox}[skillbox, title=CraftWoodSword]
    \textbf{Description:} The agent crafts a wooden sword at a crafting table using gathered wood resources. This skill builds upon wood acquisition and crafting table placement to teach the agent foundational combat tool creation. \\
    \textbf{Skill Type:} Crafting \\
    \textbf{Assigned Reward:} 1 \\
    \textbf{Success Condition:} \code{agent inventory sword >= 1} \\
    \textbf{Starting Conditions:}
    \begin{itemize}[noitemsep, topsep=0pt]
        \item \textbf{C1:} \code{agent inventory wood >= 1} (Requires: MineWood)
        \item \textbf{C2:} \code{agent near block: BlockType.CRAFTING\_TABLE} (Requires: PlaceCraftingTable)
    \end{itemize}
\end{tcolorbox}

\begin{tcolorbox}[skillbox, title=MineStone]
    \textbf{Skill Type:} Resource Gathering \\
    \textbf{Description:} The agent locates stone deposits and mines them using a wooden pickaxe to obtain stone resources. \\
    \textbf{Assigned Reward:} 1 \\
    \textbf{Success Condition:} \code{agent inventory stone >= 1} \\
    \textbf{Starting Conditions:}
    \begin{itemize}[noitemsep, topsep=0pt, leftmargin=*]
        \item \textbf{C1:} \code{agent inventory pickaxe >= 1} (Req: CraftWoodPickaxe)
        \item \textbf{C2:} \code{agent near block: BlockType.STONE} (Req: FindStone)
    \end{itemize}
\end{tcolorbox}

\begin{tcolorbox}[skillbox, title=CraftStonePickaxe]
    \textbf{Skill Type:} Crafting \\
    \textbf{Description:} The agent crafts a stone pickaxe at a crafting table using gathered wood and stone resources. \\
    \textbf{Assigned Reward:} 1 \\
    \textbf{Success Condition:} \code{agent inventory pickaxe >= 2} \\
    \textbf{Starting Conditions:}
    \begin{itemize}[noitemsep, topsep=0pt, leftmargin=*]
        \item \textbf{C1:} \code{agent inventory wood >= 1} (Req: MineWood)
        \item \textbf{C2:} \code{agent inventory stone >= 1} (Req: MineStone)
        \item \textbf{C3:} \code{agent near block: BlockType.CRAFTING\_TABLE} (Req: PlaceCraftingTable)
    \end{itemize}
\end{tcolorbox}

\begin{tcolorbox}[skillbox, title=FindCoal]
    \textbf{Skill Type:} Navigation \\
    \textbf{Description:} The agent explores the overworld (level 0) to locate coal deposits within a 2-block radius. \\
    \textbf{Assigned Reward:} 1 \\
    \textbf{Success Condition:} \code{(agent level == 0) \& agent near block: BlockType.COAL} \\
    \textbf{Starting Conditions:}
    \begin{itemize}[noitemsep, topsep=0pt, leftmargin=*]
        \item \textbf{C1:} \code{agent near block: BlockType.STONE} (Req: FindStone)
    \end{itemize}
\end{tcolorbox}

\begin{tcolorbox}[skillbox, title=MineCoal]
    \textbf{Skill Type:} Resource Gathering \\
    \textbf{Description:} The agent locates coal deposits identified in stone-rich areas and mines them using a wooden pickaxe. \\
    \textbf{Assigned Reward:} 1 \\
    \textbf{Success Condition:} \code{agent inventory coal >= 1} \\
    \textbf{Starting Conditions:}
    \begin{itemize}[noitemsep, topsep=0pt, leftmargin=*]
        \item \textbf{C1:} \code{agent inventory pickaxe >= 1} (Req: CraftWoodPickaxe)
        \item \textbf{C2:} \code{agent near block: BlockType.COAL} (Req: FindCoal)
    \end{itemize}
\end{tcolorbox}

\begin{tcolorbox}[skillbox, title=CraftStoneSword]
    \textbf{Skill Type:} Crafting \\
    \textbf{Description:} The agent crafts a stone sword at a crafting table using gathered wood and stone resources. \\
    \textbf{Assigned Reward:} 1 \\
    \textbf{Success Condition:} \code{agent inventory sword >= 2} \\
    \textbf{Starting Conditions:}
    \begin{itemize}[noitemsep, topsep=0pt, leftmargin=*]
        \item \textbf{C1:} \code{agent inventory wood >= 1} (Req: MineWood)
        \item \textbf{C2:} \code{agent inventory stone >= 1} (Req: MineStone)
        \item \textbf{C3:} \code{agent near block: BlockType.CRAFTING\_TABLE} (Req: PlaceCraftingTable)
    \end{itemize}
\end{tcolorbox}

\begin{tcolorbox}[skillbox, title=PlaceFurnace]
    \textbf{Skill Type:} Crafting \\
    \textbf{Description:} The agent places a furnace using gathered stone resources, creating essential infrastructure for smelting ores. \\
    \textbf{Assigned Reward:} 1 \\
    \textbf{Success Condition:} \code{agent near block: BlockType.FURNACE} \\
    \textbf{Starting Conditions:}
    \begin{itemize}[noitemsep, topsep=0pt, leftmargin=*]
        \item \textbf{C1:} \code{agent inventory stone >= 1} (Req: MineStone)
    \end{itemize}
\end{tcolorbox}

\begin{tcolorbox}[skillbox, title=CraftTorch]
    \textbf{Skill Type:} Crafting \\
    \textbf{Description:} The agent crafts torches at a crafting table using gathered wood and coal resources. \\
    \textbf{Assigned Reward:} 1 \\
    \textbf{Success Condition:} \code{agent inventory torches >= 1} \\
    \textbf{Starting Conditions:}
    \begin{itemize}[noitemsep, topsep=0pt, leftmargin=*]
        \item \textbf{C1:} \code{agent inventory wood >= 1} (Req: MineWood)
        \item \textbf{C2:} \code{agent inventory coal >= 1} (Req: MineCoal)
        \item \textbf{C3:} \code{agent near block: BlockType.CRAFTING\_TABLE} (Req: PlaceCraftingTable)
    \end{itemize}
\end{tcolorbox}

\begin{tcolorbox}[skillbox, title=MineSapling]
    \textbf{Skill Type:} Resource Gathering \\
    \textbf{Description:} The agent mines grass blocks in the overworld (level 0) near cows to obtain saplings. \\
    \textbf{Assigned Reward:} 1 \\
    \textbf{Success Condition:} \code{agent inventory sapling >= 1} \\
    \textbf{Starting Conditions:}
    \begin{itemize}[noitemsep, topsep=0pt, leftmargin=*]
        \item \textbf{C1:} \code{agent near passive mob: PassiveMobType.COW} (Req: FindCow)
    \end{itemize}
\end{tcolorbox}

\begin{tcolorbox}[skillbox, title=EatFood]
    \textbf{Skill Type:} Survival \\
    \textbf{Description:} The agent locates and kills cows in the overworld (level 0) until its food level reaches maximum capacity. \\
    \textbf{Assigned Reward:} 1 \\
    \textbf{Success Condition:} \code{(agent level == 0) \& (agent food >= (7 + 2 * agent dexterity))} \\
    \textbf{Starting Conditions:}
    \begin{itemize}[noitemsep, topsep=0pt, leftmargin=*]
        \item \textbf{C1:} \code{agent near passive mob: PassiveMobType.COW} (Req: FindCow)
    \end{itemize}
\end{tcolorbox}

\begin{tcolorbox}[skillbox, title=CraftArrow]
    \textbf{Skill Type:} Crafting \\
    \textbf{Description:} The agent crafts arrows at a crafting table using gathered wood and stone resources. \\
    \textbf{Assigned Reward:} 1 \\
    \textbf{Success Condition:} \code{agent inventory arrows >= 1} \\
    \textbf{Starting Conditions:}
    \begin{itemize}[noitemsep, topsep=0pt, leftmargin=*]
        \item \textbf{C1:} \code{agent inventory wood >= 1} (Req: MineWood)
        \item \textbf{C2:} \code{agent inventory stone >= 1} (Req: MineStone)
        \item \textbf{C3:} \code{agent near block: BlockType.CRAFTING\_TABLE} (Req: PlaceCraftingTable)
    \end{itemize}
\end{tcolorbox}

\begin{tcolorbox}[skillbox, title=KillZombie]
    \textbf{Skill Type:} Fighting Mobs \\
    \textbf{Description:} The agent locates and defeats a zombie, an aggressive melee mob that spawns in stone-rich areas. \\
    \textbf{Assigned Reward:} 1 \\
    \textbf{Success Condition:} \code{(agent level == 0) \& agent killed mob: MeleeMobType.ZOMBIE} \\
    \textbf{Starting Conditions:}
    \begin{itemize}[noitemsep, topsep=0pt, leftmargin=*]
        \item \textbf{C1:} \code{agent near melee mob: MeleeMobType.ZOMBIE} (Req: FindStone)
    \end{itemize}
\end{tcolorbox}

\begin{tcolorbox}[skillbox, title=MineIron]
    \textbf{Skill Type:} Resource Gathering \\
    \textbf{Description:} The agent mines iron ore deposits using a stone pickaxe, obtaining iron resources. \\
    \textbf{Assigned Reward:} 1 \\
    \textbf{Success Condition:} \code{agent inventory iron >= 1} \\
    \textbf{Starting Conditions:}
    \begin{itemize}[noitemsep, topsep=0pt, leftmargin=*]
        \item \textbf{C1:} \code{agent inventory pickaxe >= 2} (Req: CraftStonePickaxe)
        \item \textbf{C2:} \code{agent near block: BlockType.STONE} (Req: FindStone)
    \end{itemize}
\end{tcolorbox}

\begin{tcolorbox}[skillbox, title=KillSkeleton]
    \textbf{Skill Type:} Fighting Mobs \\
    \textbf{Description:} The agent locates and defeats a skeleton, a ranged mob that spawns in stone-rich areas. \\
    \textbf{Assigned Reward:} 1 \\
    \textbf{Success Condition:} \code{(agent level == 0) \& agent killed mob: RangedMobType.SKELETON} \\
    \textbf{Starting Conditions:}
    \begin{itemize}[noitemsep, topsep=0pt, leftmargin=*]
        \item \textbf{C1:} \code{agent inventory sword >= 1} (Req: CraftWoodSword)
        \item \textbf{C2:} \code{(agent level == 0) \& agent near ranged mob: RangedMobType.SKELETON} (Req: FindStone)
    \end{itemize}
\end{tcolorbox}

\begin{tcolorbox}[skillbox, title=SleepToFullEnergy]
    \textbf{Skill Type:} Survivability \\
    \textbf{Description:} The agent ensures its survival intrinsics are stabilized and eliminates threats before sleeping to recover energy. \\
    \textbf{Assigned Reward:} 1 \\
    \textbf{Success Condition:} \code{(agent level == 0) \& (agent energy >= (7 + 2 * agent dexterity))} \\
    \textbf{Starting Conditions:}
    \begin{itemize}[noitemsep, topsep=0pt, leftmargin=*]
        \item \textbf{C1:} \code{(agent level == 0) \& (agent food >= (7 + 2 * agent dexterity))} (Req: EatFood)
        \item \textbf{C2:} \code{(agent level == 0) \& (agent drink >= (7 + 2 * agent dexterity))} (Req: DrinkWater)
        \item \textbf{C3:} \code{not agent near melee mob: MeleeMobType.ZOMBIE} (Req: KillZombie)
        \item \textbf{C4:} \code{not agent near ranged mob: RangedMobType.SKELETON} (Req: KillSkeleton)
    \end{itemize}
\end{tcolorbox}

\begin{tcolorbox}[skillbox, title=FindLadderDown]
    \textbf{Skill Type:} Navigation \\
    \textbf{Description:} The agent explores the overworld (level 0) to locate the downward ladder required to descend to the dungeon level. \\
    \textbf{Assigned Reward:} 1 \\
    \textbf{Success Condition:} \code{(agent level == 0) \& local object exists: ItemType.LADDER\_DOWN} \\
    \textbf{Starting Conditions:} None
\end{tcolorbox}

\begin{tcolorbox}[skillbox, title=PlaceTorch]
    \textbf{Skill Type:} Survivability \\
    \textbf{Description:} The agent utilizes crafted torches to place them on valid adjacent blocks, illuminating the immediate surroundings. \\
    \textbf{Assigned Reward:} 1 \\
    \textbf{Success Condition:} \code{stationary object exists: ItemType.TORCH} \\
    \textbf{Starting Conditions:}
    \begin{itemize}[noitemsep, topsep=0pt, leftmargin=*]
        \item \textbf{C1:} \code{agent inventory torches >= 1} (Req: CraftTorch)
    \end{itemize}
\end{tcolorbox}

\begin{tcolorbox}[skillbox, title=RecoverHealth]
    \textbf{Skill Type:} Survivability \\
    \textbf{Description:} The agent ensures its survival intrinsics are fully replenished and in a safe environment, allowing passive health regeneration. \\
    \textbf{Assigned Reward:} 1 \\
    \textbf{Success Condition:} \code{(agent level == 0) \& (agent health >= (8 + agent strength))} \\
    \textbf{Starting Conditions:}
    \begin{itemize}[noitemsep, topsep=0pt, leftmargin=*]
        \item \textbf{C1:} \code{(agent level == 0) \& (agent energy >= (7 + 2 * agent dexterity))} (Req: SleepToFullEnergy)
    \end{itemize}
\end{tcolorbox}

\begin{tcolorbox}[skillbox, title=StepOntoLadderDown]
    \textbf{Skill Type:} Navigation \\
    \textbf{Description:} The agent moves from being within a 2-block radius of the downward ladder to standing exactly on the ladder block. \\
    \textbf{Assigned Reward:} 1 \\
    \textbf{Success Condition:} \code{(agent level == 0) \& (agent standing on: ItemType.LADDER\_DOWN)} \\
    \textbf{Starting Conditions:}
    \begin{itemize}[noitemsep, topsep=0pt, leftmargin=*]
        \item \textbf{C1:} \code{local object exists: ItemType.LADDER\_DOWN} (Req: FindLadderDown)
    \end{itemize}
\end{tcolorbox}

\begin{tcolorbox}[skillbox, title=FindIron]
    \textbf{Skill Type:} Resource Gathering \\
    \textbf{Description:} The agent explores the overworld (level 0) to locate iron ore deposits within a 2-block radius. \\
    \textbf{Assigned Reward:} 1 \\
    \textbf{Success Condition:} \code{(agent level == 0) \& agent near block: BlockType.IRON} \\
    \textbf{Starting Conditions:}
    \begin{itemize}[noitemsep, topsep=0pt, leftmargin=*]
        \item \textbf{C1:} \code{(agent level == 0) \& agent near block: BlockType.STONE} (Req: FindStone)
    \end{itemize}
\end{tcolorbox}

\begin{tcolorbox}[skillbox, title=DescendLadder]
    \textbf{Skill Type:} Navigation \\
    \textbf{Description:} The agent executes the descend action while positioned on the downward ladder in the overworld to transition to the dungeon. \\
    \textbf{Assigned Reward:} 1 \\
    \textbf{Success Condition:} \code{agent level == 1} \\
    \textbf{Starting Conditions:}
    \begin{itemize}[noitemsep, topsep=0pt, leftmargin=*]
        \item \textbf{C1:} \code{(agent level == 0) \& (agent standing on: ItemType.LADDER\_DOWN)} (Req: StepOntoLadderDown)
    \end{itemize}
\end{tcolorbox}

\begin{tcolorbox}[skillbox, title=FindChestLevel1]
    \textbf{Skill Type:} Navigation \\
    \textbf{Description:} The agent navigates the dungeon level (level 1) to locate a chest within a 2-block radius. \\
    \textbf{Assigned Reward:} 1 \\
    \textbf{Success Condition:} \code{(agent level == 1) \& agent near block: BlockType.CHEST} \\
    \textbf{Starting Conditions:}
    \begin{itemize}[noitemsep, topsep=0pt, leftmargin=*]
        \item \textbf{C1:} \code{agent level == 1} (Req: DescendLadder)
    \end{itemize}
\end{tcolorbox}

\begin{tcolorbox}[skillbox, title=FindOrcSoldierLevel1]
    \textbf{Skill Type:} Navigation \\
    \textbf{Description:} The agent explores the dungeon level (level 1) to locate an Orc Soldier, a melee mob that spawns in dungeon rooms, within a 2-block radius. \\
    \textbf{Assigned Reward:} 1 \\
    \textbf{Success Condition:} \code{(agent level == 1) \& agent near melee mob: MeleeMobType.ORC\_SOLDIER} \\
    \textbf{Starting Conditions:}
    \begin{itemize}[noitemsep, topsep=0pt, leftmargin=*]
        \item \textbf{C1:} \code{agent level == 1} (Req: DescendLadder)
    \end{itemize}
\end{tcolorbox}

\begin{tcolorbox}[skillbox, title=FaceLadderDown]
    \textbf{Skill Type:} Navigation \\
    \textbf{Description:} The agent positions itself adjacent to the downward ladder on the overworld (level 0) and rotates to face the ladder block.  \\
    \textbf{Assigned Reward:} 1 \\
    \textbf{Success Condition:} \code{(agent level == 0) \& (agent facing: ItemType.LADDER\_DOWN)} \\
    \textbf{Starting Conditions:}
    \begin{itemize}[noitemsep, topsep=0pt, leftmargin=*]
        \item \textbf{C1:} \code{(agent level == 0) \& local object exists: ItemType.LADDER\_DOWN} (Req: FindLadderDown)
    \end{itemize}
\end{tcolorbox}

\begin{tcolorbox}[skillbox, title=KillOrcWarrior]
    \textbf{Skill Type:} Fighting Mobs \\
    \textbf{Description:} The agent locates and defeats an Orc Warrior, a melee mob that spawns in dungeon rooms on level 1. \\
    \textbf{Assigned Reward:} 1 \\
    \textbf{Success Condition:} \code{(agent level == 1) \& agent killed mob: MeleeMobType.ORC\_SOLDIER} \\
    \textbf{Starting Conditions:}
    \begin{itemize}[noitemsep, topsep=0pt, leftmargin=*]
        \item \textbf{C1:} \code{agent level == 1} (Req: DescendLadder)
        \item \textbf{C2:} \code{agent inventory sword >= 1} (Req: CraftWoodSword)
        \item \textbf{C3:} \code{(agent level == 1) \& agent near melee mob: MeleeMobType.ORC\_SOLDIER} (Req: FindOrcSoldierLevel1)
    \end{itemize}
\end{tcolorbox}

\begin{tcolorbox}[skillbox, title=OpenChestForBowLevel1]
    \textbf{Skill Type:} Resource Gathering \\
    \textbf{Description:} The agent locates and opens the first chest on the dungeon level (level 1) to obtain a bow. \\
    \textbf{Assigned Reward:} 1 \\
    \textbf{Success Condition:} \code{(agent level == 1) \& (agent inventory bow >= 1)} \\
    \textbf{Starting Conditions:}
    \begin{itemize}[noitemsep, topsep=0pt, leftmargin=*]
        \item \textbf{C1:} \code{agent level == 1} (Req: DescendLadder)
        \item \textbf{C2:} \code{(agent level == 1) \& agent near block: BlockType.CHEST} (Req: FindChestLevel1)
    \end{itemize}
\end{tcolorbox}

\begin{tcolorbox}[skillbox, title=FindFountainLevel1]
    \textbf{Skill Type:} Navigation \\
    \textbf{Description:} The agent leverages its ability to locate dungeon chests to efficiently explore adjacent rooms and identify fountains. \\
    \textbf{Assigned Reward:} 1 \\
    \textbf{Success Condition:} \code{(agent level == 1) \& agent near block: BlockType.FOUNTAIN} \\
    \textbf{Starting Conditions:}
    \begin{itemize}[noitemsep, topsep=0pt, leftmargin=*]
        \item \textbf{C1:} \code{agent level == 1} (Req: DescendLadder)
        \item \textbf{C2:} \code{agent near block: BlockType.CHEST} (Req: FindChestLevel1)
    \end{itemize}
\end{tcolorbox}

\begin{tcolorbox}[skillbox, title=FindDungeonTorch]
    \textbf{Skill Type:} Navigation \\
    \textbf{Description:} The agent explores the dungeon level (level 1) to locate pre-placed torches in room corners, serving as navigation landmarks.  \\
    \textbf{Assigned Reward:} 1 \\
    \textbf{Success Condition:} \code{(agent level == 1) \& stationary object exists: ItemType.TORCH} \\
    \textbf{Starting Conditions:}
    \begin{itemize}[noitemsep, topsep=0pt, leftmargin=*]
        \item \textbf{C1:} \code{agent level == 1} (Req: DescendLadder)
    \end{itemize}
\end{tcolorbox}

\begin{tcolorbox}[skillbox, title=OpenChestForArrowsLevel1]
    \textbf{Skill Type:} Resource Gathering \\
    \textbf{Description:} The agent locates and opens a chest on the dungeon level (level 1) to obtain arrows. \\
    \textbf{Assigned Reward:} 1 \\
    \textbf{Success Condition:} \code{agent inventory arrows >= 1} \\
    \textbf{Starting Conditions:}
    \begin{itemize}[noitemsep, topsep=0pt, leftmargin=*]
        \item \textbf{C1:} \code{agent level == 1} (Req: DescendLadder)
        \item \textbf{C2:} \code{agent near block: BlockType.CHEST} (Req: FindChestLevel1)
    \end{itemize}
\end{tcolorbox}

\begin{tcolorbox}[skillbox, title=DrinkFromFountainLevel1]
    \textbf{Skill Type:} Survivability \\
    \textbf{Description:} The agent locates a fountain in the dungeon level (level 1) and drinks from it to replenish thirst. \\
    \textbf{Assigned Reward:} 1 \\
    \textbf{Success Condition:} \code{(agent level == 1) \& (agent drink >= (7 + 2 * agent dexterity))} \\
    \textbf{Starting Conditions:}
    \begin{itemize}[noitemsep, topsep=0pt, leftmargin=*]
        \item \textbf{C1:} \code{(agent level == 1) \& agent near block: BlockType.FOUNTAIN} (Req: FindFountainLevel1)
    \end{itemize}
\end{tcolorbox}

\begin{tcolorbox}[skillbox, title=PlaceFurnaceAdjacent]
    \textbf{Skill Type:} Crafting \\
    \textbf{Description:} The agent locates the existing crafting table and strategically places a furnace in an adjacent block (including diagonally).  \\
    \textbf{Assigned Reward:} 1 \\
    \textbf{Success Condition:} \code{agent near block: BlockType.CRAFTING\_TABLE \& agent near block: BlockType.FURNACE} \\
    \textbf{Starting Conditions:}
    \begin{itemize}[noitemsep, topsep=0pt, leftmargin=*]
        \item \textbf{C1:} \code{agent near block: BlockType.CRAFTING\_TABLE} (Req: PlaceCraftingTable)
        \item \textbf{C2:} \code{agent inventory stone >= 1} (Req: MineStone)
    \end{itemize}
\end{tcolorbox}

\begin{tcolorbox}[skillbox, title=CraftIronArmour]
    \textbf{Skill Type:} Crafting \\
    \textbf{Description:} The agent crafts a piece of iron armour at a crafting table while adjacent to a furnace using gathered iron and coal resources. \\
    \textbf{Assigned Reward:} 1 \\
    \textbf{Success Condition:} \code{agent inventory armour >= 1} \\
    \textbf{Starting Conditions:}
    \begin{itemize}[noitemsep, topsep=0pt, leftmargin=*]
        \item \textbf{C1:} \code{agent inventory iron >= 3} (Req: MineIron)
        \item \textbf{C2:} \code{agent inventory coal >= 3} (Req: MineCoal)
        \item \textbf{C3:} \code{agent near block: BlockType.CRAFTING\_TABLE} (Req: PlaceCraftingTable)
        \item \textbf{C4:} \code{agent near block: BlockType.FURNACE} (Req: PlaceFurnace)
    \end{itemize}
\end{tcolorbox}

\begin{tcolorbox}[skillbox, title=CraftIronPickaxe]
    \textbf{Skill Type:} Crafting \\
    \textbf{Description:} The agent crafts an iron pickaxe at a crafting table while ensuring the furnace is placed adjacently. \\
    \textbf{Assigned Reward:} 1 \\
    \textbf{Success Condition:} \code{agent inventory pickaxe >= 3} \\
    \textbf{Starting Conditions:}
    \begin{itemize}[noitemsep, topsep=0pt, leftmargin=*]
        \item \textbf{C1-C4:} \code{Wood, Stone, Iron, Coal >= 1}
        \item \textbf{C5:} \code{agent near block: BlockType.CRAFTING\_TABLE \& agent near block: BlockType.FURNACE} (Req: PlaceFurnaceAdjacent)
    \end{itemize}
\end{tcolorbox}

\begin{tcolorbox}[skillbox, title=FindLadderDownLevel1]
    \textbf{Skill Type:} Navigation \\
    \textbf{Description:} The agent explores the dungeon level (level 1) to locate the downward ladder by leveraging dungeon torches as navigation landmarks. \\
    \textbf{Assigned Reward:} 1 \\
    \textbf{Success Condition:} \code{(agent level == 1) \& local object exists: ItemType.LADDER\_DOWN} \\
    \textbf{Starting Conditions:}
    \begin{itemize}[noitemsep, topsep=0pt, leftmargin=*]
        \item \textbf{C1:} \code{agent level == 1} (Req: DescendLadder)
        \item \textbf{C2:} \code{stationary object exists: ItemType.TORCH} (Req: FindDungeonTorch)
    \end{itemize}
\end{tcolorbox}

\begin{tcolorbox}[skillbox, title=CraftIronSword]
    \textbf{Skill Type:} Crafting \\
    \textbf{Description:} The agent crafts an iron sword at a crafting table while ensuring furnace adjacency. \\
    \textbf{Assigned Reward:} 1 \\
    \textbf{Success Condition:} \code{agent inventory sword >= 3} \\
    \textbf{Starting Conditions:}
    \begin{itemize}[noitemsep, topsep=0pt, leftmargin=*]
        \item \textbf{C1-C4:} \code{Wood, Stone, Iron, Coal >= 1}
        \item \textbf{C5:} \code{agent near block: BlockType.CRAFTING\_TABLE \& agent near block: BlockType.FURNACE} (Req: PlaceFurnaceAdjacent)
    \end{itemize}
\end{tcolorbox}

\begin{tcolorbox}[skillbox, title=StepOntoLadderDownLevel1]
    \textbf{Skill Type:} Navigation \\
    \textbf{Description:} The agent navigates from proximity to the downward ladder on the dungeon level (level 1) to stand precisely on the ladder block. \\
    \textbf{Assigned Reward:} 1 \\
    \textbf{Success Condition:} \code{(agent level == 1) \& (agent standing on: ItemType.LADDER\_DOWN)} \\
    \textbf{Starting Conditions:}
    \begin{itemize}[noitemsep, topsep=0pt, leftmargin=*]
        \item \textbf{C1:} \code{(agent level == 1) \& local object exists: ItemType.LADDER\_DOWN} (Req: FindLadderDownLevel1)
    \end{itemize}
\end{tcolorbox}

\begin{tcolorbox}[skillbox, title=FindUpLadderLevel1]
    \textbf{Skill Type:} Navigation \\
    \textbf{Description:} The agent explores the starting room on the dungeon level (level 1) to locate the upward ladder. \\
    \textbf{Assigned Reward:} 1 \\
    \textbf{Success Condition:} \code{(agent level == 1) \& ladder object exists: ItemType.LADDER\_UP} \\
    \textbf{Starting Conditions:}
    \begin{itemize}[noitemsep, topsep=0pt, leftmargin=*]
        \item \textbf{C1:} \code{agent level == 1} (Req: DescendLadder)
    \end{itemize}
\end{tcolorbox}

\begin{tcolorbox}[skillbox, title=MineGrassForSapling]
    \textbf{Skill Type:} Resource Gathering \\
    \textbf{Description:} The agent mines grass blocks in the overworld (level 0) to obtain saplings, a critical resource for establishing sustainable food sources. \\
    \textbf{Assigned Reward:} 1 \\
    \textbf{Success Condition:} \code{agent inventory sapling >= 1} \\
    \textbf{Starting Conditions:}
    \begin{itemize}[noitemsep, topsep=0pt, leftmargin=*]
        \item \textbf{C1:} \code{(agent level == 0) \& agent near block: BlockType.GRASS} (Req: FindTree)
    \end{itemize}
\end{tcolorbox}

\begin{tcolorbox}[skillbox, title=KillOrcMage]
    \textbf{Skill Type:} Fighting mobs \\
    \textbf{Description:} The agent locates and defeats an Orc Mage, a ranged mob that spawns in dungeon rooms on level 1.  \\
    \textbf{Assigned Reward:} 1 \\
    \textbf{Success Condition:} \code{(agent level == 1) \& agent killed mob: RangedMobType.ORC\_MAGE} \\
    \textbf{Starting Conditions:}
    \begin{itemize}[noitemsep, topsep=0pt, leftmargin=*]
        \item \textbf{C1:} \code{agent level == 1} (Req: DescendLadder)
        \item \textbf{C2:} \code{agent inventory bow >= 1} (Req: OpenChestForBowLevel1)
        \item \textbf{C3:} \code{agent inventory arrows >= 1} (Req: CraftArrow)
        \item \textbf{C4:} \code{agent near ranged mob: RangedMobType.ORC\_MAGE} (Req: FindDungeonTorch)
    \end{itemize}
\end{tcolorbox}

\begin{tcolorbox}[skillbox, title=ExitRoomToCorridorLevel1]
    \textbf{Skill Type:} Navigation \\
    \textbf{Description:} The agent learns to exit a dungeon room by moving away from corner torches into the connecting corridor.  \\
    \textbf{Assigned Reward:} 1 \\
    \textbf{Success Condition:} \code{not stationary object exists: ItemType.TORCH} \\
    \textbf{Starting Conditions:}
    \begin{itemize}[noitemsep, topsep=0pt, leftmargin=*]
        \item \textbf{C1:} \code{stationary object exists: ItemType.TORCH} (Req: FindDungeonTorch)
    \end{itemize}
\end{tcolorbox}

\begin{tcolorbox}[skillbox, title=EnterAdjacentRoomLevel1]
    \textbf{Skill Type:} Navigation \\
    \textbf{Description:} The agent navigates from the starting room (containing the upward ladder) on the dungeon level to an adjacent room using torch landmarks. \\
    \textbf{Assigned Reward:} 1 \\
    \textbf{Success Condition:} \code{(agent level == 1) \& stationary object exists: ItemType.TORCH \& not ladder object exists: ItemType.LADDER\_UP} \\
    \textbf{Starting Conditions:}
    \begin{itemize}[noitemsep, topsep=0pt, leftmargin=*]
        \item \textbf{C1:} \code{(agent level == 1) \& ladder object exists: ItemType.LADDER\_UP} (Req: FindUpLadderLevel1)
    \end{itemize}
\end{tcolorbox}

\begin{tcolorbox}[skillbox, title=EnterNewRoomFromCorridorLevel1]
    \textbf{Skill Type:} Navigation \\
    \textbf{Description:} The agent navigates from the dungeon corridor (after exiting a room) on level 1 to enter an adjacent room by detecting torches. \\
    \textbf{Assigned Reward:} 1 \\
    \textbf{Success Condition:} \code{stationary object exists: ItemType.TORCH} \\
    \textbf{Starting Conditions:}
    \begin{itemize}[noitemsep, topsep=0pt, leftmargin=*]
        \item \textbf{C1:} \code{not stationary object exists: ItemType.TORCH} (Req: ExitRoomToCorridorLevel1)
    \end{itemize}
\end{tcolorbox}

\begin{tcolorbox}[skillbox, title=FindSnailLevel1]
    \textbf{Skill Type:} Navigation \\
    \textbf{Description:} The agent explores the dungeon level (level 1) to locate a snail, a passive mob that spawns in dungeon rooms. \\
    \textbf{Assigned Reward:} 1 \\
    \textbf{Success Condition:} \code{(agent level == 1) \& agent near passive mob: PassiveMobType.SNAIL} \\
    \textbf{Starting Conditions:}
    \begin{itemize}[noitemsep, topsep=0pt, leftmargin=*]
        \item \textbf{C1:} \code{agent level == 1} (Req: DescendLadder)
        \item \textbf{C2:} \code{stationary object exists: ItemType.TORCH} (Req: FindDungeonTorch)
    \end{itemize}
\end{tcolorbox}

\begin{tcolorbox}[skillbox, title=KillSnailLevel1]
    \textbf{Skill Type:} Fighting mobs \\
    \textbf{Description:} The agent locates and defeats a snail, a passive mob that spawns in dungeon rooms on level 1, to obtain food resources. \\
    \textbf{Assigned Reward:} 1 \\
    \textbf{Success Condition:} \code{(agent level == 1) \& agent killed mob: PassiveMobType.SNAIL} \\
    \textbf{Starting Conditions:}
    \begin{itemize}[noitemsep, topsep=0pt, leftmargin=*]
        \item \textbf{C1:} \code{(agent level == 1)} (Req: DescendLadder)
        \item \textbf{C2:} \code{agent inventory sword >= 1} (Req: CraftWoodSword)
        \item \textbf{C3:} \code{agent near passive mob: PassiveMobType.SNAIL} (Req: FindFountainLevel1)
    \end{itemize}
\end{tcolorbox}

\begin{tcolorbox}[skillbox, title=EnterNewRoomExcludingUpLadderLevel1]
    \textbf{Skill Type:} Navigation \\
    \textbf{Description:} The agent navigates from the corridor to enter an adjacent room that does not contain the upward ladder, confirming new room entry. \\
    \textbf{Assigned Reward:} 1 \\
    \textbf{Success Condition:} \code{(agent level == 1) \& stationary object exists: ItemType.TORCH \& not ladder object exists: ItemType.LADDER\_UP} \\
    \textbf{Starting Conditions:}
    \begin{itemize}[noitemsep, topsep=0pt, leftmargin=*]
        \item \textbf{C1:} \code{not stationary object exists: ItemType.TORCH} (Req: ExitRoomToCorridorLevel1)
    \end{itemize}
\end{tcolorbox}

\begin{tcolorbox}[skillbox, title=KillEightMobsLevel1]
    \textbf{Skill Type:} Fighting mobs \\
    \textbf{Description:} The agent eliminates 8 hostile or passive mobs to satisfy the requirement for opening the downward ladder to level 2. \\
    \textbf{Assigned Reward:} 1 \\
    \textbf{Success Condition:} \code{(agent level == 1) \& (agent monsters killed level 1 >= 8)} \\
    \textbf{Starting Conditions:}
    \begin{itemize}[noitemsep, topsep=0pt, leftmargin=*]
        \item \textbf{C1:} \code{agent level == 1} (Req: DescendLadder)
        \item \textbf{C2:} \code{agent inventory sword >= 1} (Req: CraftWoodSword)
    \end{itemize}
\end{tcolorbox}

\begin{tcolorbox}[skillbox, title=FaceLadderDownLevel1]
    \textbf{Skill Type:} Navigation \\
    \textbf{Description:} The agent positions itself adjacent to the downward ladder on the dungeon level and rotates to face the ladder block. \\
    \textbf{Assigned Reward:} 1 \\
    \textbf{Success Condition:} \code{(agent level == 1) \& (agent facing: ItemType.LADDER\_DOWN)} \\
    \textbf{Starting Conditions:}
    \begin{itemize}[noitemsep, topsep=0pt, leftmargin=*]
        \item \textbf{C1:} \code{(agent level == 1) \& local object exists: ItemType.LADDER\_DOWN} (Req: FindLadderDownLevel1)
    \end{itemize}
\end{tcolorbox}

\begin{tcolorbox}[skillbox, title=ApproachLadderDownLevel1]
    \textbf{Skill Type:} Navigation \\
    \textbf{Description:} The agent navigates from proximity to the downward ladder to a position directly adjacent (1 block away). \\
    \textbf{Assigned Reward:} 1 \\
    \textbf{Success Condition:} \code{(agent level == 1) \& (agent adjacent to: ItemType.LADDER\_DOWN)} \\
    \textbf{Starting Conditions:}
    \begin{itemize}[noitemsep, topsep=0pt, leftmargin=*]
        \item \textbf{C1:} \code{(agent level == 1) \& local object exists: ItemType.LADDER\_DOWN} (Req: FindLadderDownLevel1)
    \end{itemize}
\end{tcolorbox}

\begin{tcolorbox}[skillbox, title=DescendLadderLevel1]
    \textbf{Skill Type:} Navigation \\
    \textbf{Description:} The agent executes the descend action while positioned on the downward ladder to transition to the Gnomish Mines (level 2).  \\
    \textbf{Assigned Reward:} 1 \\
    \textbf{Success Condition:} \code{agent level == 2} \\
    \textbf{Starting Conditions:}
    \begin{itemize}[noitemsep, topsep=0pt, leftmargin=*]
        \item \textbf{C1:} \code{(agent level == 1) \& (agent monsters killed level 1 >= 8)} (Req: KillEightMobsLevel1)
        \item \textbf{C2:} \code{(agent level == 1) \& (agent standing on: ItemType.LADDER\_DOWN)} (Req: StepOntoLadderDownLevel1)
    \end{itemize}
\end{tcolorbox}

\begin{tcolorbox}[skillbox, title=FindWaterLevel2]
    \textbf{Skill Type:} Navigation \\
    \textbf{Description:} The agent navigates the Gnomish Mines level (level 2) to locate a water pool within a 2-block radius. \\
    \textbf{Assigned Reward:} 1 \\
    \textbf{Success Condition:} \code{(agent level == 2) \& agent near block: BlockType.WATER} \\
    \textbf{Starting Conditions:}
    \begin{itemize}[noitemsep, topsep=0pt, leftmargin=*]
        \item \textbf{C1:} \code{(agent level == 2)} (Req: DescendLadderLevel1)
    \end{itemize}
\end{tcolorbox}

\begin{tcolorbox}[skillbox, title=FindUpLadderLevel2]
    \textbf{Skill Type:} Navigation \\
    \textbf{Description:} The agent explores the Gnomish Mines level (level 2) to locate the upward ladder (leading back to level 1). \\
    \textbf{Assigned Reward:} 1 \\
    \textbf{Success Condition:} \code{(agent level == 2) \& ladder object exists: ItemType.LADDER\_UP} \\
    \textbf{Starting Conditions:}
    \begin{itemize}[noitemsep, topsep=0pt, leftmargin=*]
        \item \textbf{C1:} \code{agent level == 2} (Req: DescendLadderLevel1)
    \end{itemize}
\end{tcolorbox}

\begin{tcolorbox}[skillbox, title=FindWaterWithTorchLevel2]
    \textbf{Skill Type:} Navigation \\
    \textbf{Description:} The agent leverages placed torches to illuminate the dark Gnomish Mines and locates water pools within a 2-block radius. \\
    \textbf{Assigned Reward:} 1 \\
    \textbf{Success Condition:} \code{(agent level == 2) \& agent near block: BlockType.WATER} \\
    \textbf{Starting Conditions:}
    \begin{itemize}[noitemsep, topsep=0pt, leftmargin=*]
        \item \textbf{C1:} \code{agent level == 2} (Req: DescendLadderLevel1)
        \item \textbf{C2:} \code{stationary object exists: ItemType.TORCH} (Req: PlaceTorch)
    \end{itemize}
\end{tcolorbox}

\begin{tcolorbox}[skillbox, title=MoveToEdgeOfLightLevel2]
    \textbf{Skill Type:} Navigation \\
    \textbf{Description:} The agent moves from the spawn point on level 2 to a position where the light level drops below 0.5.  \\
    \textbf{Assigned Reward:} 1 \\
    \textbf{Success Condition:} \code{(agent level == 2) \& (agent light level < 0.5)} \\
    \textbf{Starting Conditions:}
    \begin{itemize}[noitemsep, topsep=0pt, leftmargin=*]
        \item \textbf{C1:} \code{agent level == 2} (Req: DescendLadderLevel1)
    \end{itemize}
\end{tcolorbox}

\begin{tcolorbox}[skillbox, title=FindStalagmiteLevel2]
    \textbf{Skill Type:} Navigation \\
    \textbf{Description:} The agent explores the Gnomish Mines level (level 2) to locate stalagmites within a 2-block radius. \\
    \textbf{Assigned Reward:} 1 \\
    \textbf{Success Condition:} \code{(agent level == 2) \& agent near block: BlockType.STALAGMITE} \\
    \textbf{Starting Conditions:}
    \begin{itemize}[noitemsep, topsep=0pt, leftmargin=*]
        \item \textbf{C1:} \code{agent level == 2} (Req: DescendLadderLevel1)
    \end{itemize}
\end{tcolorbox}

\begin{tcolorbox}[skillbox, title=MineStalagmiteLevel2]
    \textbf{Skill Type:} Resource Gathering \\
    \textbf{Description:} The agent mines stalagmites in the Gnomish Mines (level 2) to obtain stone resources. \\
    \textbf{Assigned Reward:} 1 \\
    \textbf{Success Condition:} \code{agent inventory stone >= 1} \\
    \textbf{Starting Conditions:}
    \begin{itemize}[noitemsep, topsep=0pt, leftmargin=*]
        \item \textbf{C1:} \code{agent level == 2} (Req: DescendLadderLevel1)
        \item \textbf{C2:} \code{stationary object exists: ItemType.TORCH} (Req: PlaceTorch)
        \item \textbf{C3:} \code{agent inventory pickaxe >= 1} (Req: CraftWoodPickaxe)
    \end{itemize}
\end{tcolorbox}

\begin{tcolorbox}[skillbox, title=NavigateToWaterFromStalagmiteLevel2]
    \textbf{Skill Type:} Navigation \\
    \textbf{Description:} The agent leverages stalagmites as environmental landmarks in the Gnomish Mines to navigate toward nearby water pools. \\
    \textbf{Assigned Reward:} 1 \\
    \textbf{Success Condition:} \code{(agent level == 2) \& agent near block: BlockType.WATER} \\
    \textbf{Starting Conditions:}
    \begin{itemize}[noitemsep, topsep=0pt, leftmargin=*]
        \item \textbf{C1:} \code{(agent level == 2) \& agent near block: BlockType.STALAGMITE} (Req: FindStalagmiteLevel2)
        \item \textbf{C2:} \code{stationary object exists: ItemType.TORCH} (Req: PlaceTorch)
    \end{itemize}
\end{tcolorbox}

\begin{tcolorbox}[skillbox, title=PlaceTorchAtEdgeLevel2]
    \textbf{Skill Type:} Navigation \\
    \textbf{Description:} The agent strategically places a torch at the boundary of existing light (light level < 0.5) in the Gnomish Mines. \\
    \textbf{Assigned Reward:} 1 \\
    \textbf{Success Condition:} \code{(agent level == 2) \& (agent standing on: ItemType.TORCH)} \\
    \textbf{Starting Conditions:}
    \begin{itemize}[noitemsep, topsep=0pt, leftmargin=*]
        \item \textbf{C1:} \code{(agent level == 2) \& (agent light level < 0.5)} (Req: MoveToEdgeOfLightLevel2)
        \item \textbf{C2:} \code{agent inventory torches >= 1} (Req: CraftTorch)
    \end{itemize}
\end{tcolorbox}

\begin{tcolorbox}[skillbox, title=MoveIntoLitAreaLevel2]
    \textbf{Skill Type:} Survivability \\
    \textbf{Description:} The agent moves from the torch placement position at the edge of darkness into the newly illuminated area. \\
    \textbf{Assigned Reward:} 1 \\
    \textbf{Success Condition:} \code{(agent level == 2) \& (agent light level >= 0.5)} \\
    \textbf{Starting Conditions:}
    \begin{itemize}[noitemsep, topsep=0pt, leftmargin=*]
        \item \textbf{C1:} \code{(agent level == 2) \& (agent light level < 0.5)} (Req: MoveToEdgeOfLightLevel2)
        \item \textbf{C2:} \code{(agent level == 2) \& stationary object exists: ItemType.TORCH} (Req: PlaceTorchAtEdgeLevel2)
    \end{itemize}
\end{tcolorbox}

\begin{tcolorbox}[skillbox, title=MoveToNextEdgeOfLightLevel2]
    \textbf{Skill Type:} Navigation \\
    \textbf{Description:} The agent moves from the center of a newly created lit area to the next edge of darkness, enabling progressive exploration. \\
    \textbf{Assigned Reward:} 1 \\
    \textbf{Success Condition:} \code{(agent level == 2) \& (agent light level < 0.5)} \\
    \textbf{Starting Conditions:}
    \begin{itemize}[noitemsep, topsep=0pt, leftmargin=*]
        \item \textbf{C1:} \code{agent level == 2} (Req: DescendLadderLevel1)
        \item \textbf{C2:} \code{(agent level == 2) \& (agent light level >= 0.5)} (Req: MoveIntoLitAreaLevel2)
    \end{itemize}
\end{tcolorbox}

\begin{tcolorbox}[skillbox, title=MoveAwayFromUpLadderLevel2]
    \textbf{Skill Type:} Navigation \\
    \textbf{Description:} The agent navigates away from the upward ladder spawn point in the Gnomish Mines to a position at least 5 blocks distant. \\
    \textbf{Assigned Reward:} 1 \\
    \textbf{Success Condition:} \code{(agent level == 2) \& (distance to up ladder >= 5)} \\
    \textbf{Starting Conditions:}
    \begin{itemize}[noitemsep, topsep=0pt, leftmargin=*]
        \item \textbf{C1:} \code{(agent level == 2) \& ladder object exists: ItemType.LADDER\_UP} (Req: FindUpLadderLevel2)
    \end{itemize}
\end{tcolorbox}

\begin{tcolorbox}[skillbox, title=FindBatLevel2]
    \textbf{Skill Type:} Navigation \\
    \textbf{Description:} The agent explores the Gnomish Mines level (level 2) by leveraging water pools (located via torches) to find bats, passive mobs that spawn near water sources. \\
    \textbf{Assigned Reward:} 1 \\
    \textbf{Success Condition:} \code{(agent level == 2) \& agent near passive mob: PassiveMobType.BAT} \\
    \textbf{Starting Conditions:}
    \begin{itemize}[noitemsep, topsep=0pt, leftmargin=*]
        \item \textbf{C1:} \code{agent level == 2} (Req: DescendLadderLevel1)
        \item \textbf{C2:} \code{(agent level == 2) \& agent near block: BlockType.WATER} (Req: FindWaterWithTorchLevel2)
    \end{itemize}
\end{tcolorbox}

\begin{tcolorbox}[skillbox, title=MineCoalNearWaterLevel2]
    \textbf{Skill Type:} Resource Gathering \\
    \textbf{Description:} The agent leverages water pools as reliable landmarks in the Gnomish Mines to mine coal deposits within stone-rich areas.  \\
    \textbf{Assigned Reward:} 1 \\
    \textbf{Success Condition:} \code{(agent level == 2) \& (agent inventory coal >= 1)} \\
    \textbf{Starting Conditions:}
    \begin{itemize}[noitemsep, topsep=0pt, leftmargin=*]
        \item \textbf{C1:} \code{agent inventory pickaxe >= 1} (Req: CraftWoodPickaxe)
        \item \textbf{C2:} \code{stationary object exists: ItemType.TORCH} (Req: PlaceTorch)
        \item \textbf{C3:} \code{(agent level == 2) \& agent near block: BlockType.WATER} (Req: NavigateToWaterFromStalagmiteLevel2)
    \end{itemize}
\end{tcolorbox}

\begin{tcolorbox}[skillbox, title=KillGnomeWarriorNearUpLadderLevel2]
    \textbf{Skill Type:} Fighting mobs \\
    \textbf{Description:} The agent remains within the naturally lit area around the upward ladder spawn point and eliminates a Gnome Warrior. \\
    \textbf{Assigned Reward:} 1 \\
    \textbf{Success Condition:} \code{(agent level == 2) \& agent killed mob: MeleeMobType.GNOME\_WARRIOR} \\
    \textbf{Starting Conditions:}
    \begin{itemize}[noitemsep, topsep=0pt, leftmargin=*]
        \item \textbf{C1:} \code{agent level == 2} (Req: DescendLadderLevel1)
        \item \textbf{C2:} \code{(agent level == 2) \& ladder object exists: ItemType.LADDER\_UP} (Req: FindUpLadderLevel2)
        \item \textbf{C3:} \code{agent inventory sword >= 1} (Req: CraftWoodSword)
    \end{itemize}
\end{tcolorbox}

\begin{tcolorbox}[skillbox, title=KillBatLevel2]
    \textbf{Skill Type:} Fighting mobs \\
    \textbf{Description:} The agent locates and defeats a bat, a passive mob that spawns near water pools in the Gnomish Mines level (level 2), to obtain food resources. \\
    \textbf{Assigned Reward:} 1 \\
    \textbf{Success Condition:} \code{(agent level == 2) \& agent killed mob: PassiveMobType.BAT} \\
    \textbf{Starting Conditions:}
    \begin{itemize}[noitemsep, topsep=0pt, leftmargin=*]
        \item \textbf{C1:} \code{agent level == 2} (Req: DescendLadderLevel1)
        \item \textbf{C2:} \code{agent inventory sword >= 1} (Req: CraftWoodSword)
        \item \textbf{C3:} \code{stationary object exists: ItemType.TORCH} (Req: PlaceTorch)
        \item \textbf{C4:} \code{agent near passive mob: PassiveMobType.BAT} (Req: FindBatLevel2)
    \end{itemize}
\end{tcolorbox}

\begin{tcolorbox}[skillbox, title=CheckLadderInLitAreaLevel2]
    \textbf{Skill Type:} Navigation \\
    \textbf{Description:} After moving into a newly illuminated area in the Gnomish Mines, the agent systematically checks for the downward ladder. \\
    \textbf{Assigned Reward:} 1 \\
    \textbf{Success Condition:} \code{(agent level == 2) \& local object exists: ItemType.LADDER\_DOWN} \\
    \textbf{Starting Conditions:}
    \begin{itemize}[noitemsep, topsep=0pt, leftmargin=*]
        \item \textbf{C1:} \code{(agent level == 2) \& (agent light level >= 0.5)} (Req: MoveIntoLitAreaLevel2)
    \end{itemize}
\end{tcolorbox}

\begin{tcolorbox}[skillbox, title=DrinkWaterLevel2]
    \textbf{Skill Type:} Survivability \\
    \textbf{Description:} The agent locates a water pool in the Gnomish Mines level (level 2) and drinks from it to replenish thirst. \\
    \textbf{Assigned Reward:} 1 \\
    \textbf{Success Condition:} \code{(agent level == 2) \& (agent drink >= (7 + 2 * agent dexterity))} \\
    \textbf{Starting Conditions:}
    \begin{itemize}[noitemsep, topsep=0pt, leftmargin=*]
        \item \textbf{C1:} \code{(agent level == 2) \& agent near block: BlockType.WATER} (Req: NavigateToWaterFromStalagmiteLevel2)
    \end{itemize}
\end{tcolorbox}

\begin{tcolorbox}[skillbox, title=ApproachWaterLevel2]
    \textbf{Skill Type:} Survivability \\
    \textbf{Description:} The agent moves from proximity (within 2 blocks) to a water pool in the Gnomish Mines to an adjacent position (1 block away). \\
    \textbf{Assigned Reward:} 1 \\
    \textbf{Success Condition:} \code{(agent level == 2) \& (agent adjacent to: BlockType.WATER)} \\
    \textbf{Starting Conditions:}
    \begin{itemize}[noitemsep, topsep=0pt, leftmargin=*]
        \item \textbf{C1:} \code{(agent level == 2) \& agent near block: BlockType.WATER} (Req: NavigateToWaterFromStalagmiteLevel2)
    \end{itemize}
\end{tcolorbox}

\begin{tcolorbox}[skillbox, title=CheckLadderAfterTorchLevel2]
    \textbf{Skill Type:} Navigation \\
    \textbf{Description:} After placing a torch at the edge of darkness, the agent immediately checks the full 11x11 illuminated area for the downward ladder.  \\
    \textbf{Assigned Reward:} 1 \\
    \textbf{Success Condition:} \code{(agent level == 2) \& stationary object exists: ItemType.LADDER\_DOWN} \\
    \textbf{Starting Conditions:}
    \begin{itemize}[noitemsep, topsep=0pt, leftmargin=*]
        \item \textbf{C1:} \code{(agent level == 2) \& stationary object exists: ItemType.TORCH} (Req: PlaceTorchAtEdgeLevel2)
    \end{itemize}
\end{tcolorbox}

\begin{tcolorbox}[skillbox, title=ApproachLadderDownLevel2]
    \textbf{Skill Type:} Navigation \\
    \textbf{Description:} The agent moves from proximity to the downward ladder on the Gnomish Mines level (level 2) to a position directly adjacent. \\
    \textbf{Assigned Reward:} 1 \\
    \textbf{Success Condition:} \code{(agent level == 2) \& (agent adjacent to: ItemType.LADDER\_DOWN)} \\
    \textbf{Starting Conditions:}
    \begin{itemize}[noitemsep, topsep=0pt, leftmargin=*]
        \item \textbf{C1:} \code{(agent level == 2) \& local object exists: ItemType.LADDER\_DOWN} (Req: CheckLadderInLitAreaLevel2)
    \end{itemize}
\end{tcolorbox}

\begin{tcolorbox}[skillbox, title=CheckLadderInFullLitAreaLevel2]
    \textbf{Skill Type:} Navigation \\
    \textbf{Description:} After moving into a newly illuminated area, the agent performs a comprehensive 10-block radius scan to detect the downward ladder. \\
    \textbf{Assigned Reward:} 1 \\
    \textbf{Success Condition:} \code{(agent level == 2) \& stationary object exists: ItemType.LADDER\_DOWN} \\
    \textbf{Starting Conditions:}
    \begin{itemize}[noitemsep, topsep=0pt, leftmargin=*]
        \item \textbf{C1:} \code{agent level == 2} (Req: DescendLadderLevel1)
        \item \textbf{C2:} \code{(agent level == 2) \& (agent light level >= 0.5)} (Req: MoveIntoLitAreaLevel2)
    \end{itemize}
\end{tcolorbox}

\begin{tcolorbox}[skillbox, title=KillEightMobsLevel2]
    \textbf{Skill Type:} Fighting mobs \\
    \textbf{Description:} The agent explores the Gnomish Mines level (level 2) while maintaining position in lit areas to eliminate 8 hostile or passive mobs. \\
    \textbf{Assigned Reward:} 1 \\
    \textbf{Success Condition:} \code{(agent level == 2) \& (agent monsters killed level 2 >= 8)} \\
    \textbf{Starting Conditions:}
    \begin{itemize}[noitemsep, topsep=0pt, leftmargin=*]
        \item \textbf{C1:} \code{agent level == 2} (Req: DescendLadderLevel1)
        \item \textbf{C2:} \code{agent inventory sword >= 1} (Req: CraftWoodSword)
        \item \textbf{C3:} \code{(agent level == 2) \& (agent light level >= 0.5)} (Req: MoveIntoLitAreaLevel2)
    \end{itemize}
\end{tcolorbox}

\begin{tcolorbox}[skillbox, title=MineCoalLevel2]
    \textbf{Skill Type:} Resource Gathering \\
    \textbf{Description:} The agent locates and mines coal deposits in the Gnomish Mines (level 2) using a wooden pickaxe to obtain coal resources. \\
    \textbf{Assigned Reward:} 1 \\
    \textbf{Success Condition:} \code{(agent level == 2) \& (agent inventory coal > prev inventory coal)} \\
    \textbf{Starting Conditions:}
    \begin{itemize}[noitemsep, topsep=0pt, leftmargin=*]
        \item \textbf{C1:} \code{agent level == 2} (Req: DescendLadderLevel1)
        \item \textbf{C2:} \code{agent inventory pickaxe >= 1} (Req: CraftWoodPickaxe)
        \item \textbf{C3:} \code{(agent level == 2) \& agent near block: BlockType.COAL} (Req: FindStalagmiteLevel2)
    \end{itemize}
\end{tcolorbox}

\begin{tcolorbox}[skillbox, title=FaceLadderDownLevel2]
    \textbf{Skill Type:} Navigation \\
    \textbf{Description:} The agent positions itself adjacent to the downward ladder on the Gnomish Mines level (level 2) and rotates to face the ladder block. \\
    \textbf{Assigned Reward:} 1 \\
    \textbf{Success Condition:} \code{(agent level == 2) \& (agent facing: ItemType.LADDER\_DOWN)} \\
    \textbf{Starting Conditions:}
    \begin{itemize}[noitemsep, topsep=0pt, leftmargin=*]
        \item \textbf{C1:} \code{(agent level == 2) \& (agent adjacent to: ItemType.LADDER\_DOWN)} (Req: ApproachLadderDownLevel2)
    \end{itemize}
\end{tcolorbox}

\begin{tcolorbox}[skillbox, title=FaceWaterLevel2]
    \textbf{Skill Type:} Survivability \\
    \textbf{Description:} The agent, when adjacent to a water pool in the Gnomish Mines level (level 2), rotates to face the water block.  \\
    \textbf{Assigned Reward:} 1 \\
    \textbf{Success Condition:} \code{(agent level == 2) \& (agent facing: BlockType.WATER)} \\
    \textbf{Starting Conditions:}
    \begin{itemize}[noitemsep, topsep=0pt, leftmargin=*]
        \item \textbf{C1:} \code{(agent level == 2) \& (agent adjacent to: BlockType.WATER)} (Req: ApproachWaterLevel2)
    \end{itemize}
\end{tcolorbox}

\begin{tcolorbox}[skillbox, title=ApproachUpLadderLevel2]
    \textbf{Skill Type:} Navigation \\
    \textbf{Description:} The agent moves from proximity to the upward ladder on the Gnomish Mines level (level 2) to a position directly adjacent. \\
    \textbf{Assigned Reward:} 1 \\
    \textbf{Success Condition:} \code{(agent level == 2) \& (agent adjacent to: ItemType.LADDER\_UP)} \\
    \textbf{Starting Conditions:}
    \begin{itemize}[noitemsep, topsep=0pt, leftmargin=*]
        \item \textbf{C1:} \code{(agent level == 2) \& ladder object exists: ItemType.LADDER\_UP} (Req: FindUpLadderLevel2)
    \end{itemize}
\end{tcolorbox}

\begin{tcolorbox}[skillbox, title=MoveForwardAfterTorchLevel2]
    \textbf{Skill Type:} Navigation \\
    \textbf{Description:} After placing a torch at the edge of darkness, the agent moves forward one block into the newly illuminated area. \\
    \textbf{Assigned Reward:} 1 \\
    \textbf{Success Condition:} \code{(agent level == 2) \& (agent light level >= 0.5)} \\
    \textbf{Starting Conditions:}
    \begin{itemize}[noitemsep, topsep=0pt, leftmargin=*]
        \item \textbf{C1:} \code{(agent level == 2) \& (agent light level < 0.5)} (Req: PlaceTorchAtEdgeLevel2)
    \end{itemize}
\end{tcolorbox}

\begin{tcolorbox}[skillbox, title=RecoverManaSafe]
    \textbf{Skill Type:} Survivability \\
    \textbf{Description:} The agent ensures survival intrinsics are fully replenished via SleepToFullEnergy, then actively verifies and eliminates threats before recovering mana. \\
    \textbf{Assigned Reward:} 1 \\
    \textbf{Success Condition:} \code{(agent level == 0) \& (agent mana >= (6 + agent intelligence))} \\
    \textbf{Starting Conditions:}
    \begin{itemize}[noitemsep, topsep=0pt, leftmargin=*]
        \item \textbf{C1:} \code{(agent level == 0) \& (agent energy >= (7 + 2 * agent dexterity))} (Req: SleepToFullEnergy)
        \item \textbf{C2:} \code{not agent near melee mob: MeleeMobType.ZOMBIE} (Req: KillZombie)
        \item \textbf{C3:} \code{not agent near ranged mob: RangedMobType.SKELETON} (Req: KillSkeleton)
    \end{itemize}
\end{tcolorbox}

\begin{tcolorbox}[skillbox, title=StepOntoUpLadderLevel2]
    \textbf{Skill Type:} Navigation \\
    \textbf{Description:} The agent moves from an adjacent position to the upward ladder on the Gnomish Mines level (level 2) to stand exactly on the ladder block. \\
    \textbf{Assigned Reward:} 1 \\
    \textbf{Success Condition:} \code{(agent level == 2) \& (agent standing on: ItemType.LADDER\_UP)} \\
    \textbf{Starting Conditions:}
    \begin{itemize}[noitemsep, topsep=0pt, leftmargin=*]
        \item \textbf{C1:} \code{(agent level == 2) \& (agent adjacent to: ItemType.LADDER\_UP)} (Req: ApproachUpLadderLevel2)
    \end{itemize}
\end{tcolorbox}

\begin{tcolorbox}[skillbox, title=FaceUpLadderLevel2]
    \textbf{Skill Type:} Navigation \\
    \textbf{Description:} The agent positions itself adjacent to the upward ladder on the Gnomish Mines level (level 2) and rotates to face the ladder block. \\
    \textbf{Assigned Reward:} 1 \\
    \textbf{Success Condition:} \code{(agent level == 2) \& (agent facing: ItemType.LADDER\_UP)} \\
    \textbf{Starting Conditions:}
    \begin{itemize}[noitemsep, topsep=0pt, leftmargin=*]
        \item \textbf{C1:} \code{(agent level == 2) \& (agent adjacent to: ItemType.LADDER\_UP)} (Req: ApproachUpLadderLevel2)
    \end{itemize}
\end{tcolorbox}

\begin{tcolorbox}[skillbox, title=AscendLadderLevel2]
    \textbf{Skill Type:} Navigation \\
    \textbf{Description:} The agent executes the ascend action while positioned on the upward ladder in the Gnomish Mines to transition back to the dungeon level.  \\
    \textbf{Assigned Reward:} 1 \\
    \textbf{Success Condition:} \code{agent level == 1} \\
    \textbf{Starting Conditions:}
    \begin{itemize}[noitemsep, topsep=0pt, leftmargin=*]
        \item \textbf{C1:} \code{(agent level == 2) \& (agent standing on: ItemType.LADDER\_UP)} (Req: StepOntoUpLadderLevel2)
    \end{itemize}
\end{tcolorbox}

\begin{tcolorbox}[skillbox, title=ApproachFountainLevel1]
    \textbf{Skill Type:} Navigation \\
    \textbf{Description:} The agent moves from proximity (within 2 blocks) to a fountain in the dungeon level to an adjacent position (1 block away). \\
    \textbf{Assigned Reward:} 1 \\
    \textbf{Success Condition:} \code{(agent level == 1) \& (agent adjacent to: BlockType.FOUNTAIN)} \\
    \textbf{Starting Conditions:}
    \begin{itemize}[noitemsep, topsep=0pt, leftmargin=*]
        \item \textbf{C1:} \code{(agent level == 1) \& agent near block: BlockType.FOUNTAIN} (Req: FindFountainLevel1)
    \end{itemize}
\end{tcolorbox}

\begin{tcolorbox}[skillbox, title=StepOntoLadderDownLevel2]
    \textbf{Skill Type:} Navigation \\
    \textbf{Description:} The agent moves forward from an adjacent position while facing the downward ladder on the Gnomish Mines level to stand exactly on the ladder block. \\
    \textbf{Assigned Reward:} 1 \\
    \textbf{Success Condition:} \code{(agent level == 2) \& (agent standing on: ItemType.LADDER\_DOWN)} \\
    \textbf{Starting Conditions:}
    \begin{itemize}[noitemsep, topsep=0pt, leftmargin=*]
        \item \textbf{C1:} \code{(agent level == 2) \& (agent facing: ItemType.LADDER\_DOWN)} (Req: FaceLadderDownLevel2)
    \end{itemize}
\end{tcolorbox}

\begin{tcolorbox}[skillbox, title=DrinkWaterLevel2Improved]
    \textbf{Skill Type:} Survivability \\
    \textbf{Description:} The agent, after precisely approaching and correctly facing a water pool in the Gnomish Mines, drinks from it to replenish thirst.  \\
    \textbf{Assigned Reward:} 1 \\
    \textbf{Success Condition:} \code{(agent level == 2) \& (agent drink >= (7 + 2 * agent dexterity))} \\
    \textbf{Starting Conditions:}
    \begin{itemize}[noitemsep, topsep=0pt, leftmargin=*]
        \item \textbf{C1:} \code{(agent level == 2) \& (agent adjacent to: BlockType.WATER)} (Req: ApproachWaterLevel2)
        \item \textbf{C2:} \code{(agent level == 2) \& (agent facing: BlockType.WATER)} (Req: FaceWaterLevel2)
    \end{itemize}
\end{tcolorbox}

\begin{tcolorbox}[skillbox, title=MoveAwayFromWaterLevel2]
    \textbf{Skill Type:} Navigation \\
    \textbf{Description:} The agent moves from an adjacent position to a water pool to a location where it is no longer adjacent, enabling exploration of surrounding areas. \\
    \textbf{Assigned Reward:} 1 \\
    \textbf{Success Condition:} \code{(agent level == 2) \& not (agent adjacent to: BlockType.WATER)} \\
    \textbf{Starting Conditions:}
    \begin{itemize}[noitemsep, topsep=0pt, leftmargin=*]
        \item \textbf{C1:} \code{(agent level == 2) \& (agent adjacent to: BlockType.WATER)} (Req: ApproachWaterLevel2)
    \end{itemize}
\end{tcolorbox}

\begin{tcolorbox}[skillbox, title=KillGnomeWarriorInLitAreaLevel2]
    \textbf{Skill Type:} Fighting mobs \\
    \textbf{Description:} The agent, while positioned in a fully illuminated area on the Gnomish Mines level, locates and eliminates a Gnome Warrior. \\
    \textbf{Assigned Reward:} 1 \\
    \textbf{Success Condition:} \code{(agent level == 2) \& agent killed mob: MeleeMobType.GNOME\_WARRIOR} \\
    \textbf{Starting Conditions:}
    \begin{itemize}[noitemsep, topsep=0pt, leftmargin=*]
        \item \textbf{C1:} \code{agent level == 2} (Req: DescendLadderLevel1)
        \item \textbf{C2:} \code{agent inventory sword >= 1} (Req: CraftWoodSword)
        \item \textbf{C3:} \code{(agent level == 2) \& (agent light level >= 0.5)} (Req: MoveIntoLitAreaLevel2)
        \item \textbf{C4:} \code{agent near melee mob: MeleeMobType.GNOME\_WARRIOR} (Req: FindStalagmiteLevel2)
    \end{itemize}
\end{tcolorbox}

\begin{tcolorbox}[skillbox, title=KillGnomeArcherWithSwordLevel2]
    \textbf{Skill Type:} Fighting mobs \\
    \textbf{Description:} The agent leverages its sword and light management to locate and defeat a Gnome Archer in melee range within illuminated areas. \\
    \textbf{Assigned Reward:} 1 \\
    \textbf{Success Condition:} \code{(agent level == 2) \& agent killed mob: RangedMobType.GNOME\_ARCHER} \\
    \textbf{Starting Conditions:}
    \begin{itemize}[noitemsep, topsep=0pt, leftmargin=*]
        \item \textbf{C1:} \code{agent level == 2} (Req: DescendLadderLevel1)
        \item \textbf{C2:} \code{agent inventory sword >= 1} (Req: CraftWoodSword)
        \item \textbf{C3:} \code{(agent level == 2) \& (agent light level >= 0.5)} (Req: MoveIntoLitAreaLevel2)
        \item \textbf{C4:} \code{agent near ranged mob: RangedMobType.GNOME\_ARCHER} (Req: FindStalagmiteLevel2)
    \end{itemize}
\end{tcolorbox}

\begin{tcolorbox}[skillbox, title=DescendLadderLevel2]
    \textbf{Skill Type:} Navigation \\
    \textbf{Description:} The agent executes the descend action while positioned on the downward ladder in the Gnomish Mines level to transition to the Sewers (level 3).  \\
    \textbf{Assigned Reward:} 1 \\
    \textbf{Success Condition:} \code{agent level == 3} \\
    \textbf{Starting Conditions:}
    \begin{itemize}[noitemsep, topsep=0pt, leftmargin=*]
        \item \textbf{C1:} \code{agent level == 2} (Req: DescendLadderLevel1)
        \item \textbf{C2:} \code{(agent level == 2) \& (agent monsters killed level 2 >= 8)} (Req: KillEightMobsLevel2)
        \item \textbf{C3:} \code{(agent level == 2) \& (agent standing on: ItemType.LADDER\_DOWN)} (Req: StepOntoLadderDownLevel2)
    \end{itemize}
\end{tcolorbox}

\end{document}